\documentclass[11pt]{article}

\usepackage{graphicx}
\usepackage{mathptmx}      
\usepackage{amsmath}
\usepackage{bm}
\usepackage{amssymb}
\usepackage{xspace}
\usepackage{multirow}
\usepackage[abs]{overpic}
\usepackage{grffile}		
\usepackage{epstopdf}		
\usepackage[margin=1in]{geometry}
\usepackage[T1]{fontenc}
\usepackage{authblk}

\makeatletter
\DeclareRobustCommand\onedot{\futurelet\@let@token\@onedot}
\newcommand{\@onedot}{\ifx\@let@token.\else.\null\fi\xspace}

\newcommand{\ie}{i.e\onedot}

\newcommand{\etc}{\emph{etc}\onedot}

\newcommand{\etal}{et al\onedot}
\makeatother
\newcommand{\sgn}{\operatorname{sgn}}
\newcommand{\argmax}{\operatornamewithlimits{argmax}}

\begin{document}
\title{The Image Torque Operator for Contour Processing}
\author{Morimichi Nishigaki and Cornelia Ferm\"{u}ller\\
Institute for Advanced Computer Studies, University of Maryland\\ College Park, MD 20742, U.S.A.}

\maketitle

\begin{abstract}
Contours are salient features for image description, but the detection and localization of boundary contours is still considered a challenging problem. This paper introduces a new tool for edge processing implementing  the Gestaltism idea of edge grouping. This tool is
a mid-level  image operator, called the \emph{Torque} operator, that is  designed to help detect closed contours in images.
The torque operator  takes as input the raw image and  creates  an image map by computing from the image gradients within regions of multiple sizes  a measure of how well the edges are aligned to form  closed, convex contours. Fundamental  properties of the torque are explored and illustrated through examples. Then it is applied in pure bottom-up processing in a variety of applications, including  edge detection, visual attention and segmentation and experimentally demonstrated  a useful  tool that can improve existing techniques. Finally, its extension as a more general grouping mechanism and  application  in  object recognition is discussed.
\end{abstract}
\section{Introduction}
\label{Introduction}
Visual scene interpretation is very complex and involves computations  at  different levels of abstraction. Most theorists of vision adapt a categorization of visual processes and representations into low-, mid- and high-level vision \cite{Marr82}.
The idea is that low level vision is about computing  features, such as local edges, color, texture and image motion; mid-level vision groups  local features to obtain object surfaces and  global scene information, such as  3D motion and  lighting; and
high level vision  utilizes semantic information to recognize objects, actions and scenes.
The most influential ideas of mid-level vision are due to the Gestalt theorists. These psychologists of the early twentieth century \cite{Wertheimer1923} argued that human vision organizes  image features at the early stages of interpretation through a process of  figure ground segmentation. They suggested that certain principles are applied to group pieces of image and locate borders of figures.
In this  view, mid-level vision is about implementing organizational principles, such as similarity, symmetry, common fate (i.e. common motion),  closure, bias from prior experience, \etc, in order to 
identify the image regions which are object-related for further processing.

A very important cue of mid-level vision  is the contour. Objects and parts of objects are delineated from their surroundings by closed contours, which make up their boundary. Many papers over the years have focused on contour detection \cite{Martin2004,Ferrari2006,Toshev2012}, and  there has been a renewed interest in recent years. A number of recent papers also have discussed mid-level cues and inspiration from Gestalt theory for contour processing in applications of detection \cite{DollarZitnick2013,Kennedy2011,Ming2012}, recognition \cite{Opelt2006,Lim2013}, and segmentation \cite{Arbelaez2009,Yu2004}. These  works mostly learn from data \cite{Ren2005} to acquire mid-level representations.
Here we pursue  a very different approach; we propose  a bottom-up mid-level mechanism for  grouping edges into closed contours. Thus this mechanisms implements the so-called principle of \emph{closure}, which refers to the idea that objects and object parts are perceived as whole, and simple features tend to be grouped together into closed figures even when they are not complete.

\begin{figure}[htbp]
  \begin{center}
  \begin{tabular}[htbp]{@{}c@{}c@{}}

	\begin{minipage}{0.25\columnwidth}
    \begin{center}
    \includegraphics[width=1\columnwidth,keepaspectratio=true,clip]
    {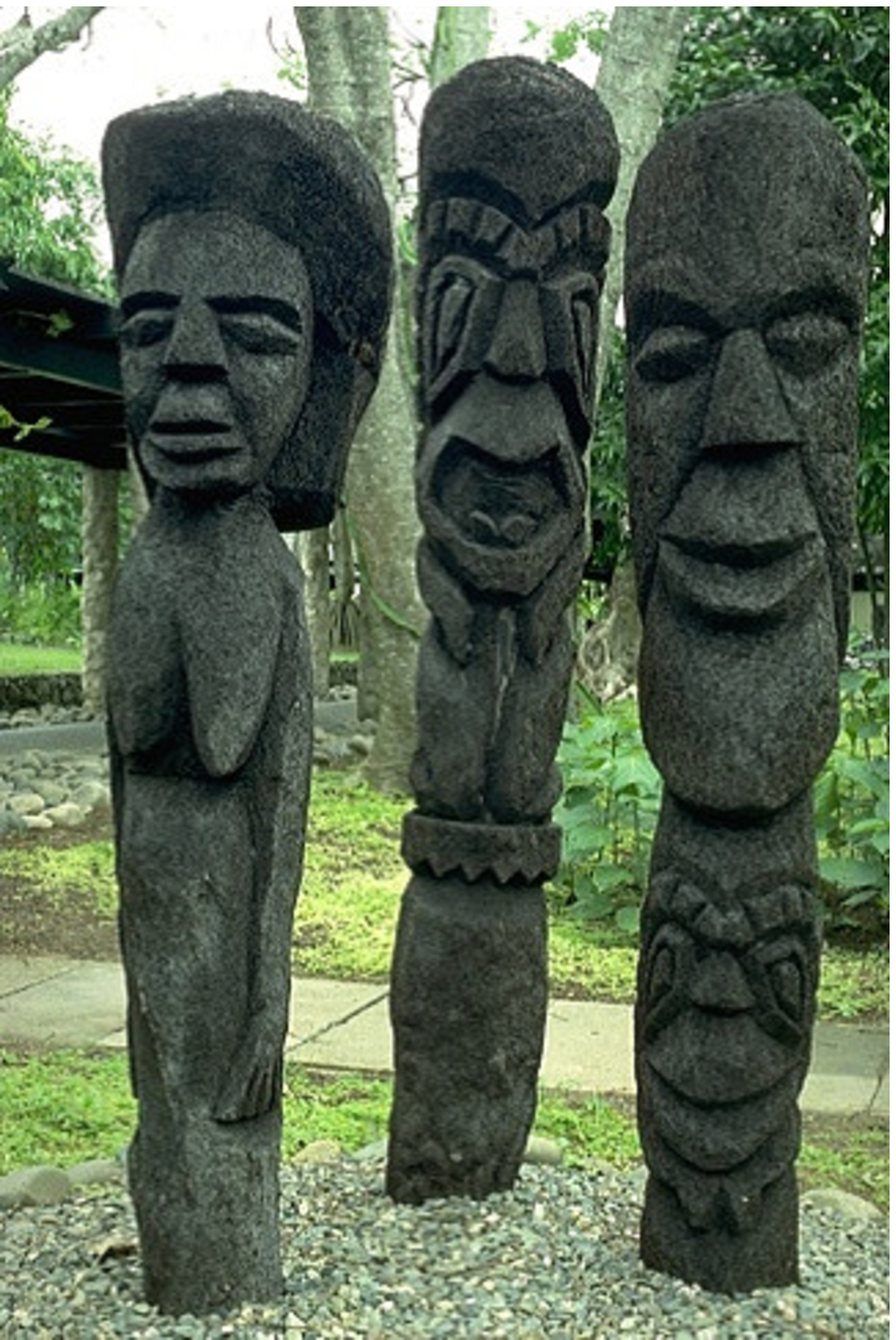}
    \end{center}
    \end{minipage}&

    \begin{minipage}{0.75\columnwidth}
    \begin{center}
    \includegraphics[width=1\columnwidth,keepaspectratio=true,clip]
    {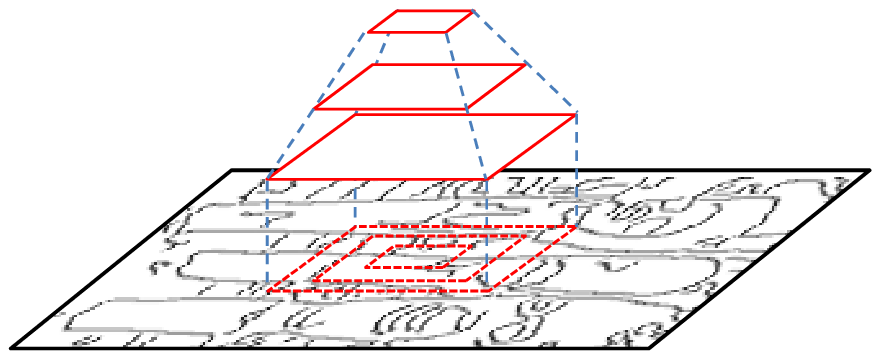}
    \end{center}
    \end{minipage}

  \end{tabular}\vspace{1mm}

  \begin{tabular}[htbp]{@{\extracolsep{0.02\columnwidth}}c@{}c@{}c@{}c@{}}

    \begin{minipage}{0.23\columnwidth}
    \begin{center}
    \includegraphics[width=1\columnwidth,keepaspectratio=true]
    {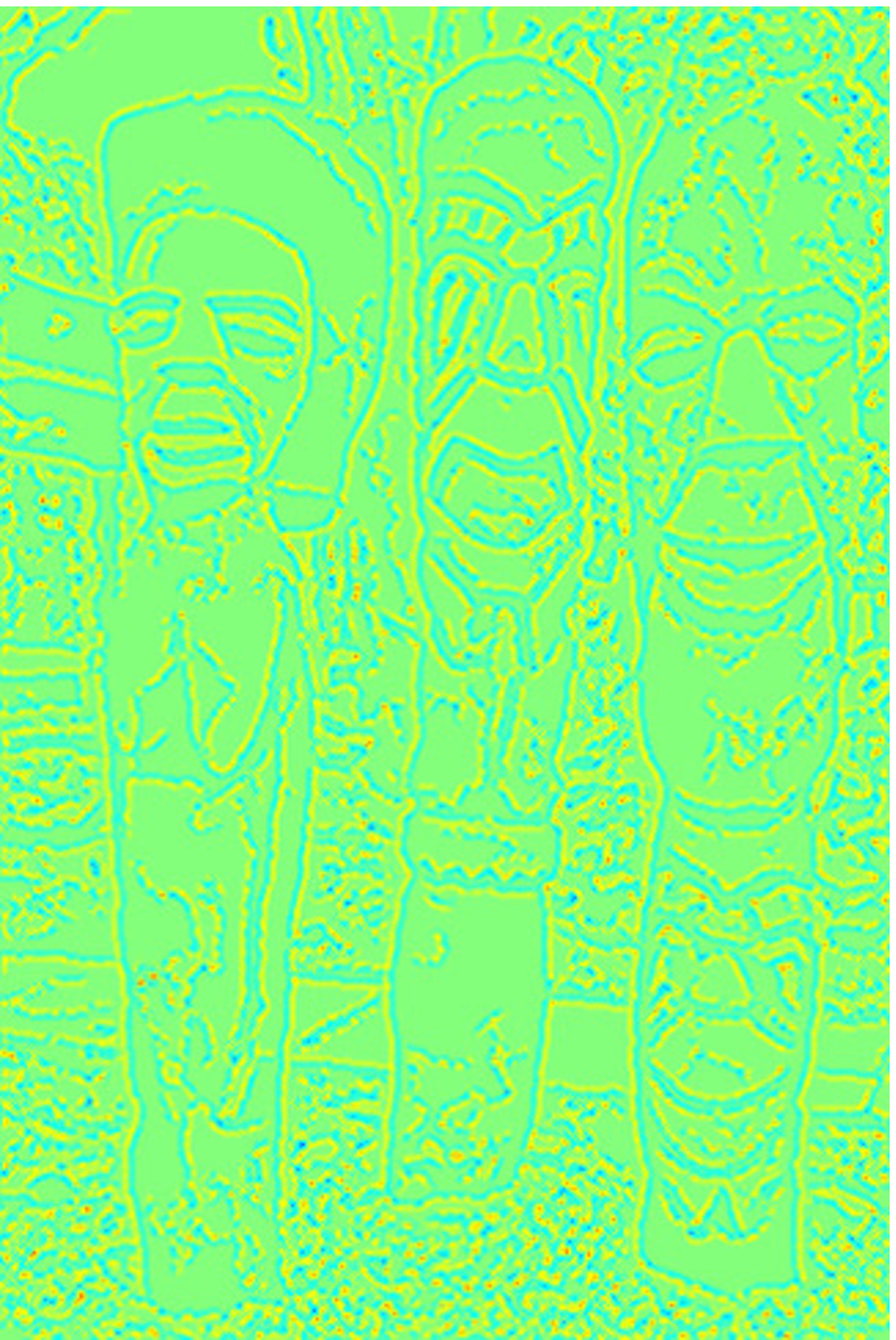}
    \end{center}
    \end{minipage}&

    \begin{minipage}{0.23\columnwidth}
    \begin{center}
    \includegraphics[width=1\columnwidth,keepaspectratio=true]
    {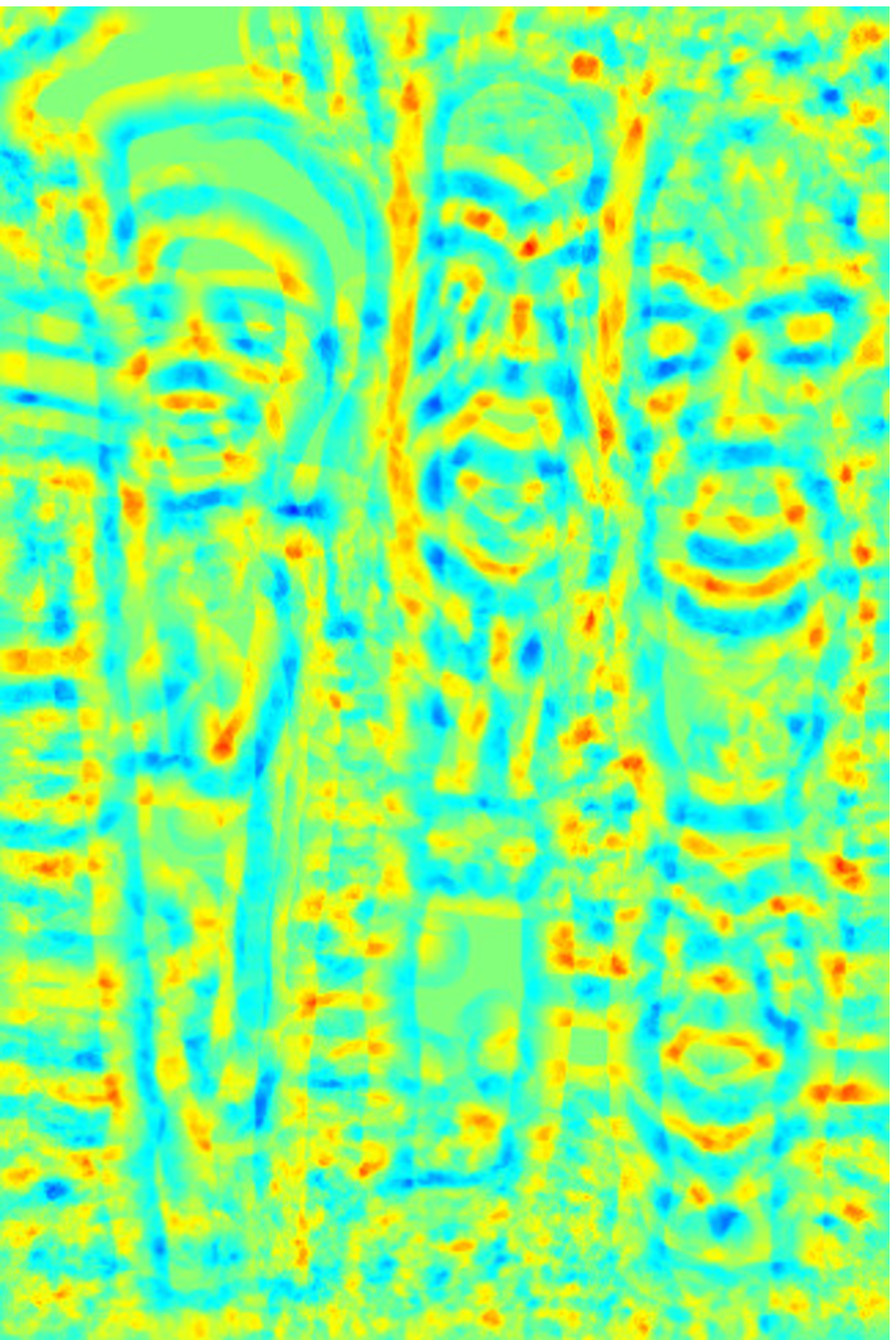}
    \end{center}
    \end{minipage}&

    \begin{minipage}{0.23\columnwidth}
    \begin{center}
    \includegraphics[width=1\columnwidth,keepaspectratio=true]
    {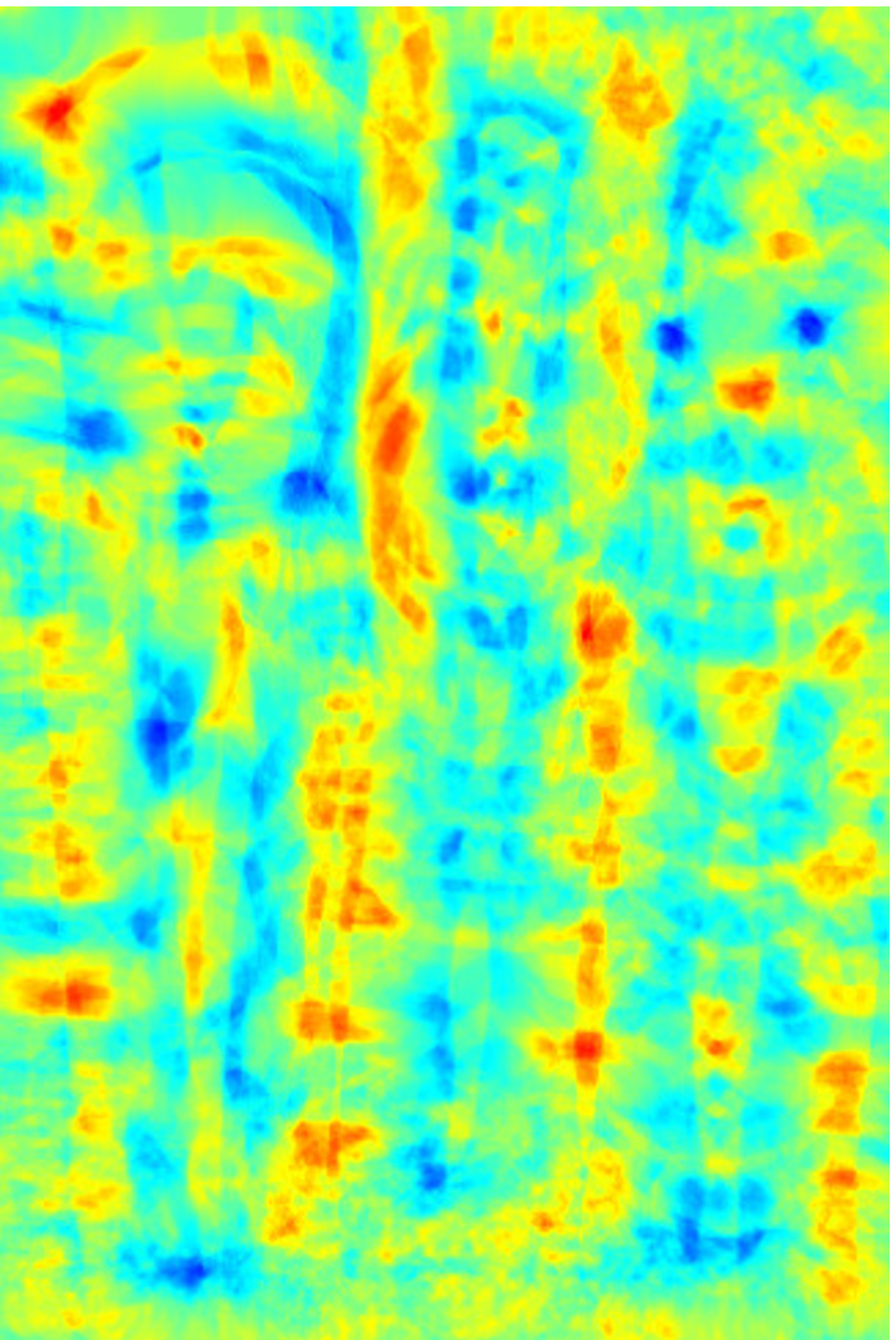}
    \end{center}
    \end{minipage}&

    \begin{minipage}{0.23\columnwidth}
    \begin{center}
    \includegraphics[width=1\columnwidth,keepaspectratio=true]
    {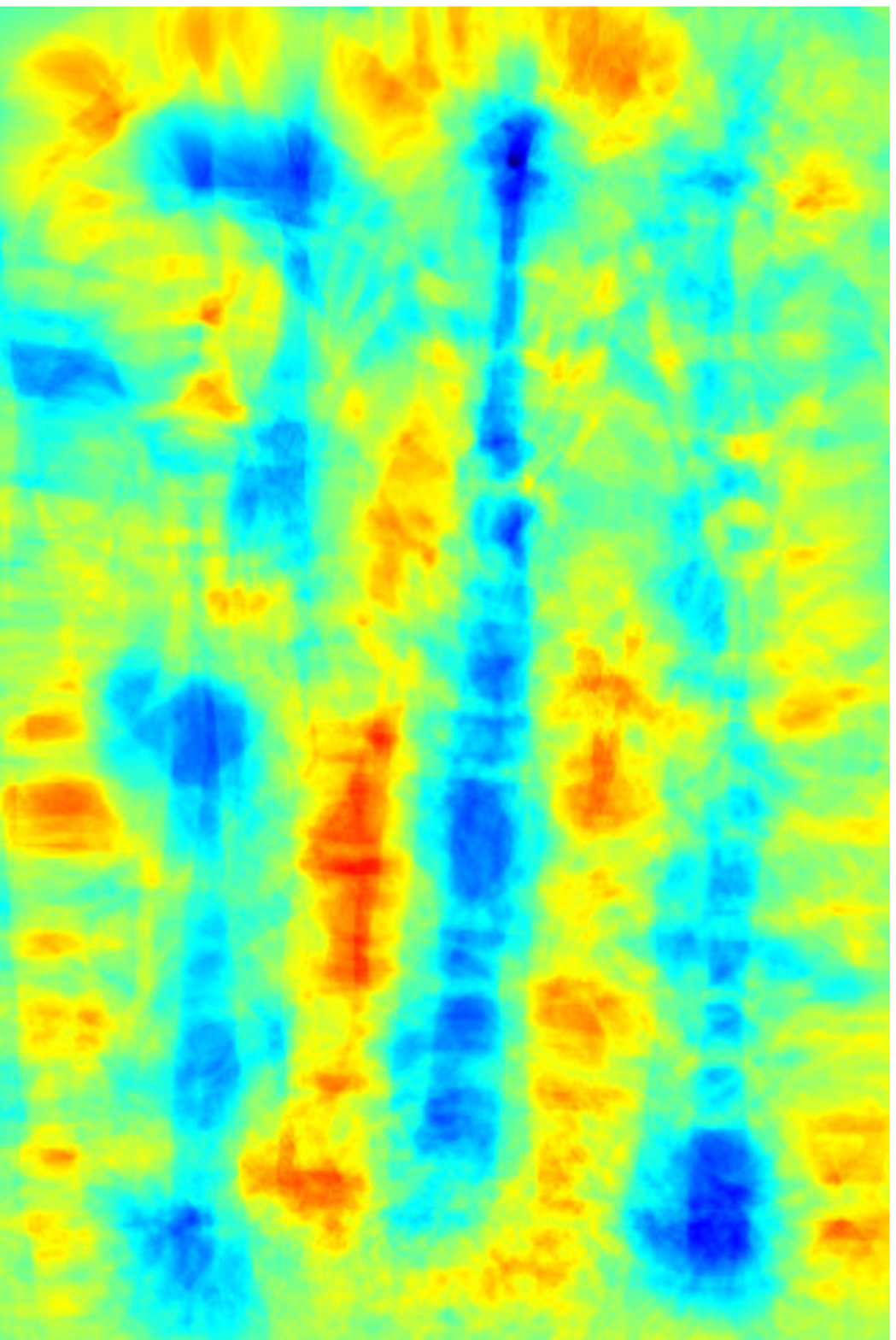}
    \end{center}
    \end{minipage}\\
    5 & 21 & 45 & 81
  \end{tabular}

  \begin{tabular}[htbp]{@{}c@{}c@{}c@{}c@{}}

    \begin{minipage}{0.35\columnwidth}
    \begin{center}
    \includegraphics[width=1\columnwidth,keepaspectratio=true]
    {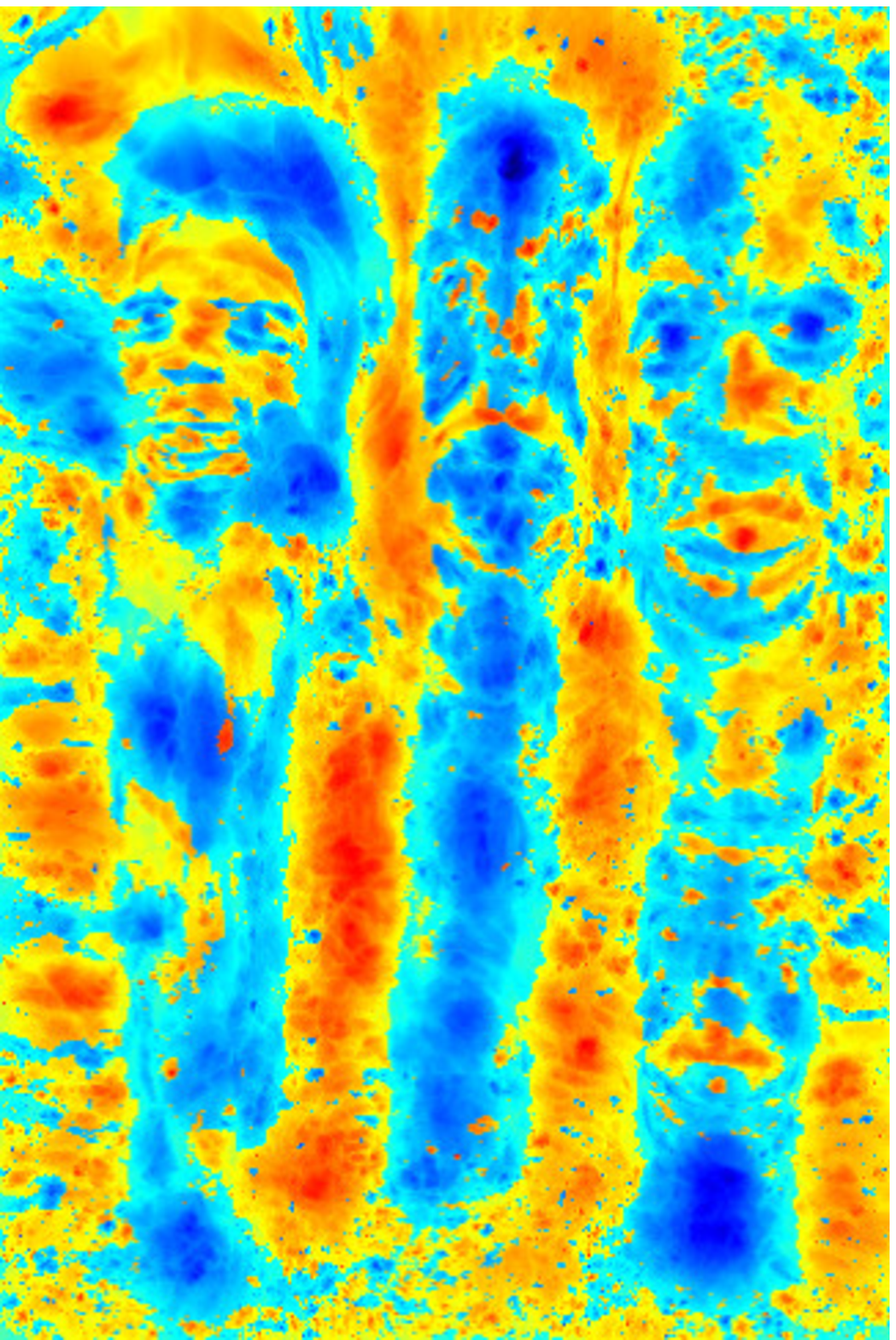}
    \end{center}
    \end{minipage}&

    \begin{minipage}{0.092\columnwidth}
    \begin{flushleft}
    \includegraphics[width=1\columnwidth,keepaspectratio=true]
    {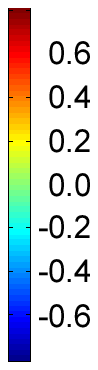}
    \end{flushleft}
    \end{minipage}

    \begin{minipage}{0.35\columnwidth}
    \begin{center}
    \includegraphics[width=1\columnwidth,keepaspectratio=true]
    {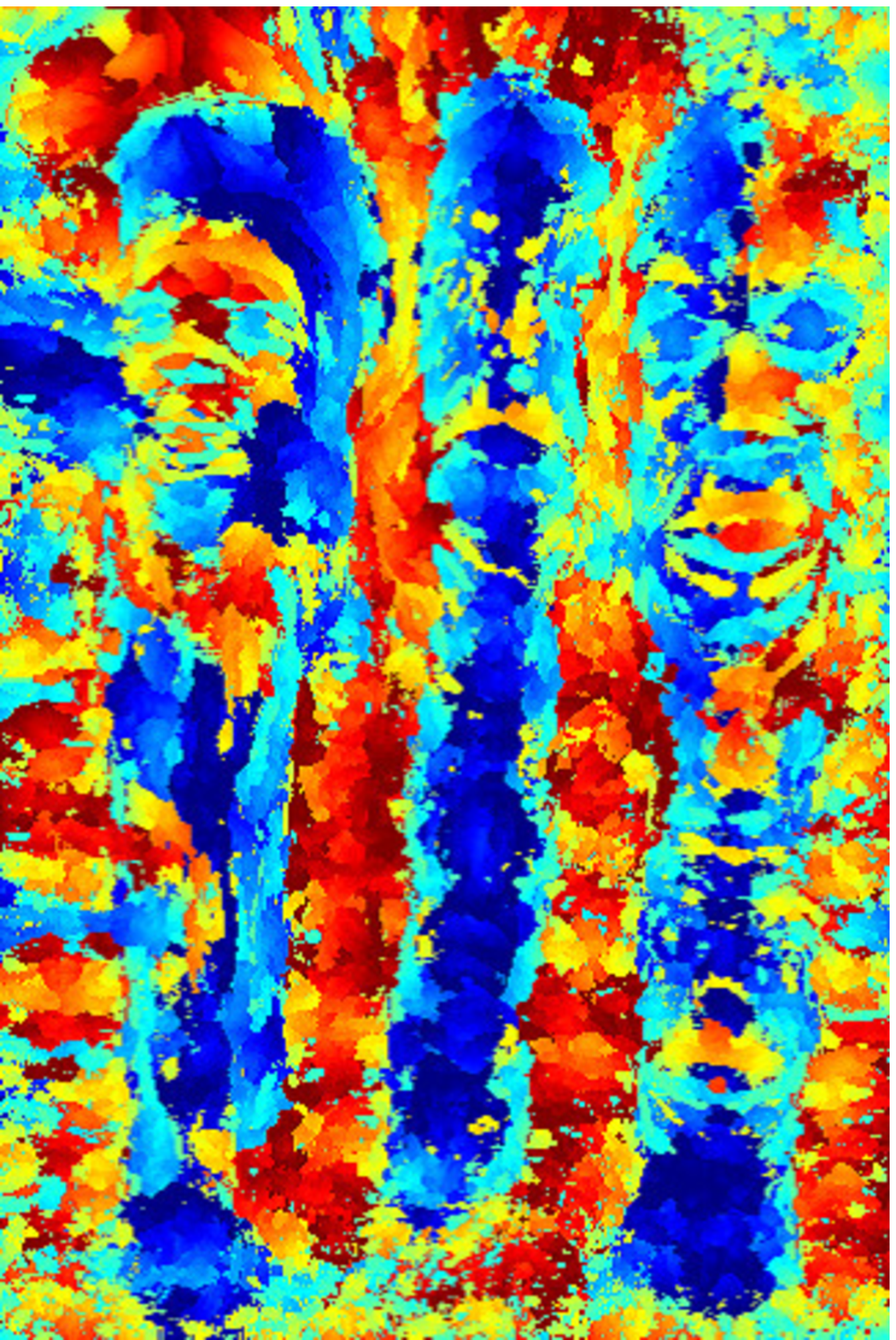}
    \end{center}
    \end{minipage}&

    \begin{minipage}{0.092\columnwidth}
    \begin{flushleft}
    \includegraphics[width=1.0\columnwidth,keepaspectratio=true]
    {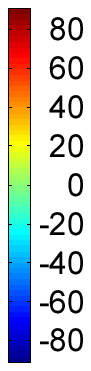}
    \end{flushleft}
    \end{minipage}

  \end{tabular}
  \end{center}
\caption{Illustration of the torque.
Upper row: left: Test image. right: Usage of the torque operator. The torque operator is applied to multi-scale image patches at every pixel in an image.
Middle row: For the test image, examples of torque maps are shown for  four patches of increasing size.
Lower row: Combination of the torque at all patch sizes into one map, called the value map and the corresponding scale map.
\label{fig:Illustration of the torque}}
\end{figure}

The grouping of edges is implemented with a  semi-global image operator, which we call the \emph{torque operator}. This operator is defined on oriented edges and provides a measure of the edge structure within a patch. It takes on  large value when the  patch contains a  closed contour.
 The operator was designed as a tool to help  detect regions likely to contain object or parts of objects. This is achieved by collecting edge information from regions of different extent in a way that enforces edges on convex contours and attenuates random  edges due to texture.
  The torque within a patch is computed by taking  at every edge point the value of  the cross-product of the oriented edge and the vector from the center point of the patch, and summing over all values. To help  the reader  get a quick grasp of the basic idea, the concept is illustrated in
Figure~\ref{fig:Illustration of the torque} before  a proper definition of the operator will be given in Section~\ref{sec:Torque Operator}.

Referring to the figure, for the image on the upper left from the Berkeley data set \cite{Martin2001}, torque maps were  computed for  four different patch sizes ( $5 \times 5$, $21 \times 21$, $45 \times 45$, and $81 \times 81$ ). In these maps every pixel $p$ encodes  the torque value from the patch of   given size centered at $p$.
We used the color coding as explained in the third row. Because edges are oriented, torque values can be positive (shown in red)  and negative (shown in blue).
We then define data structures: We call the three-dimensional structure of the  torque maps at different scale the \emph{torque volume}, and we combine the different scales into  two-dimensional maps, which we
call  the \emph{torque value map} and the \emph{scale map} (as shown in the third row of Fig.~\ref{fig:Illustration of the torque}). The torque value map at every pixel codes the value of largest absolute value over all scales, and the scale map codes the scale and the sign corresponding to the largest absolute torque value.
These data structures will be used as tools to solve a number of applications.

The paper is organized as follows.
After a description of related work in the next section, we provide a formal definition of the torque operator and discussion of its  properties.
Then we apply the torque  operator in a few Computer Vision applications, specifically the problems of boundary detection, visual attention, segmentation, and object recognition, and we verify its usefulness as a tool that can improve existing techniques.
Finally we conclude with a discussion on extending this operator to the spatio-temporal domain and as a grouping mechanism of high-level semantic edge information. Parts of this paper have previously appeared in \cite{Nishigaki2012}.

\section{Related Work}

Contours are an essential cue for  many vision applications, including segmentation, tracking and recognition. By the term contour, we generally refer to  extended edges delineating objects and parts of objects. Already in  the early days of Computer Vision
it became clear that simple point-wise edge responses, computed with filters, are not sufficient to obtain salient edges corresponding to contours; some grouping mechanisms are necessary \cite{RenMalik2002}. Earlier approaches employed  semi-local edge-linking processes to obtain extended edges from edgels computed with local filters.
For example, the well-known Canny edge detector traces edges using hysteresis thresholding. In more recent years the top performing algorithms employ edge detection methods  specifically tuned to boundaries, and they use various  linking and globalization processes, designed either for segmentation or recognition.

{\noindent \emph{Boundary detection:}}
Data-driven approaches to contour detection have been championed in the work of Martin \etal \cite{Martin2004}.
In this study, local cues, such as brightness, color, texture, and their gradients are combined, and weights for each cue are learned using labeled image data sets to distinguish edge points at boundaries from others.
In similar spirit Doll{\'a}r et al.~\cite{Dollar2006} learn edge classifiers from simple features in image patches, Ren \cite{Ren2008} combines information of local operators from multiple scales, and the high performance contour detection algorithm in \cite{Arbelaez2011} includes a globalization process to combine local edges based on the  affinity of distant pixels. The latter method was further improved by  Ren and Bo \cite{RenBo2012}, who replaced the hand-selected features using patches obtained through sparse coding and dictionary learning, and integrated them   through multi-scale pooling in the globalization. Zheng et al. \cite{Zheng2010} proposed an object-specific boundary detector by combining low-level information using the BEL edge detector \cite{Dollar2006} and  foreground background cues, middle-level information about short and long-range connections of pixels, and high-level information from shape priors.

Recently Lim et al. \cite{Lim2013} proposed an interesting generic mid-level detector, which they call sketch tokens. From patches of human-generated contours they learn different classes of edge features. Then using random forests,  edge pixels are detected and classified as the center points of edge tokens. The approach was demonstrated in the bottom-up task of  contour detection and the top-down task of   object detection. In \cite{DollarZitnick2013} a real-time algorithm for edge detection was proposed using the sketch tokens as features of structured information in a random forest \cite{Kontschieder2011}. This way edge detection is formulated as the prediction of local segmentation masks.

Over the years many computational models have been proposed for contour completion \cite{ElderZucker96,GuyMedioni93}. For example in more recent work Kokkinos \cite{Kokkinos2010} first classifies boundary pixels based on SIFT features, and then groups the edge pixels using fractional linear programming based on the edge strength and a smoothness constraint, and Ren \cite{Ren2008} linearly approximates edges  and connects pixels  using a conditional random field (CRF), which  captures the  statistics of continuity and different junction types. Most approaches to  contour completion  model  the  Gestalt rules of proximity and good-continuity. The principle of closure has also been proposed \cite{ElderZucker96}, for example in \cite{Schoenemann2011} for the  task of segmentation.
In recent work Ming et al. \cite{Ming2012} used  a higher-order CRF to model short and long range connections between  edgels and junctions and implemented this way  various edge completion principles, including the one of closure.

Most closely related to our work are the ideas of Lindeberg \cite{Lindeberg1994,Lindeberg1998} and  Craft et al.~\cite{Craft2007}.
In his seminal work on a  computational theory  of  scale space, Lindeberg introduces, among other image feature detectors, the blob detector. Circular blobs in general can be  detected using the Laplacian (or Difference of Gaussian) operator and scale space blobs are computed by filtering the image with scale-normalized Laplacians and detecting  the local extrema in space and time. Lindeberg \cite{Lindeberg1993} also discussed the scale selection mechanism as a tool for selecting the focus of attention. The torque mechanism, introduced here, can be used in a similar way for attention selection, and if implemented with square windows or circular windows, has a very similar behavior. The torque, however, is more general. First, even if implemented with square (circular) windows, the torque value map can detect a larger range of shapes not just circular ones, and if used with differently shaped windows, it can also be tuned to certain structures.
Secondly, we view the torque as a tool that can be used in many applications by using both the value and scale map.

Craft et al. \cite{Craft2007} developed a computational neural model for figure-ground organization using a circular grouping mechanism. Their motivation was to provide an explanatory  mechanism that can account for border-ownership and attention processes. At the first layer  (modeling cells in area V1) oriented edge responses are computed and stored as pairs, one for each orientation. These cells are grouped in a second layer (modeling  area V2) with top-down modulation from a third higher layer of larger-receptive cells, which tunes the grouping so it prefers  annular patterns of different  size (similar to a Laplacian filter). Our torque mechanism works in a similar way, but is more general. Instead of  being tuned only to circular figures, the torque can group any closed figure, and it also has a scale selection mechanism. Craft et al. \cite{Craft2007} proposed their model, not to solve Computer Vision applications, but  to explain neurophysiological data. They also discussed   physiological evidence  in support of their biological model. Since the torque is a grouping mechanism for closure, it is linked to border-ownership and  attention, and all  of the arguments provided in  \cite{Craft2007} also apply to the torque mechanism.  The most important ones are: Early edge signals involved in border-ownership representation code not only direction, but also orientation \cite{Heydt2005}. The speed of  processing and size of fibers \cite{Zhou2000} indicate that  border-ownership is not  implemented through lateral connections, but  rather requires a processing in higher order areas that feed back the signal. The processing time in border ownership signals was found  not to dependent on the  figure size. Finally, since figure-ground processes  interact with higher-level processes, it must play a role in visual attention selection, and the two processes must be closely related. Our torque mechanism has all these features: (a) it uses orientation of edges; (b) is an additional mechanism (higher level than edge-linking) which can feed back to simple edges; (c) as a scale-space mechanism its computation does not depend on the size of object, and (d) it lends itself  naturally as a tool for attention selection.


{\noindent \emph{Segmentation:}}
The fundamental approach to segmentation is to separate surfaces, assuming that  individual  surfaces are homogeneous in some local measurements, such as color intensity, texture \cite{Alpert2007,Shi2000,Bitsakos2010}, motion \cite{Criminisi2006,Yin2007,Ogale2005a,Fermuller99} or depth \cite{Kolmogorov2005,Woo2000}, and neighboring surfaces are separated by discontinuities in these cues.
We here differentiate between approaches that treat segmentation as the problem of dividing
the image into multiple regions, and approaches that consider the problem as separating one foreground object from background.
Examples of the former are the mean-shift method \cite{Comaniciu2002}
and graph-partitioning methods using the  graph cuts algorithm \cite{Boykov2004} or the normalized cut approach \cite{Shi2000}.
These segmentation approaches are usually based on local cues  as input to a global optimization, but recently many methods first compute super-pixels \cite{Ren2003,Kohli2009} by over-segmenting the image into perceptual uniform  regions based on the statistics in neighborhoods or affinity between points.

Approaches that consider foreground-background segmentation include probabilistic models that formulate the problem as optimization for a binary labeling and use belief propagation  or graph cut  algorithms \cite{Rother2004,Kolmogorov2005} and continuous models  based on differential equations using methods such as active contours and variational approaches \cite{Blake2000,Mumford1989,Osher1988}. Our evaluation here uses graphcut methods. We use the classical graphcut segmentation into multiple segments \cite{Boykov2001} and the foreground-background separation approach of \cite{Mishra2009b}. The latter method segments in the polar coordinate system by minimizing for  a closed contour surrounding a fixation point using a graph cut formulation defined on edges only.
All segmentation approaches have the problem of being biased, usually towards small regions with small and smooth boundary.
This is because of texture edges, which in real images are always present, in conjunction with  minimizations biased towards certain shapes.
For example graph cuts \cite{Kolmogorov2005} are known to favor small areas, the polar coordinate representation favors circular blobs, and variational minimizations \cite{Vese2002} also prefer smaller segments, as they explicitly minimize the length and/or smoothness of the bounding contour. By using the  torque  in a preprocessing step to find object-like regions,
 it can  alleviate the bias.  It can help  locate  the regions surrounded by contours, locate contours of larger extent, and separate texture edges from boundary edges.

{\noindent \emph{Visual attention:}}
Attention mechanisms are classified into bottom-up and top-down processes.
Top-down attention is more complex because it represents objects in memory \cite{Hollingworth2001} and uses the memory to detect likely objects in an attended visual scene.
Bottom up attention is driven by low level processes.
The best  known  model of visual attention has been proposed by Itti \etal \cite{Itti1998}.
In this model, first local feature maps are computed from color, intensity and orientation information as the difference of Gaussian filters at multiple scales, which approximate the \emph{center-surround} differences of neurons.
Larger center surround differences are considered more \emph{conspicuous}.
Then a combined saliency map is constructed by adding the normalized feature maps. Related approaches differ in the choice of feature vectors and combination of features.
In our experimental section we will compare against the method of Harel \etal \cite{Harel2006}, which computes a saliency map based on the dissimilarity of features in regions using a graph-based approach.
Harel \etal evaluated the performance of their detector on its ability to predict  human attention  using the  human fixation data of Einh\"auser \etal \cite{Einhauser2006}, and  reported to achieve  98\% of the ROC area of a human based control, while the model by Itti \etal achieved only 84\%.
Recent works in fixation and attention \cite{Holm2008,Einhauser2008}  offer an alternative to the traditional \emph{early} feature saliency theories.
Based on  systematic psychophysical experiments, \cite{Holm2008} suggests that observers look at objects as if they knew them before they became aware of their identity, and \cite{Einhauser2008} shows that the hypothesis that humans attend to objects has a better predictive power in the data than any other theory. A number of works recently proposed algorithms for the implementation of this idea calling it proto-segmentation and object proposal \cite{Alexe2012,Cheng2014,Hosang2016}.
In our paper we will use the torque measure to derive a saliency map for visual attention.
The extrema of the torque measure often appear at the locations in the image where objects are surrounded by  edges.
Thus the torque appears to be a good measure to model object driven visual attention.

{\noindent \emph{Object recognition:}}
 The best known object recognition methods use descriptors  based on point detectors  tuned to texture features \cite{Lowe2004,DT05}, but object  recognition from contours has also received significant attention. We can roughly classify existing   approaches  according to how they  detect and describe  local contour features and how they  represent and classify  the overall contour.
Often, so-called contour patches, or shape  fragments, are  used as  local descriptors. For example, the shape context descriptor \cite{Belongie2002} encodes the spatial distribution of edge points in log polar space, or the feature detector defined by \cite{Jurie2004} is based on the saliency of local convexity. Some approaches  acquire the contour fragments and their detectors \cite{Kumar2004,Shotton2005,Opelt2006,Leibe2004,Mairal2008} from data using learning techniques. For, example Shotton et al. \cite{Shotton2005} and  Opelt et al. \cite{Opelt2006} learn a codebook of shape fragments. To detect objects, the learned class specific shape fragments are then matched using oriented chamfer matching and voted via a star-shape model.  Leibe et al. \cite{Leibe2004} encode with the patches the relative location to the  object center, to create a codebook that in addition to appearance also represents  spatial information for particular object classes.
 For a better representation of object classes, some techniques  group contour pieces into longer lines and curves \cite{Ferrari2006,Ravishankar2008} and match object  parts. A descriptor for matching partial shape fragments was introduced in \cite{Riemenschneider2010}, and
 Toshev et. al.~\cite{Toshev2010} proposed the \emph{chordiogram} descriptor to encode  relative angles of boundary segments. Our contribution to recognition here is a contour-based patch detector and a patch descriptor. The  contour patch detector is based on extrema in the  torque map, and the contour patch descriptor is is based on  the torque values in the detected patch at multiple scales. We then use a simple bag-of-word representation and an SVM for object classification.

\section{Torque Operator}
\label{sec:Torque Operator}
After providing a definition of the torque operator,  we discuss its properties that make it useful for contour processing. Then  we provide the data-structure for  representing the torque, illustrate its application on various examples, and  discuss issues related to scale selection. Finally, we provide an efficient implementation of the torque using the so-called integral images.
\subsection{Definition}

\begin{figure}[tb]
  \begin{center}
    \begin{overpic}[width=0.695\columnwidth,keepaspectratio=true,clip]
	{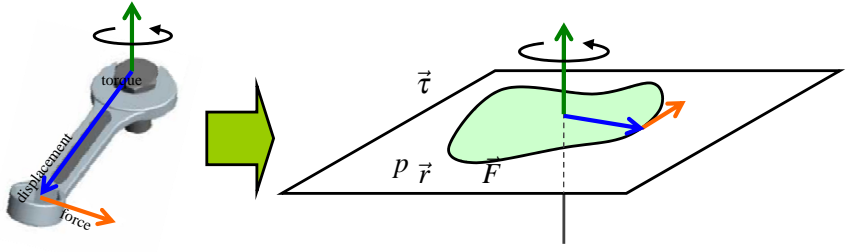}
    \put(154,39){$p$}
    \put(164,35){$\vec{r}$}
    \put(188,35){$\vec{F}$}
    \put(163,69){$\vec{\tau}$}
    \put( 11,27){\rotatebox{53}{\tiny displacement}}
    \put( 28,22){\rotatebox{-25}{\tiny force}}
    \put( 44,72){\tiny torque}
    \end{overpic}
  \end{center}
\caption{The idea of the image torque measure is inspired by the concept of the torque in physics: Given a point $p$ and an edge point $q$ in an image, we assign a unit force vector $\vec{F}$ along the tangent of the edge. Denoting as $\vec{r}$ the vector from $p$ to $q$, the torque vector at $q$ is defined as $\vec{\tau} = \vec{r} \times \vec{F}$.
Its value along the axis perpendicular the image will be called the \emph{torque}.
\label{fig:Image torque idea}}
\end{figure}

Torque, as defined in physics, is the tendency of a force to rotate an object about an axis.
Mathematically, torque is defined as the cross product of the force and the displacement vector from the point at which  torque is measured to the point to  which  the force is applied, as depicted in Fig.~\ref{fig:Image torque idea}:
\begin{align}
  \vec{\tau} = \vec{r} \times \vec{F},\label{eq:torque}
\end{align}
where $\vec{\tau}$ is the torque vector,
$\vec{r}$ is the displacement vector,
and $\vec{F}$ is the force vector.

To define a torque measure for images,
we consider  forces  applied at  edge points and parallel to the tangent of the local edge.
For an arbitrary point, called the center point, we consider a rotation axis in three-dimensional space passing through that point and perpendicular to the image plane.
Then we can measure the torque at any point in the image with respect to the rotation axis, as  defined in physics (see Fig.~\ref{fig:Image torque idea}).
Since the displacement vector and the force vector are both on the image surface, the torque vector is  perpendicular to the image surface, and we call the magnitude of the torque vector along the rotation axis simply the \emph{torque} or \emph{torque value} here and after.

Since our images are discrete, edges are  represented by a set of pixels. Let  $q$ be an edge point, whose edge we represent by  a vector $\vec{F}_q$, and let  $p$ denote the center point and $\vec{r}_{pq}$  the displacement vector from $p$ to $q$. Then the torque value at $p$ is
\begin{align}
\vec{\tau}_{pq} = \vec{r}_{pq} \times \vec{F}_q.\label{eq:torque_vector}
\end{align}
 With slight abuse of notation, here  the cross product of  two-dimensional vectors denotes the scalar obtained by cross-multiplying the vectors. If we  further assume the force vector  of unit length, the value of the torque amounts to:
 \begin{align}
\tau_{pq} = \left\|\vec{r}_{pq}\right\|\sin\theta_{pq}.\label{eq:torque for edge point}
\end{align}
Note that in our definition edges are oriented, where the orientation is defined by contrast.
Thus, the value of the torque can have positive and negative values.
We define the orientation of an edge  as perpendicular clockwise to the image gradient, such that the brighter side is on its right and the darker side on its left.

We then define an image operator on  local image patches.
The \emph{torque operator}  is defined as the sum over the torque values of  all edge points within an image patch of arbitrary shape. Applying the torque operator, $\tau_P$, to a patch  we obtain the \emph{torque of an image patch} as:
\begin{align}
\tau_{P,p} = \frac{1}{2\left|P\right|}\sum_{q\in E\left(P\right)}\tau_{pq},\label{eq:torque for patch pixel-wise}
\end{align}
where $E\left(P\right)$ is a set of edge points in the patch $P$, and $p$ is the center point of the patch.
$\left|P\right|$ is the area of the patch,
which is used for normalization to achieve independence of the patch size. This will become clear in the next section, where it is shown that the torque of a patch is related to the area under the curve.

We apply the torque operator over the entire edge map, shifting the position of the patch and  varying the size of the patch as depicted in Fig.~\ref{fig:Illustration of the torque}.
In principle, the shape of the patch could be arbitrary, but in this paper we will use disk or square patches for illustration (Sec. \ref{sec:Torque Operator}), and square and rectangle patches in our efficient implementation and the experiments (Sec. \ref{sec:app}).

\subsection{Properties of Torque Operator}

Next some of the basic properties of the torque operator are explained to motivate the usefulness of the torque in different applications.

\subsubsection{Torque and Area}

\begin{figure}[tb]
  \begin{center}
    \begin{overpic}[width=0.521\columnwidth,keepaspectratio=true,clip]
    {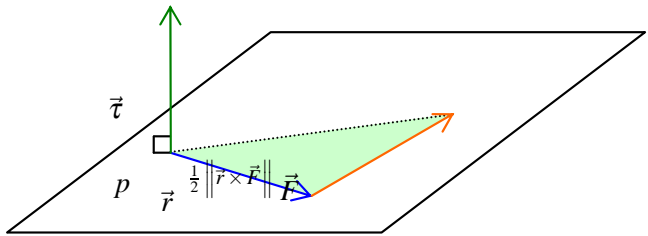}
    \put( 42,17){$p$}
    \put( 60,10){$\vec{r}$}
    \put(105,14){$\vec{F}$}
    \put( 40,45){$\vec{\tau}$}
    \put( 70,20){\scriptsize{$\frac{1}{2}\left\|\vec{r}\times \vec{F}\right\|$}}
    \end{overpic}
  \end{center}
\caption{Cross product and area.
The triangle enclosed by  vectors $\vec{r}$ and $\vec{F}$ is equivalent to $\left\|\vec{r}\times \vec{F}\right\|/2$.
\label{fig:Cross Product and Area}}
\end{figure}

\begin{figure}[tb]
  \begin{center}
  \begin{tabular}[htbp]{cc}
    \begin{minipage}{0.3\columnwidth}
    \begin{center}
    \begin{overpic}[width=1\columnwidth,keepaspectratio=true,clip]
    {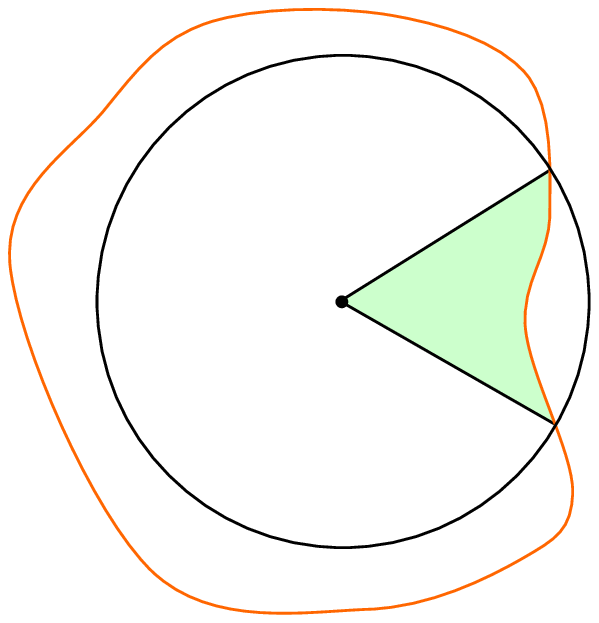}
      \put( 35, 30){$p$}
      \put( 70, 50){$q_1$}
      \put( 70, 20){$q_2$}
    \end{overpic}
    \end{center}
    \end{minipage}&

    \begin{minipage}{0.3\columnwidth}
    \begin{center}
    \begin{overpic}[width=1\columnwidth,keepaspectratio=true,clip]
    {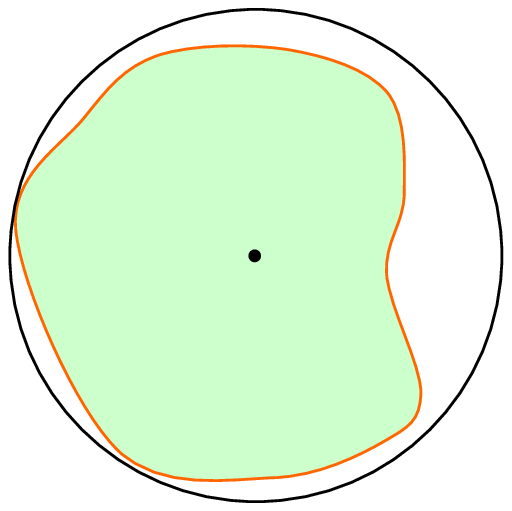}
      \put( 35, 30){$p$}
    \end{overpic}
    \end{center}
    \end{minipage}\\

    (a) & (b)
  \end{tabular}
  \end{center}
\caption{Relationship between torque and area.
Two cases are shown: (a) the disk patch is smaller than the object, and it covers only a part of the boundary; (b) the disk patch covers the whole object boundary.
\label{fig:Torque and Area}}
\end{figure}

Since the torque is defined by the cross product of vectors, it is essentially related to the area defined by these vectors as shown in Fig.~\ref{fig:Cross Product and Area}.
This relationship can  be easily extended to edge curves.
Assuming edges are clean continuous curves, the value  of the torque in a patch is related to the position of the curves in the patch and their shape. For curve segments intersecting the boundary of the patch with center $p$ at two intersection points $q_1$ and $q_2$ as in Fig.~\ref{fig:Torque and Area}(a), the torque is proportional to the area enclosed by the edge curve and the two line segments $\overline{p q_1}$ and $\overline{p q_2}$. The torque of a closed curve, completely inside the patch, is proportional to the  area under the curve (Fig.~\ref{fig:Torque and Area} (b)). The closer  the curve to the patch boundaries, the larger the torque of the image patch. We normalize for the patch size, so we can compare the torque across scales.
Intuitively, it then can be understood  that the  torque operator can be useful for finding the scale of closed curves.

\subsubsection{Texture vs Boundary}
\begin{figure}[tb]
  \begin{center}
  \begin{tabular}[htbp]{cccc}
    \begin{minipage}{0.2\columnwidth}
    \begin{center}
    \includegraphics[width=1.0\columnwidth,keepaspectratio=true,clip]
    {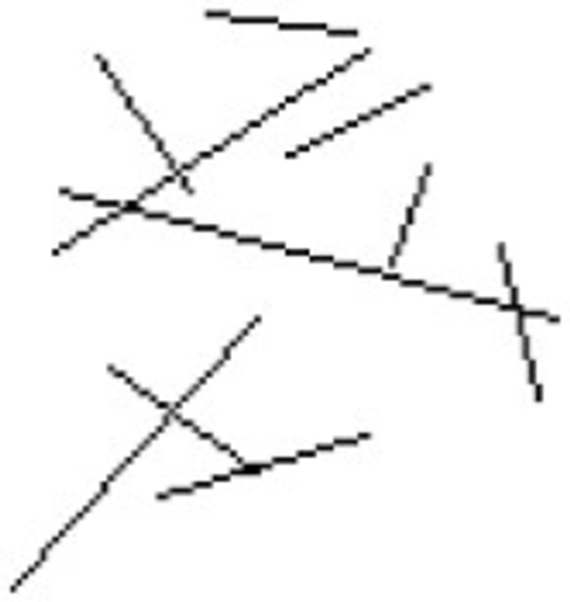}
    \end{center}
    \end{minipage}&

    \begin{minipage}{0.2\columnwidth}
    \begin{center}
    \begin{overpic}[width=1.0\columnwidth,keepaspectratio=true,clip]
    {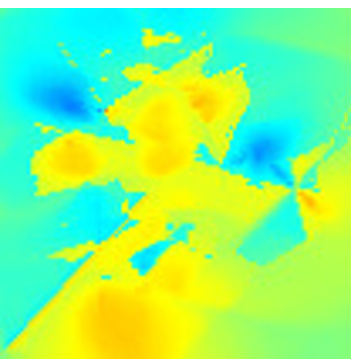}
      \put(  5,12){Max torque}
      \put( 20, 2){0.39}
    \end{overpic}
    \end{center}
    \end{minipage}&

    \begin{minipage}{0.2\columnwidth}
    \begin{center}
    \includegraphics[width=1.0\columnwidth,keepaspectratio=true,clip]
    {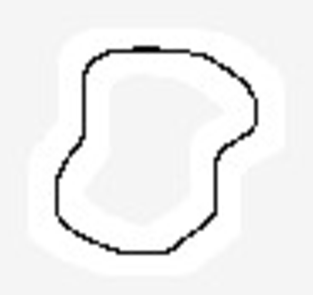}
    \end{center}
    \end{minipage}&

    \begin{minipage}{0.2\columnwidth}
    \begin{center}
    \begin{overpic}[width=1.0\columnwidth,keepaspectratio=true,clip]
    {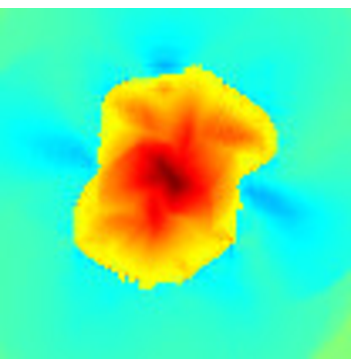}
      \put(  5, 12){Max torque}
      \put( 20,  2){0.79}
    \end{overpic}
    \end{center}
    \end{minipage}\\

    (a) & (b) & (c) & (d)
  \end{tabular}
  \end{center}
\caption{Torque on texture edges and boundary edges. The maximum absolute torque value (over all patch sizes) tends to be small for random texture edges (a and b) and large for closed boundaries (c and d).
\label{fig:Torque on Texture Edges and Boundary Edges}}
\end{figure}

 An important function of the torque as a mid-level operator is to separate aligned edge structures from random texture edges.
The torque of  a patch will be larger when the patch has long contours and it will be largest if the edges  are structured into closed contours. On the other hand, the torque is expected to be small when the edges are due to random texture as illustrated in Fig.~\ref{fig:Torque on Texture Edges and Boundary Edges}. This is, because in our  definition of the torque, edges have an orientation defined by the contrast. Therefore,  in textures (made up of blobs and textons)  where oriented edges in the patch appear in pairs, they cancel  each other. Assuming there is sufficient randomness in the  edge distribution, the  sum of contributions will be small.

\subsubsection{Extrema in Torque}
\label{sec:Extrema in Torque}

\begin{figure}[tb]
\begin{center}

  \begin{tabular}[htbp]{@{}c@{}c@{}c@{}c@{}c@{}}
    \begin{minipage}{0.15\columnwidth}
    \begin{center}
    \includegraphics[width=1.0\columnwidth,keepaspectratio=true]
    {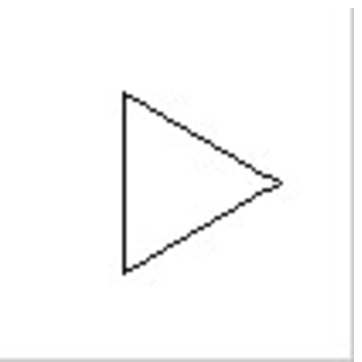}
    \end{center}
    \end{minipage}&

    \begin{minipage}{0.15\columnwidth}
    \begin{center}
    \includegraphics[width=1.0\columnwidth,keepaspectratio=true]
    {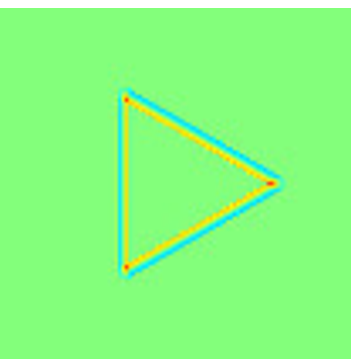}
    \end{center}
    \end{minipage}&

    \begin{minipage}{0.15\columnwidth}
    \begin{center}
    \includegraphics[width=1.0\columnwidth,keepaspectratio=true]
    {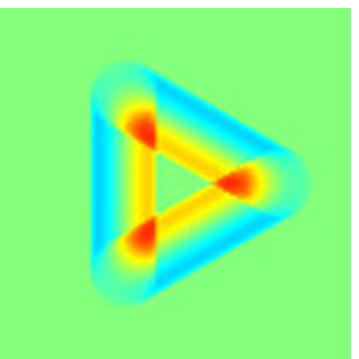}
    \end{center}
    \end{minipage}&

    \begin{minipage}{0.15\columnwidth}
    \begin{center}
    \includegraphics[width=1.0\columnwidth,keepaspectratio=true]
    {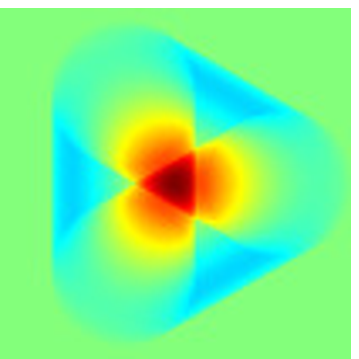}
    \end{center}
    \end{minipage}&

    \begin{minipage}{0.15\columnwidth}
    \begin{center}
    \includegraphics[width=1.0\columnwidth,keepaspectratio=true]
    {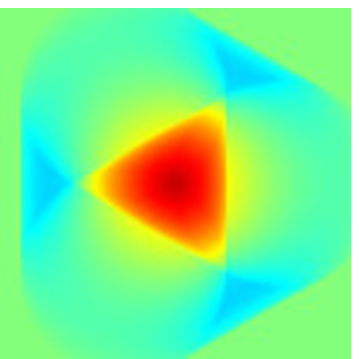}
    \end{center}
    \end{minipage}\\

    & 2 & 10 & 21 & 30
  \end{tabular}

  \begin{tabular}{@{}c@{}}

    \begin{minipage}{0.73\columnwidth}
    \begin{overpic}[width=1\textwidth,keepaspectratio=true,clip]
    {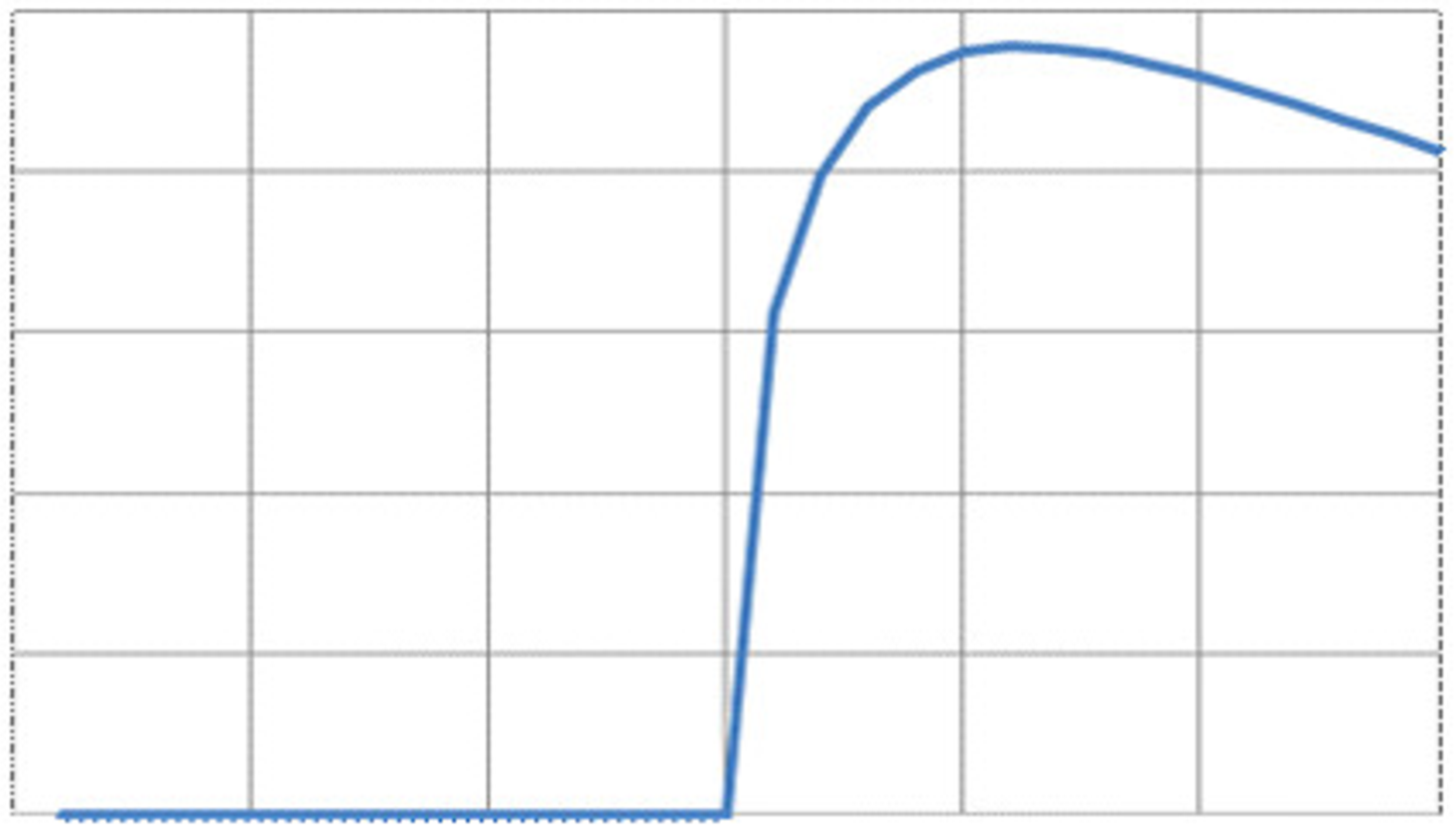}
    \put(  5,  0){$0$}
    \put( 31,  0){$5$}
    \put( 56,  0){$10$}
    \put( 83,  0){$15$}
    \put(109,  0){$20$}
    \put(135,  0){$25$}
    \put(163,  0){$30$}
    \put( 50, -8){Patch Radius (Scale)}
    \put( -5,  5){$0.0$}
    \put( -5, 23){$0.1$}
    \put( -5, 41){$0.2$}
    \put( -5, 59){$0.3$}
    \put( -5, 77){$0.4$}
    \put( -5, 95){$0.5$}
    \put(-13, 40){\rotatebox{90}{Torque}}
	\end{overpic}
    \end{minipage}

  \end{tabular}

\end{center}
\caption{First row: Torque maps for a triangle for  different patch sizes.
The numbers below the torque maps denote  the radius of the disk patch in pixels.
Second row: Torque values at the center of the triangle shown over patch sizes.
\label{fig:Torque Maps for Triangle over Patch Sizes}}
\end{figure}

The torque  tends to be large in magnitude if the patch contains extended contours close to
the boundaries of the patch.
Therefore, it is expected that the torque measure is useful for  finding the locations in the image  where edges are in structure forming closed contours. Furthermore, the torque can give us  the scale of the region of those  structured edges. This concepts is illustrated in
Figure~\ref{fig:Torque Maps for Triangle over Patch Sizes} for  an example where the structured edges form a triangle. The figure shows the torque value maps for four sizes, and graphs the torque values at the center of the triangle over all patch sizes. As can be seen, the location of structured edges, \ie the triangle, can be inferred from the maximum of the torque over space and patch sizes. The patch size of the torque maximum indicates the size of the triangle.

Next we formally define the  data-structures used in the torque computation, then we provide further examples illustrating the torque value on real objects of different shape.

\subsection{Representation of the Torque}
\label{sec:rep}
By applying the torque operator to an image using  a single patch size, we obtain a two-dimensional map of torque values.  Applying the torque operator using  multiple  patch sizes, we obtain multiple maps at different scale
(see Fig.~\ref{fig:Illustration of the torque}).
The set of maps at  different scales makes a  three-dimensional volume.
However, it is often convenient for applications to combine the  maps at different scales into  two dimensional maps, $V$ and $S$, which we describe by the following equations:
\begin{align}
V\left(x,y\right) &= \tau\left(x, y, \hat{s}\left(x,y\right)\right),\\
S\left(x,y\right) &= \sgn\left(V\left(x,y\right)\right)\cdot \hat{s}\left(x,y\right),\\
\hat{s}\left(x,y\right) &= \argmax_s \left|\tau\left(x,y,s\right)\right|.
\end{align}
$\tau\left(x,y,s\right)$ is the torque value at point $\left(x,y\right)$ with patch size (scale $s$). We call the three-dimensional volume of $\tau$ the  \emph{torque volume}, and $V$ and $S$ the  \emph{torque value map} and \emph{scale map}, respectively.

\begin{figure}[tb]
\begin{center}
  \begin{tabular}[htbp]{@{}c@{}c@{}c@{}c@{}c@{}c@{}}
    \begin{minipage}{0.166\columnwidth}
    \begin{center}
    \includegraphics[width=1.0\columnwidth,keepaspectratio=true]
    {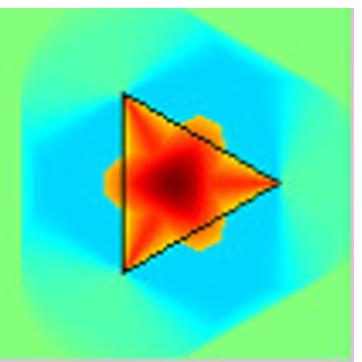}
    \end{center}
    \end{minipage}&

    \begin{minipage}{0.166\columnwidth}
    \begin{center}
    \includegraphics[width=1.0\columnwidth,keepaspectratio=true]
    {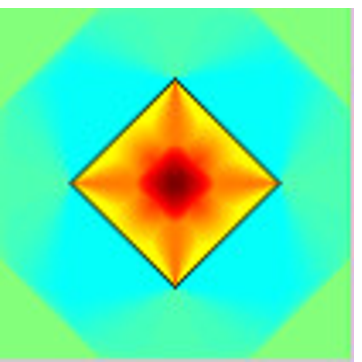}
    \end{center}
    \end{minipage}&

    \begin{minipage}{0.166\columnwidth}
    \begin{center}
    \includegraphics[width=1.0\columnwidth,keepaspectratio=true]
    {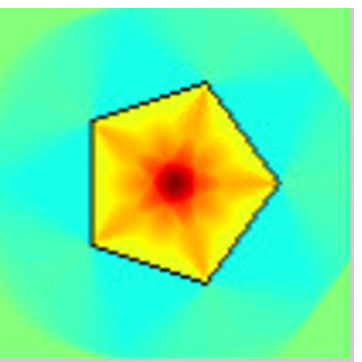}
    \end{center}
    \end{minipage}&

    \begin{minipage}{0.166\columnwidth}
    \begin{center}
    \includegraphics[width=1.0\columnwidth,keepaspectratio=true]
    {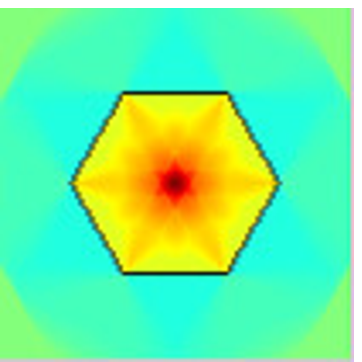}
    \end{center}
    \end{minipage}&

    \begin{minipage}{0.166\columnwidth}
    \begin{center}
    \includegraphics[width=1.0\columnwidth,keepaspectratio=true]
    {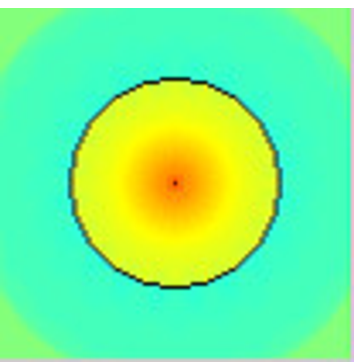}
    \end{center}
    \end{minipage}&

    \begin{minipage}{0.166\columnwidth}
    \begin{center}
    \includegraphics[width=1.0\columnwidth,keepaspectratio=true]
    {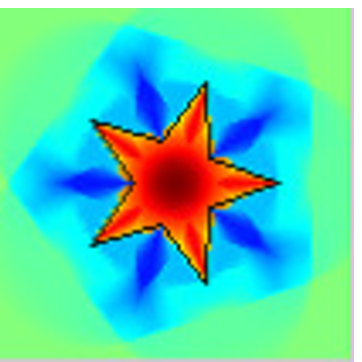}
    \end{center}
    \end{minipage}\\

    \begin{minipage}{0.166\columnwidth}
    \begin{center}
    \includegraphics[width=1.0\columnwidth,keepaspectratio=true]
    {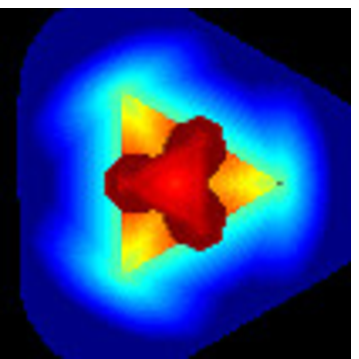}
    \end{center}
    \end{minipage}&

    \begin{minipage}{0.166\columnwidth}
    \begin{center}
    \includegraphics[width=1.0\columnwidth,keepaspectratio=true]
    {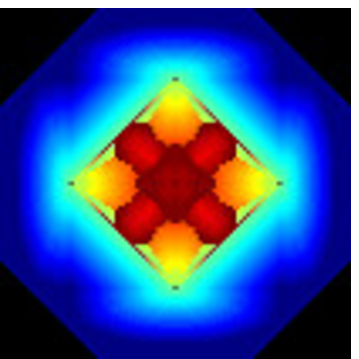}
    \end{center}
    \end{minipage}&

    \begin{minipage}{0.166\columnwidth}
    \begin{center}
    \includegraphics[width=1.0\columnwidth,keepaspectratio=true]
    {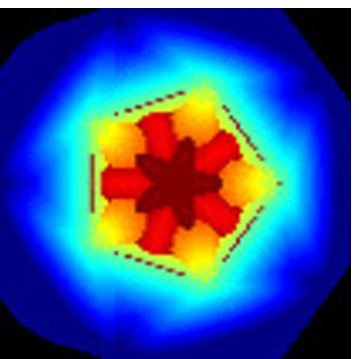}
    \end{center}
    \end{minipage}&

    \begin{minipage}{0.166\columnwidth}
    \begin{center}
    \includegraphics[width=1.0\columnwidth,keepaspectratio=true]
    {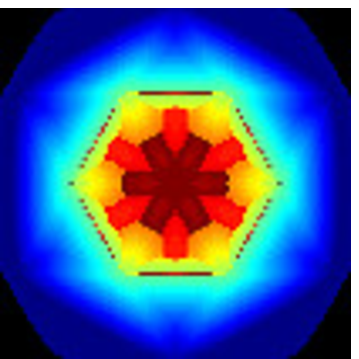}
    \end{center}
    \end{minipage}&

    \begin{minipage}{0.166\columnwidth}
    \begin{center}
    \includegraphics[width=1.0\columnwidth,keepaspectratio=true]
    {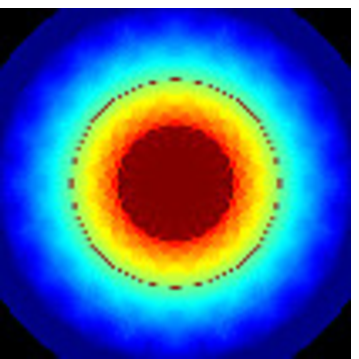}
    \end{center}
    \end{minipage}&

    \begin{minipage}{0.166\columnwidth}
    \begin{center}
    \includegraphics[width=1.0\columnwidth,keepaspectratio=true]
    {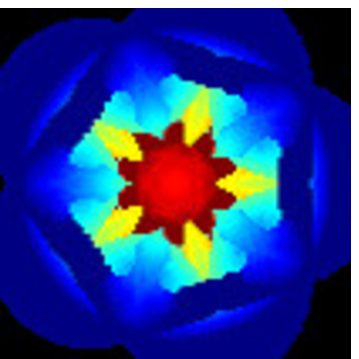}
    \end{center}
    \end{minipage}
  \end{tabular}
\end{center}
\caption{Torque value maps (upper row) and scale maps  (lower row)  for simple shapes.
The shapes are overlaid  onto the torque value with black lines.
Black regions in the scale map denote areas where  the torque value is the same over scales.
\label{fig:Torque Value Maps and Scale Maps for Simple Shapes}}
\end{figure}

Figure~\ref{fig:Torque Value Maps and Scale Maps for Simple Shapes} shows examples of torque value maps and scale maps for some simple figures. As can be seen the torque operator captures well the concept of closure. For these simple shapes, the torque value map directly provides the location of the object, i.e. the center of the polygon, and the scale map provides its scale (at the point corresponding to the extrema in the value map).

\begin{figure}[tb]
\begin{center}
  \begin{tabular}[htbp]{@{}c@{}c@{}c@{\hspace{0.03\columnwidth}}c@{}c@{}c@{}}
    \begin{minipage}{0.16\columnwidth}
    \begin{center}
    \includegraphics[width=1.0\columnwidth,keepaspectratio=true]
    {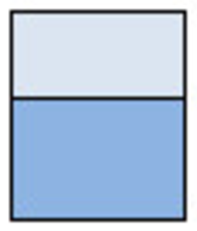}
    \end{center}
    \end{minipage}&

    \begin{minipage}{0.16\columnwidth}
    \begin{center}
    \includegraphics[width=1.0\columnwidth,keepaspectratio=true]
    {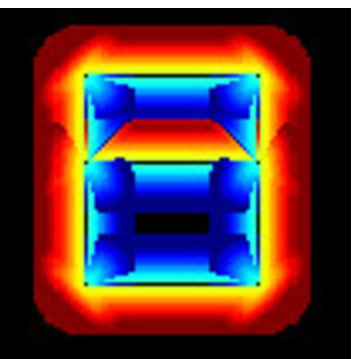}
    \end{center}
    \end{minipage}&

    \begin{minipage}{0.16\columnwidth}
    \begin{center}
    \includegraphics[width=1.0\columnwidth,keepaspectratio=true]
    {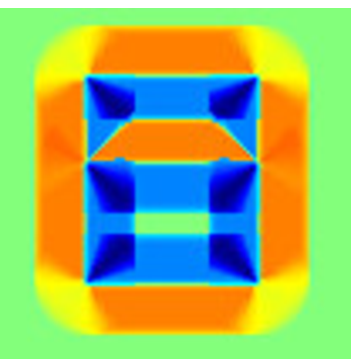}
    \end{center}
    \end{minipage}&

    \begin{minipage}{0.16\columnwidth}
    \begin{center}
    \includegraphics[width=1.0\columnwidth,keepaspectratio=true]
    {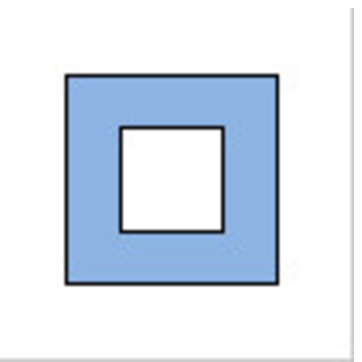}
    \end{center}
    \end{minipage}&

    \begin{minipage}{0.16\columnwidth}
    \begin{center}
    \includegraphics[width=1.0\columnwidth,keepaspectratio=true]
    {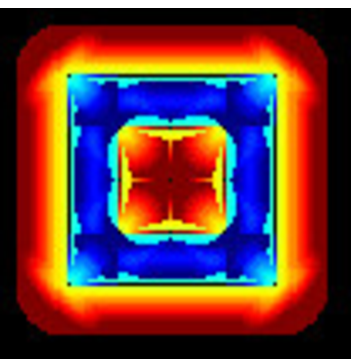}
    \end{center}
    \end{minipage}&

    \begin{minipage}{0.16\columnwidth}
    \begin{center}
    \includegraphics[width=1.0\columnwidth,keepaspectratio=true]
    {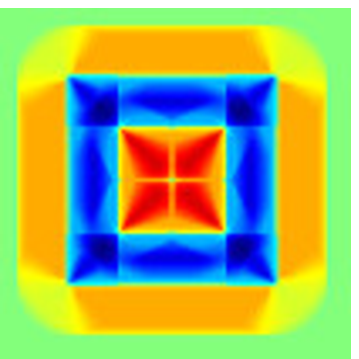}
    \end{center}
    \end{minipage}\\

    \begin{minipage}{0.16\columnwidth}
    \begin{center}
    \includegraphics[width=1.0\columnwidth,keepaspectratio=true]
    {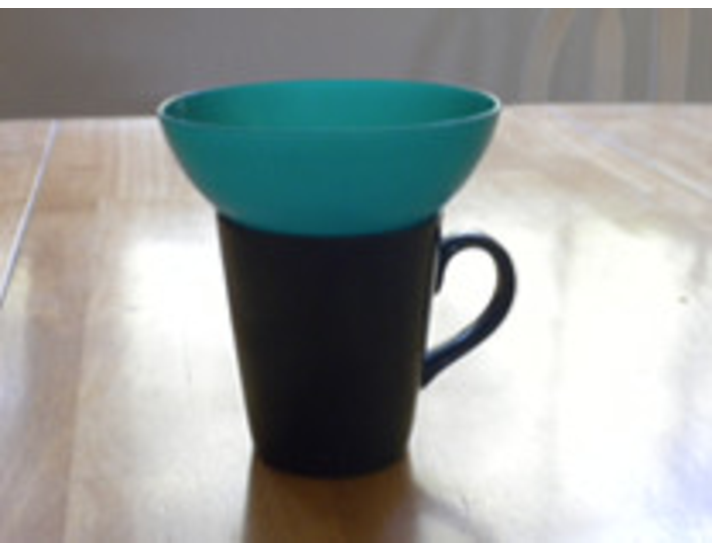}
    \end{center}
    \end{minipage}&

    \begin{minipage}{0.16\columnwidth}
    \begin{center}
    \includegraphics[width=1.0\columnwidth,keepaspectratio=true]
    {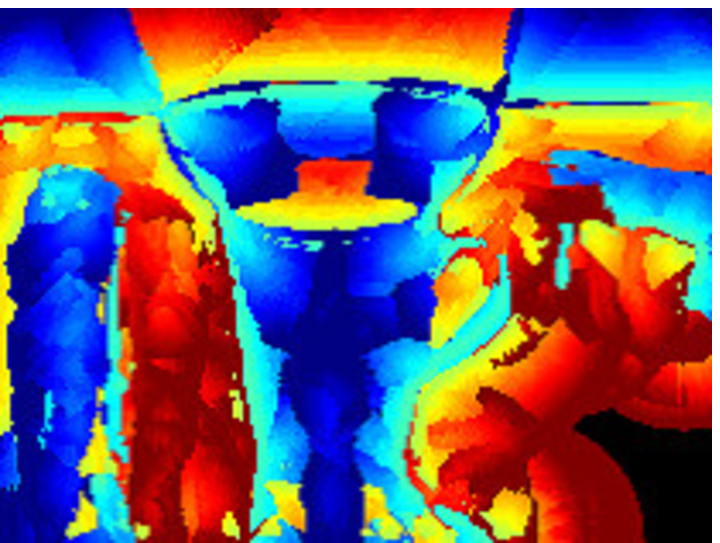}
    \end{center}
    \end{minipage}&

    \begin{minipage}{0.16\columnwidth}
    \begin{center}
    \includegraphics[width=1.0\columnwidth,keepaspectratio=true]
    {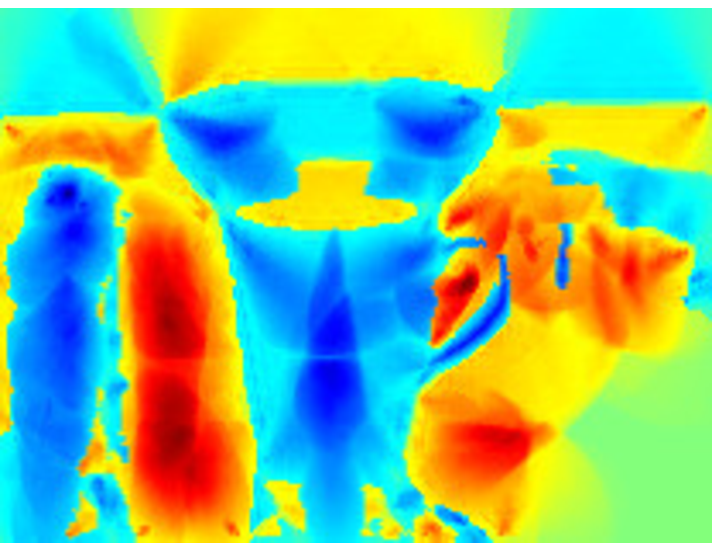}
    \end{center}
    \end{minipage}&

    \begin{minipage}{0.16\columnwidth}
    \begin{center}
    \includegraphics[width=1.0\columnwidth,keepaspectratio=true]
    {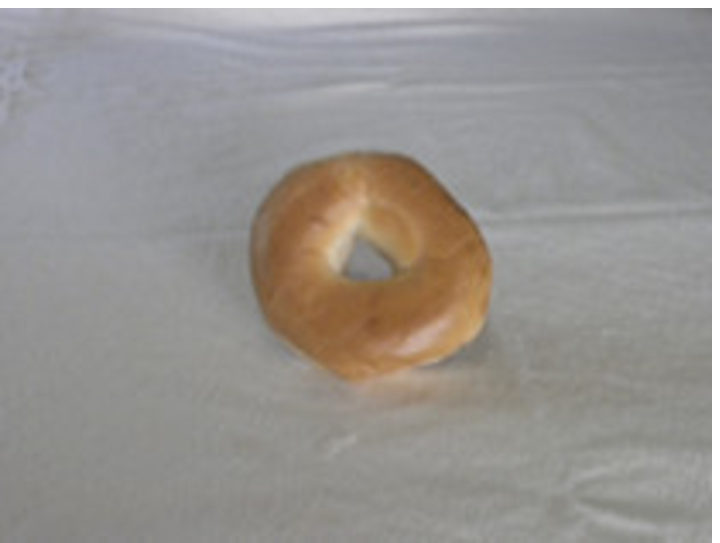}
    \end{center}
    \end{minipage}&

    \begin{minipage}{0.16\columnwidth}
    \begin{center}
    \includegraphics[width=1.0\columnwidth,keepaspectratio=true]
    {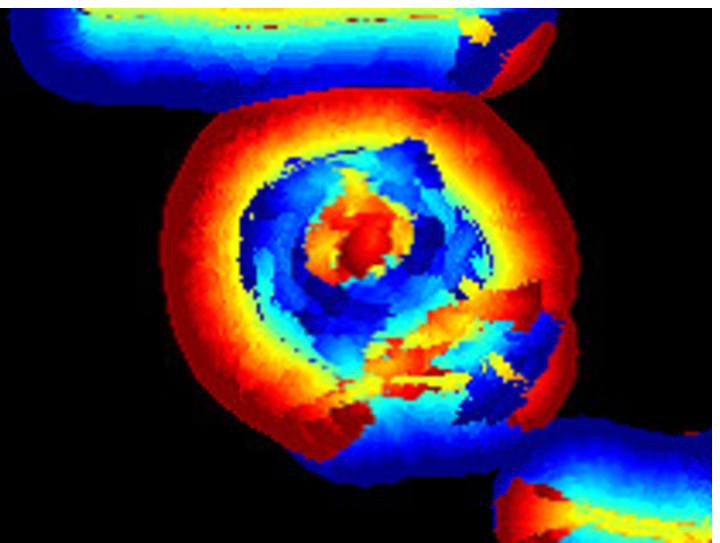}
    \end{center}
    \end{minipage}&

    \begin{minipage}{0.16\columnwidth}
    \begin{center}
    \includegraphics[width=1.0\columnwidth,keepaspectratio=true]
    {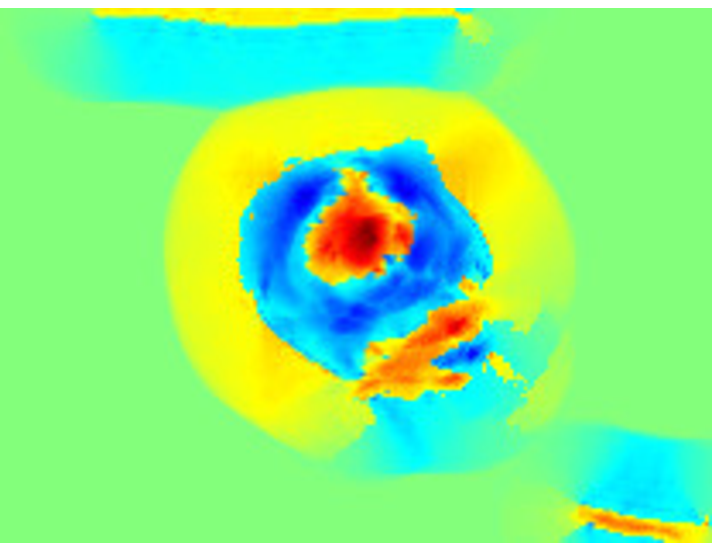}
    \end{center}
    \end{minipage}
  \end{tabular}
\end{center}
\caption{Torque for two configurations with multiple objects.
The first row shows graphical examples of the configuration, and the second row shows real images with objects of such configuration.
Columns (1 and 4), (2 and 5) and (3 and 6) show the test images, the  scale maps and  torque value maps, respectively.
\label{fig:Torque Value Maps and Scale Maps for Multiple Objects}}
\end{figure}

The torque  maps are more complex for multiple objects located close to each other, or for objects of different topology. Figure~\ref{fig:Torque Value Maps and Scale Maps for Multiple Objects}  illustrates the theoretical and empirical maps for two situations: the case of an object on top of another object, and the case  an object with a hole. The former is illustrated for two dark objects on  bright background with the object  at the bottom  darker than the one on top. The object regions have  negative torque value, but positive torque  values appear on the brighter object around the edges shared with the darker object. This is because  the edge orientation is defined by  polarity. In this region the  edge boundary abuts a darker region, which is in conflict with the other edges which abut brighter regions. Thus, these edges give
torque values of opposite sign, and result in a  positive torque value region close
to the edge. For the case where the object has a hole, we do not have such a confusion.

\begin{figure}[tb]
  \begin{center}
    \begin{tabular}{@{}c@{}c@{}c@{}c@{}c@{}}

      \begin{minipage}{0.2\columnwidth}
      \begin{center}
      \includegraphics[width=1\columnwidth,keepaspectratio=true,clip]
      {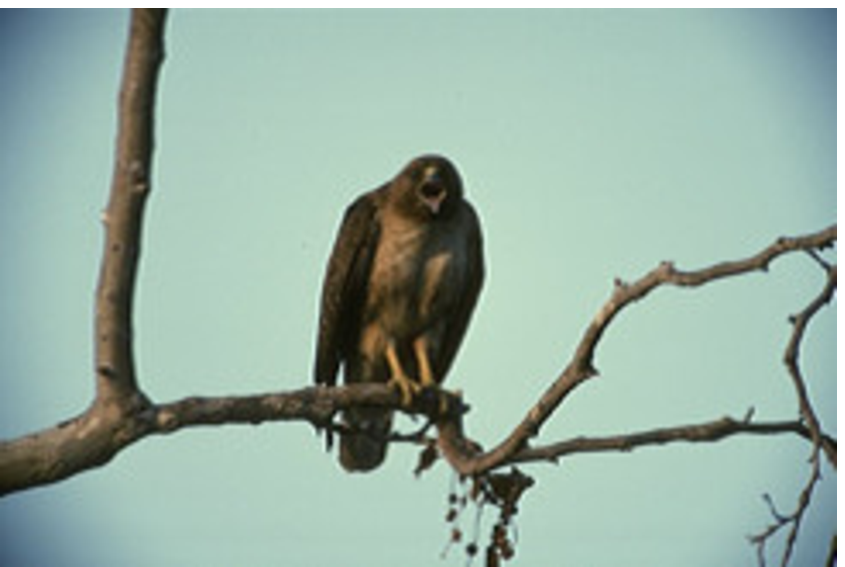}
      \end{center}
      \end{minipage}&

      \begin{minipage}{0.2\columnwidth}
      \begin{center}
      \includegraphics[width=1\columnwidth,keepaspectratio=true,clip]
      {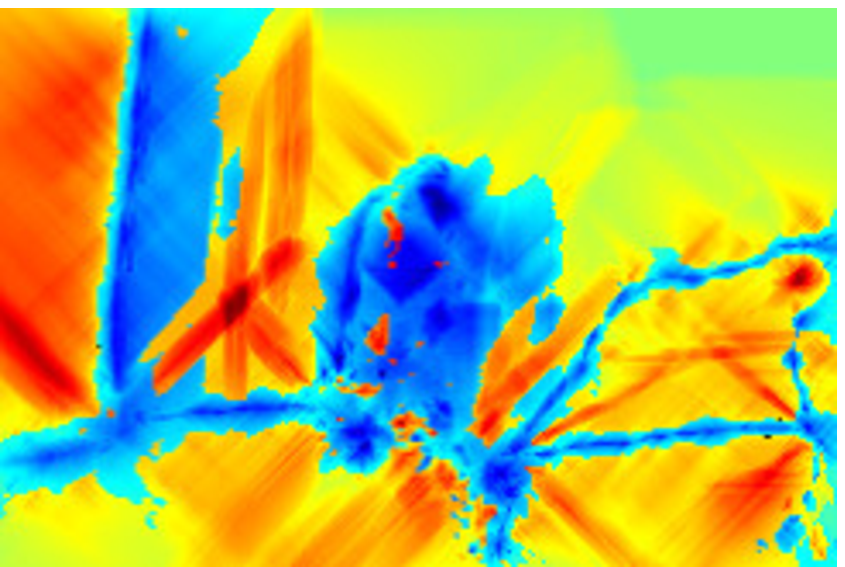}
      \end{center}
      \end{minipage}&

      \begin{minipage}{0.2\columnwidth}
      \begin{center}
      \includegraphics[width=1\columnwidth,keepaspectratio=true,clip]
      {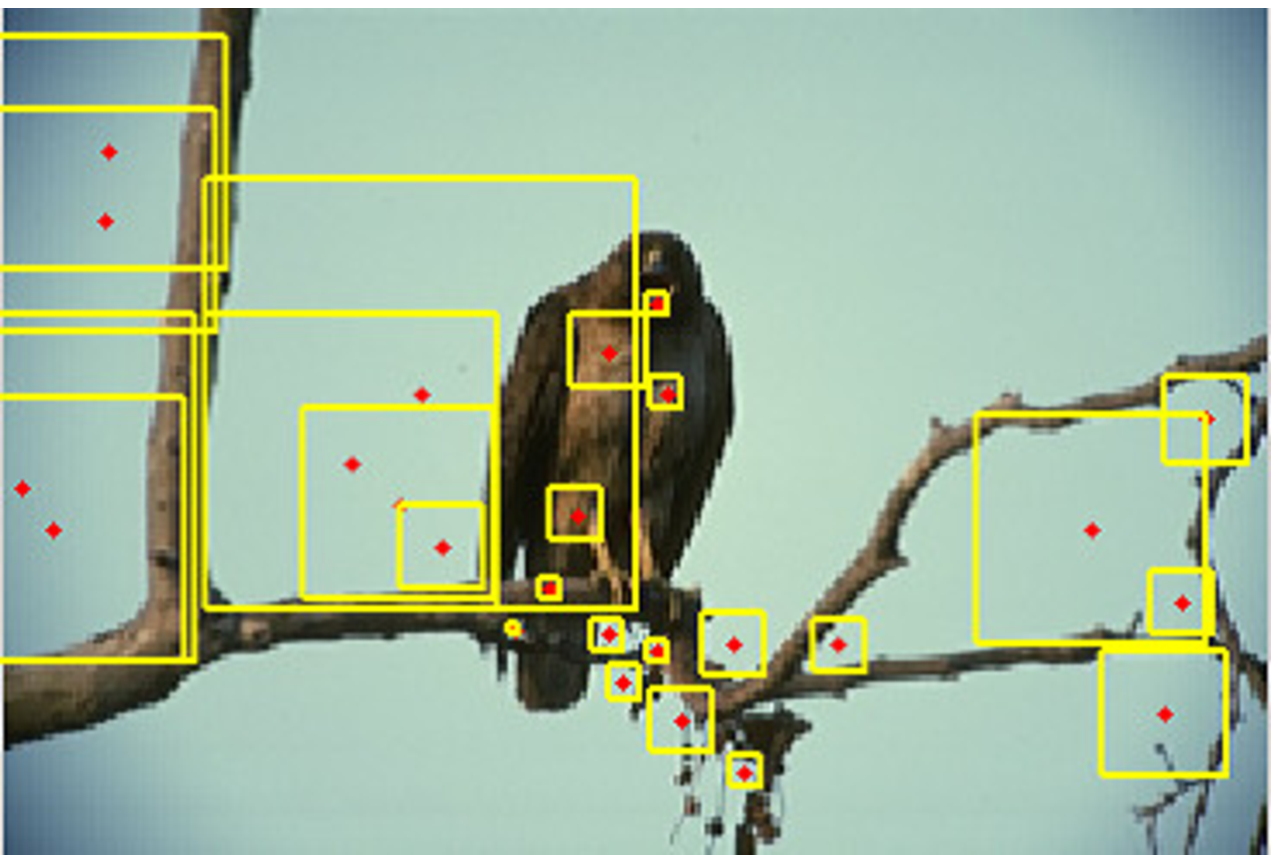}
      \end{center}
      \end{minipage}&

      \begin{minipage}{0.2\columnwidth}
      \begin{center}
      \includegraphics[width=1\columnwidth,keepaspectratio=true,clip]
      {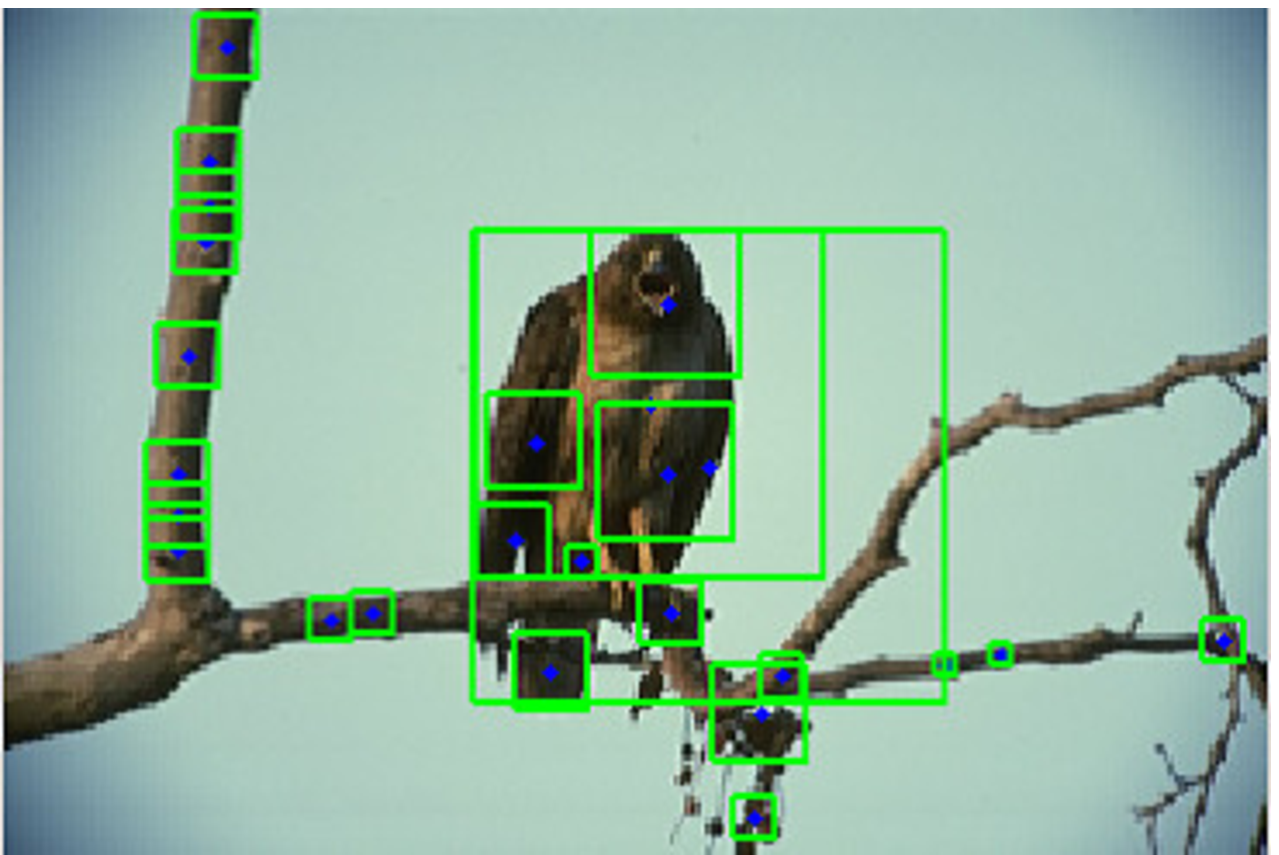}
      \end{center}
      \end{minipage}&

      \begin{minipage}{0.2\columnwidth}
      \begin{center}
      \includegraphics[width=1\columnwidth,keepaspectratio=true,clip]
      {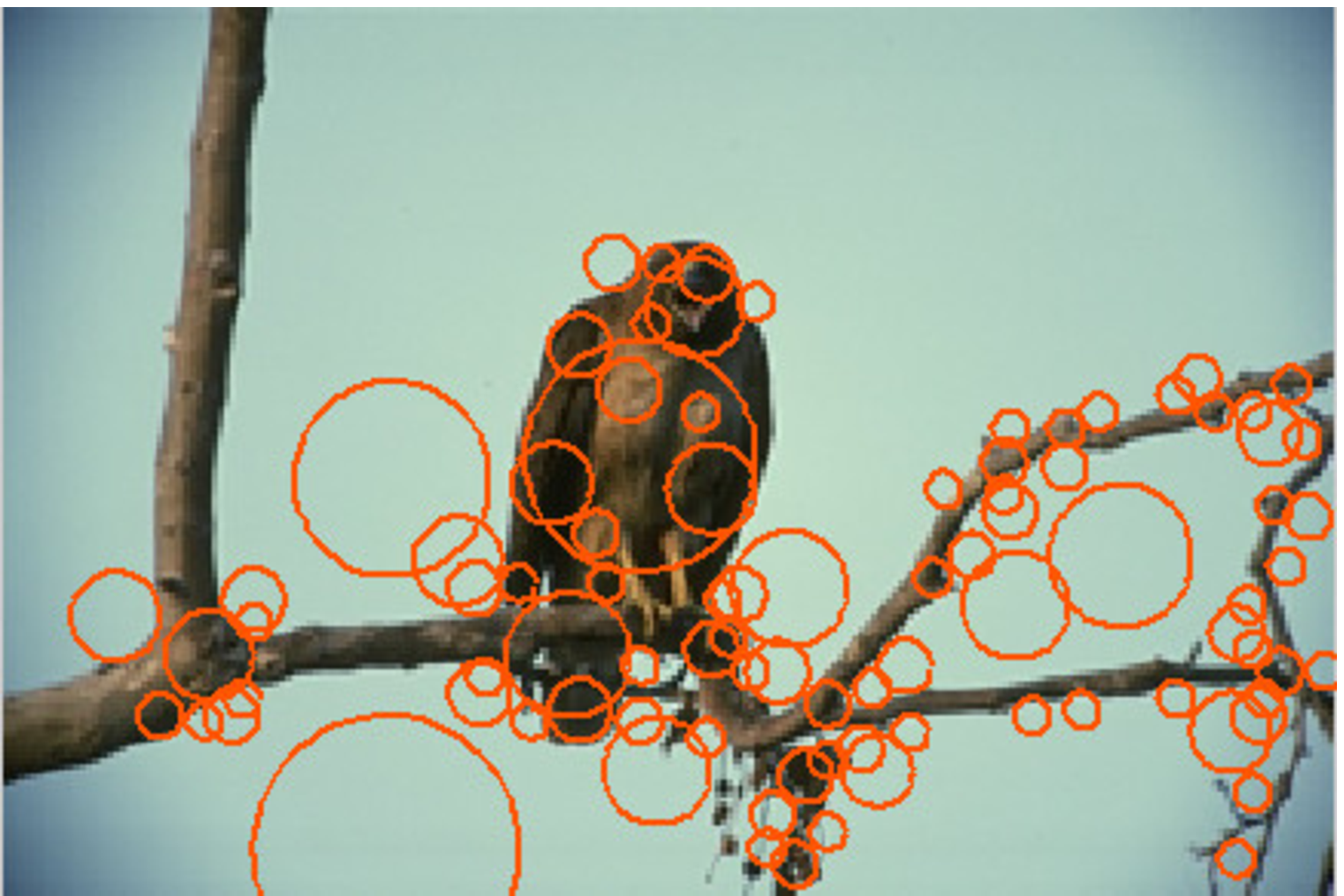}
      \end{center}
      \end{minipage}\\

      \begin{minipage}{0.2\columnwidth}
      \begin{center}
      \includegraphics[width=1\columnwidth,keepaspectratio=true,clip]
      {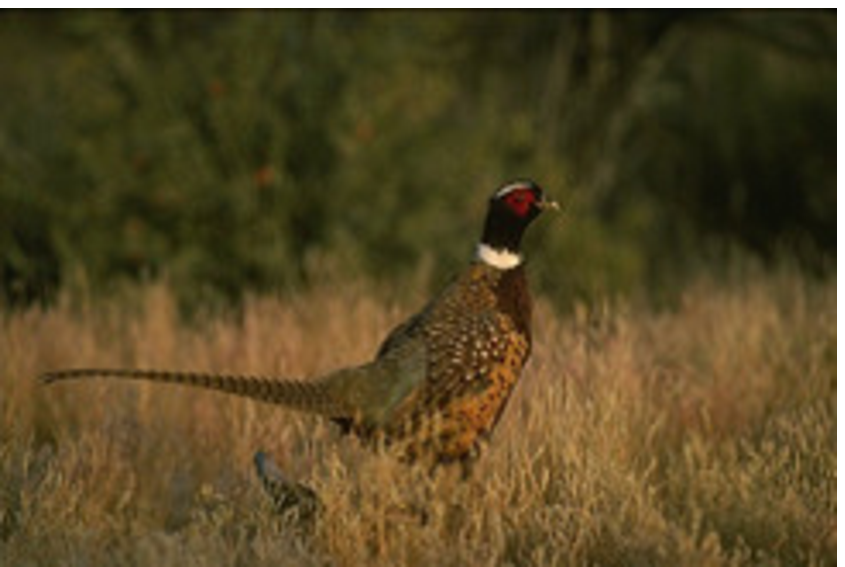}
      \end{center}
      \end{minipage}&

      \begin{minipage}{0.2\columnwidth}
      \begin{center}
      \includegraphics[width=1\columnwidth,keepaspectratio=true,clip]
      {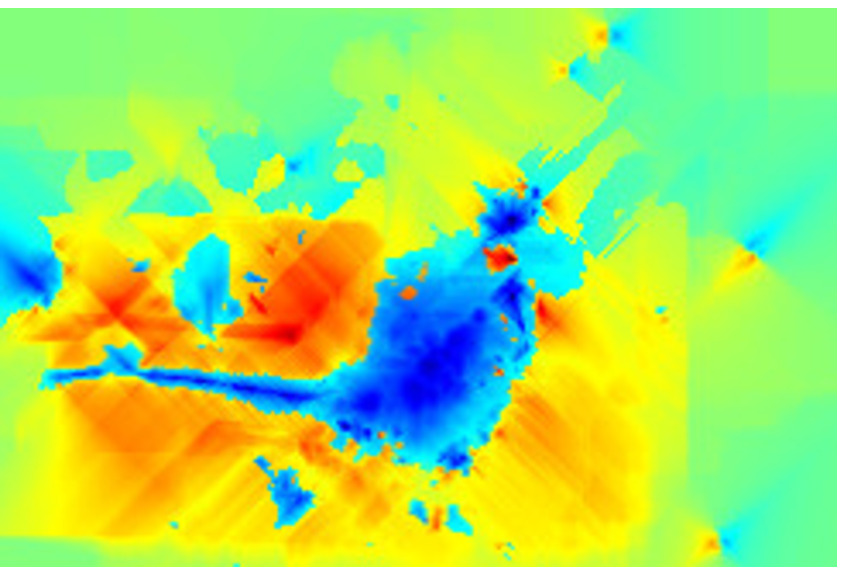}
      \end{center}
      \end{minipage}&

      \begin{minipage}{0.2\columnwidth}
      \begin{center}
      \includegraphics[width=1\columnwidth,keepaspectratio=true,clip]
      {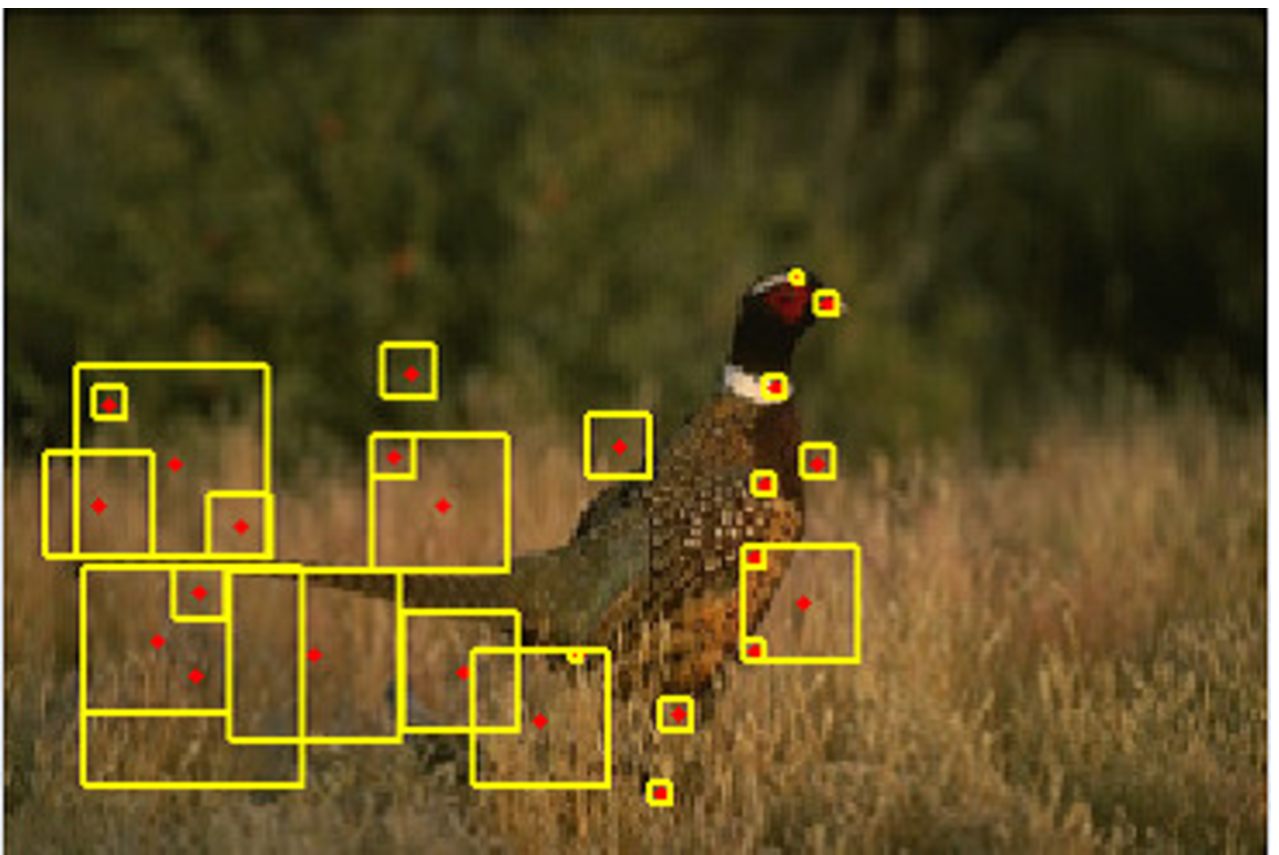}
      \end{center}
      \end{minipage}&

      \begin{minipage}{0.2\columnwidth}
      \begin{center}
      \includegraphics[width=1\columnwidth,keepaspectratio=true,clip]
      {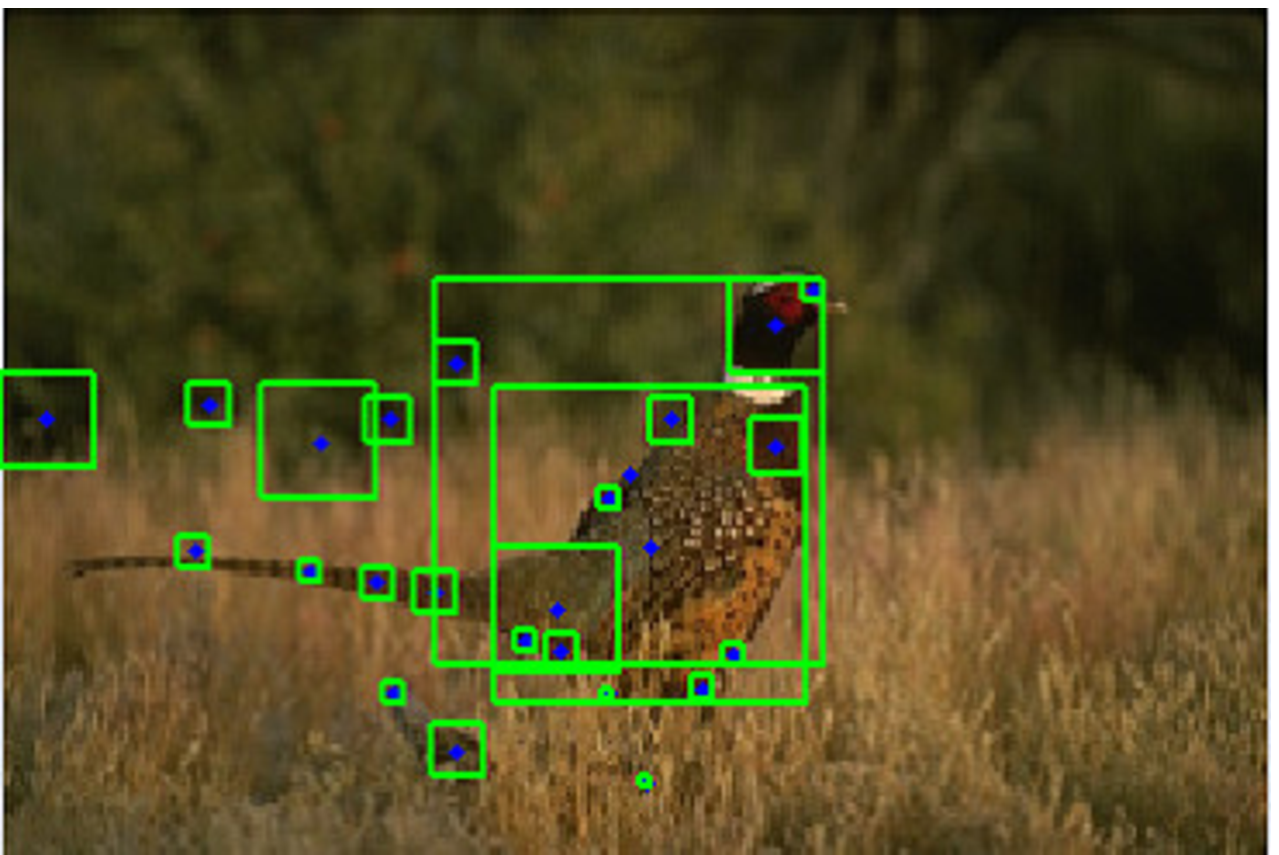}
      \end{center}
      \end{minipage}&

      \begin{minipage}{0.2\columnwidth}
      \begin{center}
      \includegraphics[width=1\columnwidth,keepaspectratio=true,clip]
      {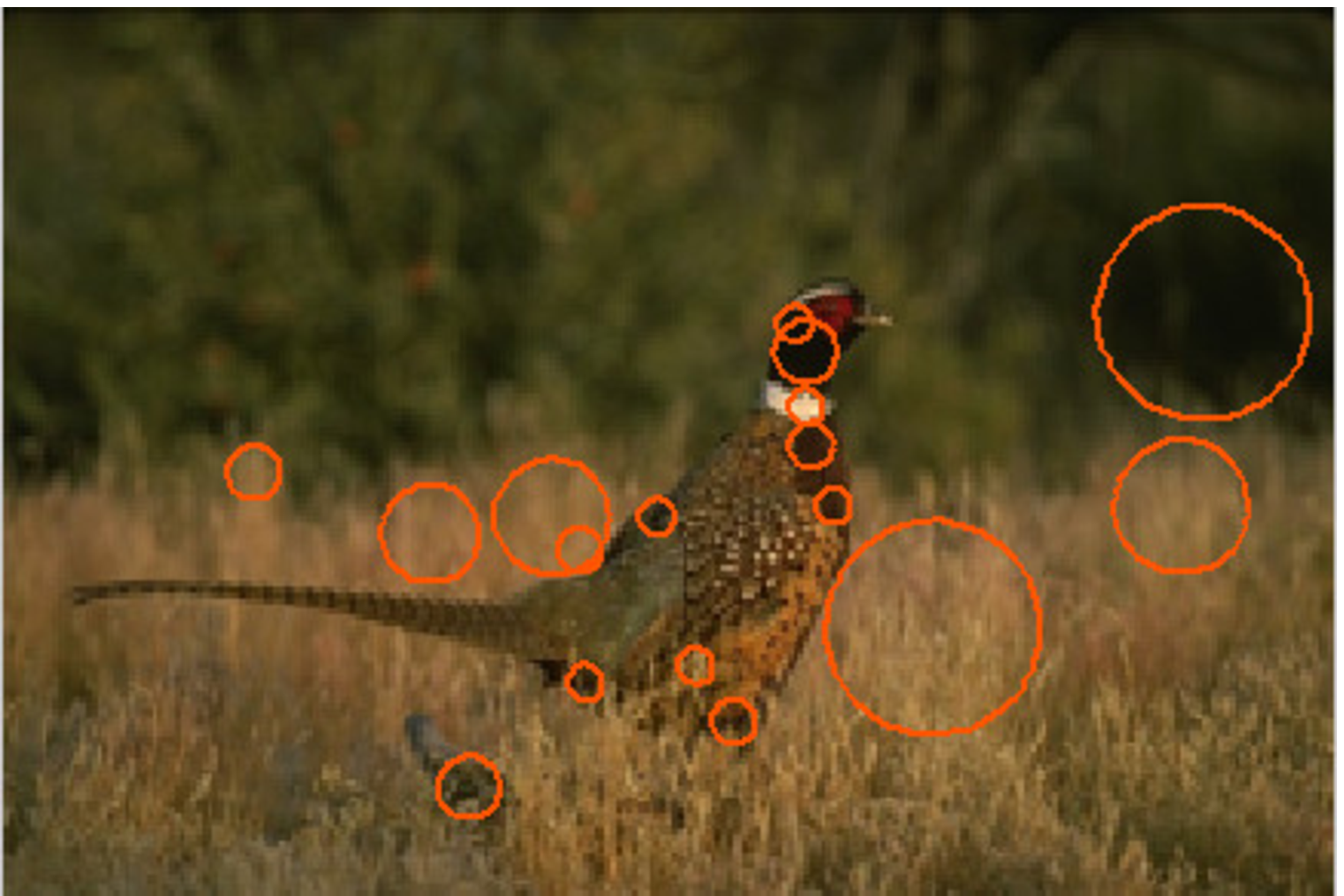}
      \end{center}
      \end{minipage}\\

      \begin{minipage}{0.2\columnwidth}
      \begin{center}
      \includegraphics[width=1\columnwidth,keepaspectratio=true,clip]
      {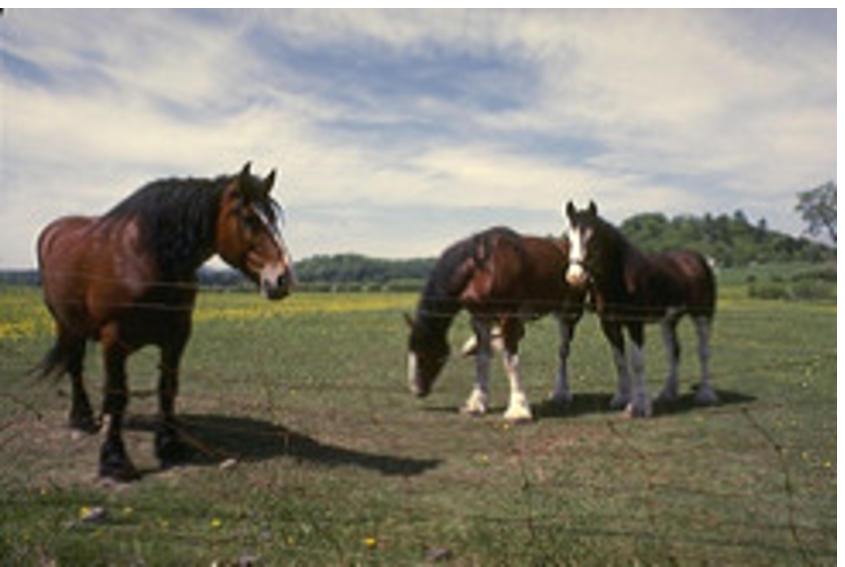}
      \end{center}
      \end{minipage}&

      \begin{minipage}{0.2\columnwidth}
      \begin{center}
      \includegraphics[width=1\columnwidth,keepaspectratio=true,clip]
      {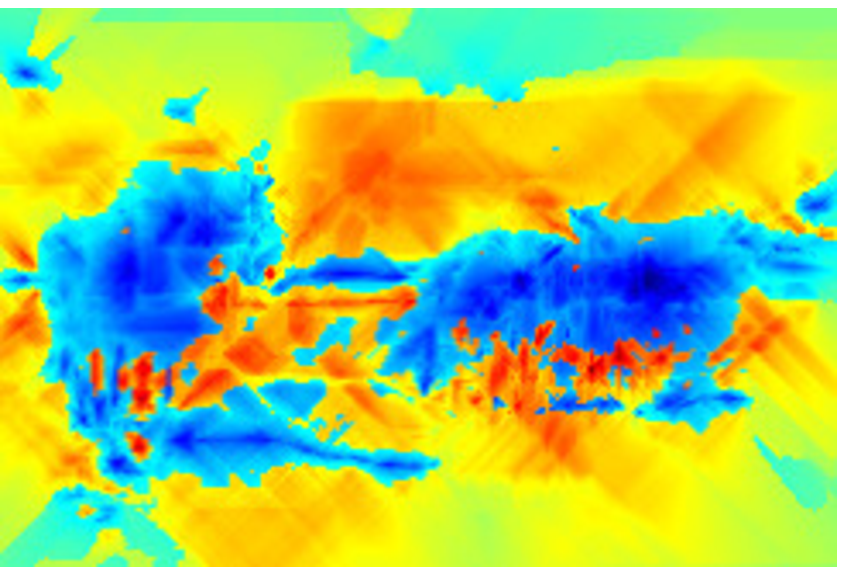}
      \end{center}
      \end{minipage}&

      \begin{minipage}{0.2\columnwidth}
      \begin{center}
      \includegraphics[width=1\columnwidth,keepaspectratio=true,clip]
      {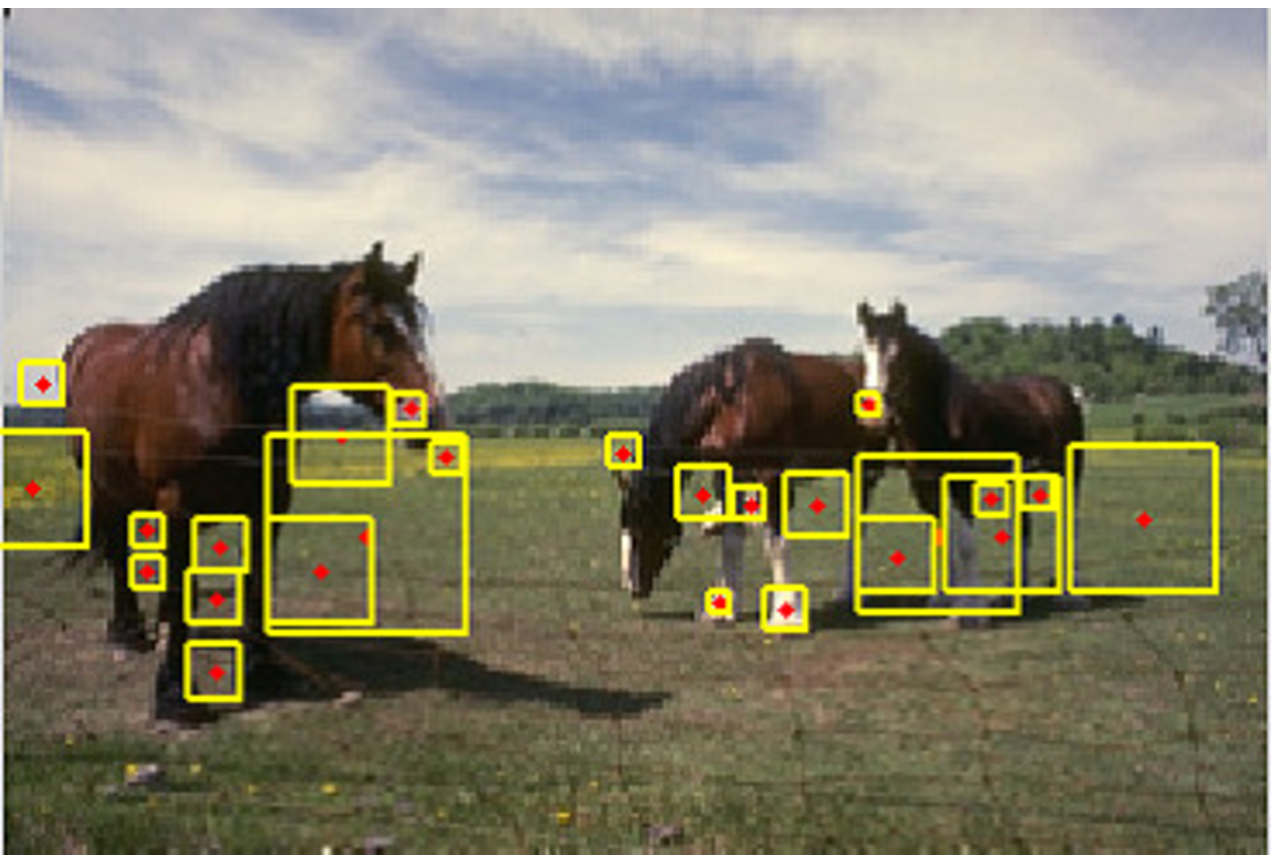}
      \end{center}
      \end{minipage}&

      \begin{minipage}{0.2\columnwidth}
      \begin{center}
      \includegraphics[width=1\columnwidth,keepaspectratio=true,clip]
      {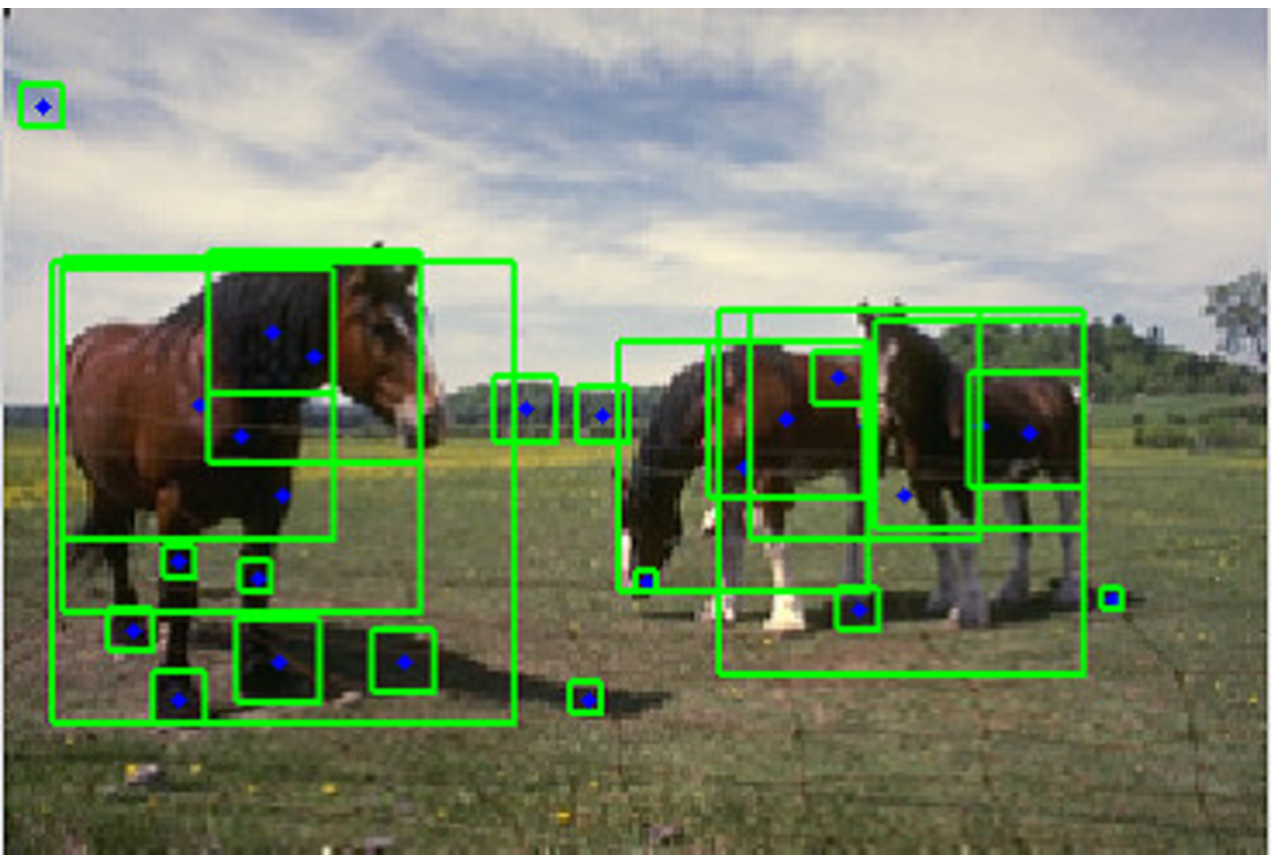}
      \end{center}
      \end{minipage}&

      \begin{minipage}{0.2\columnwidth}
      \begin{center}
      \includegraphics[width=1\columnwidth,keepaspectratio=true,clip]
      {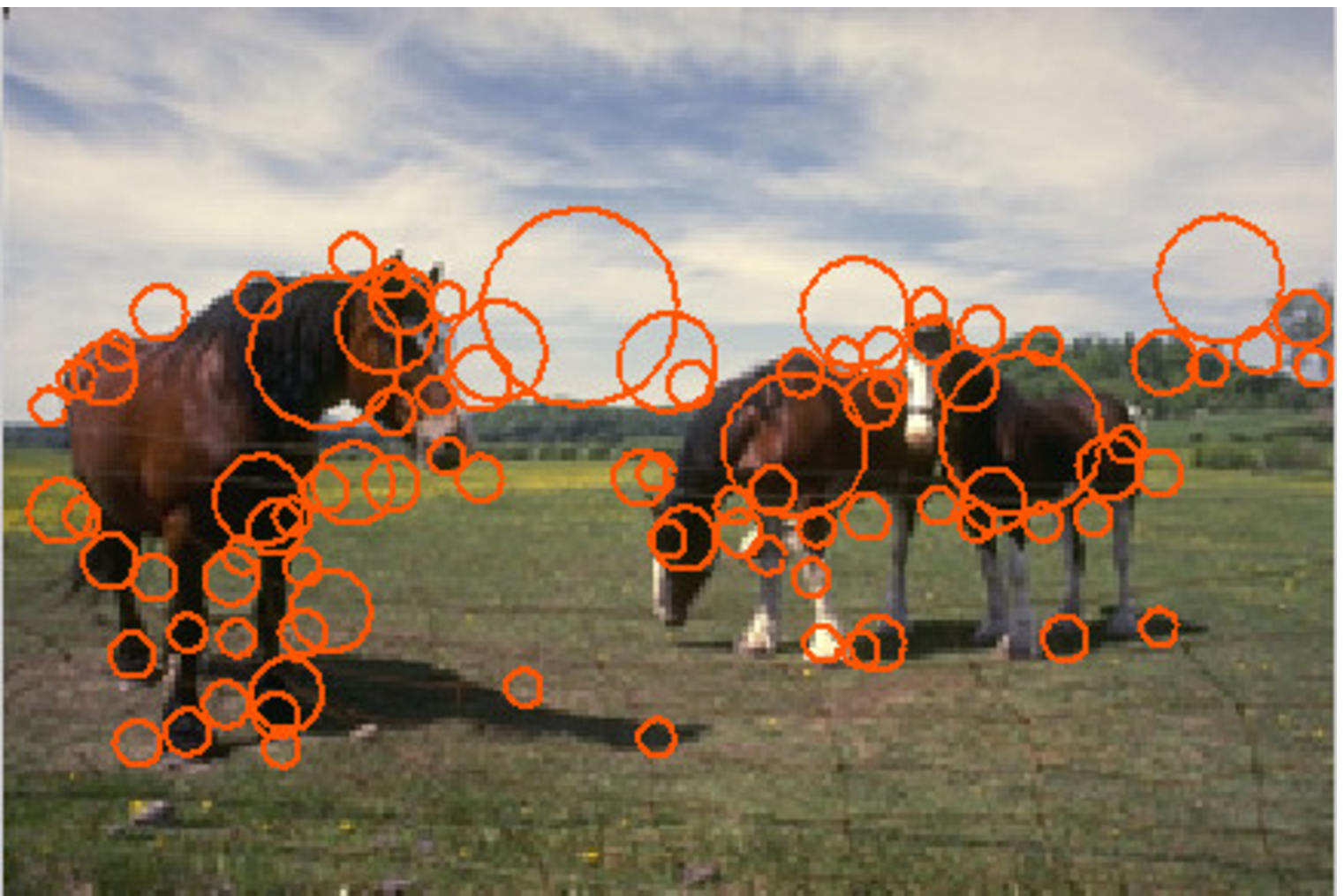}
      \end{center}
      \end{minipage}\\

      \begin{minipage}{0.2\columnwidth}
      \begin{center}
      \includegraphics[width=1\columnwidth,keepaspectratio=true,clip]
      {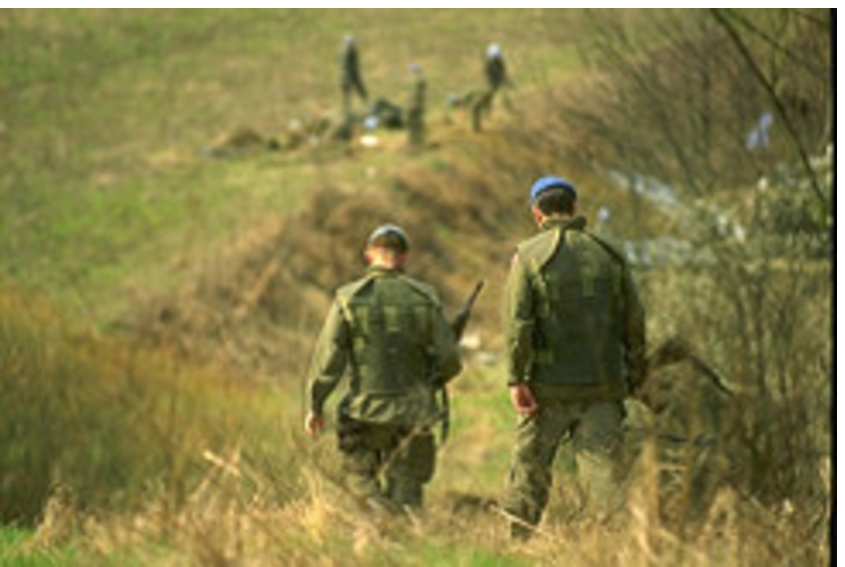}
      \end{center}
      \end{minipage}&

      \begin{minipage}{0.2\columnwidth}
      \begin{center}
      \includegraphics[width=1\columnwidth,keepaspectratio=true,clip]
      {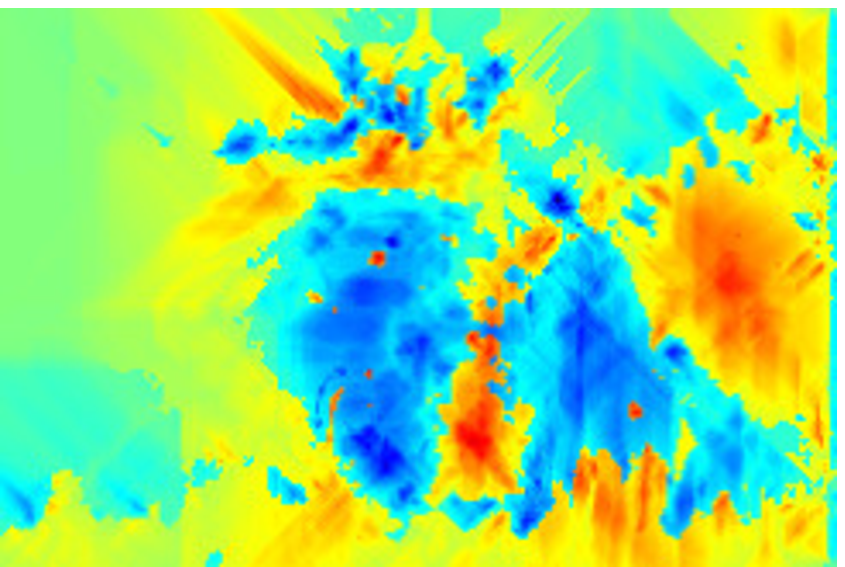}
      \end{center}
      \end{minipage}&

      \begin{minipage}{0.2\columnwidth}
      \begin{center}
      \includegraphics[width=1\columnwidth,keepaspectratio=true,clip]
      {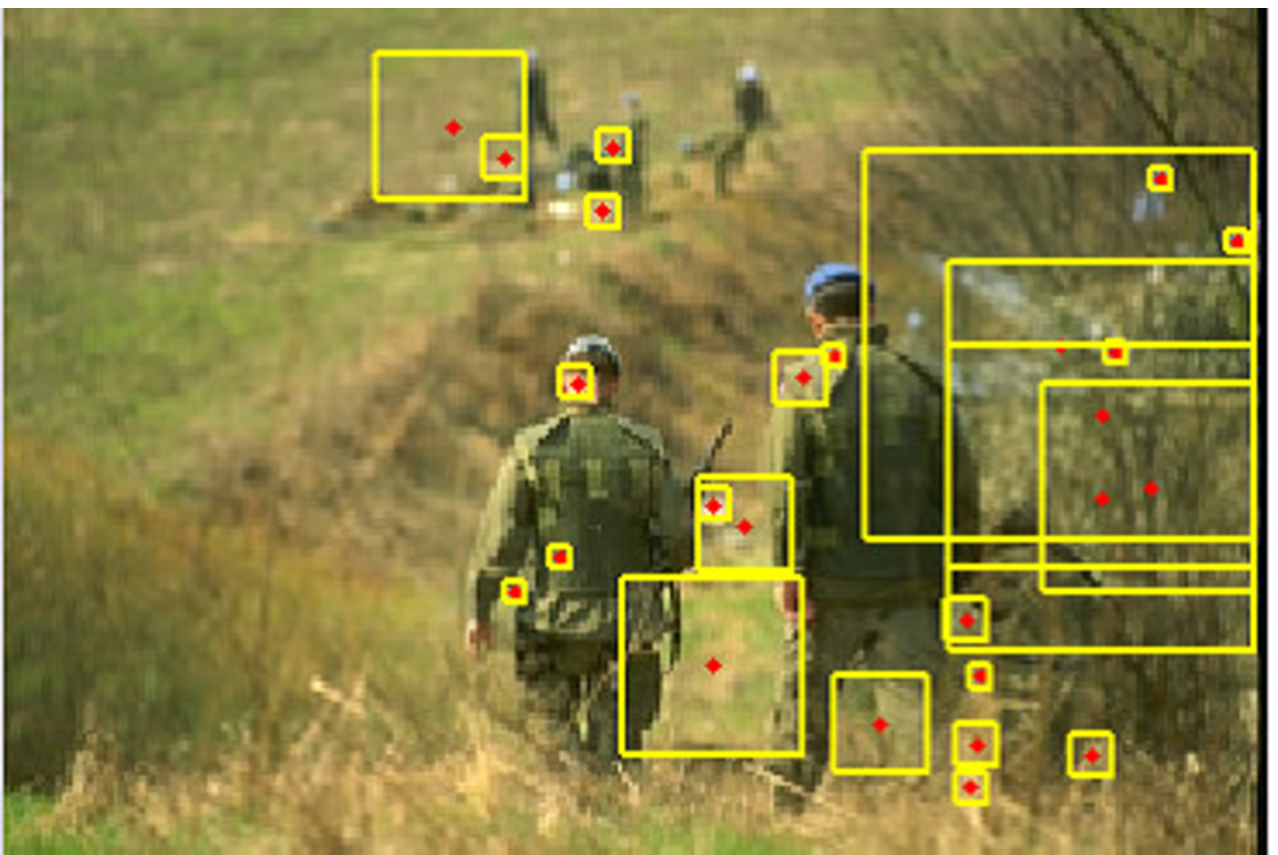}
      \end{center}
      \end{minipage}&

      \begin{minipage}{0.2\columnwidth}
      \begin{center}
      \includegraphics[width=1\columnwidth,keepaspectratio=true,clip]
      {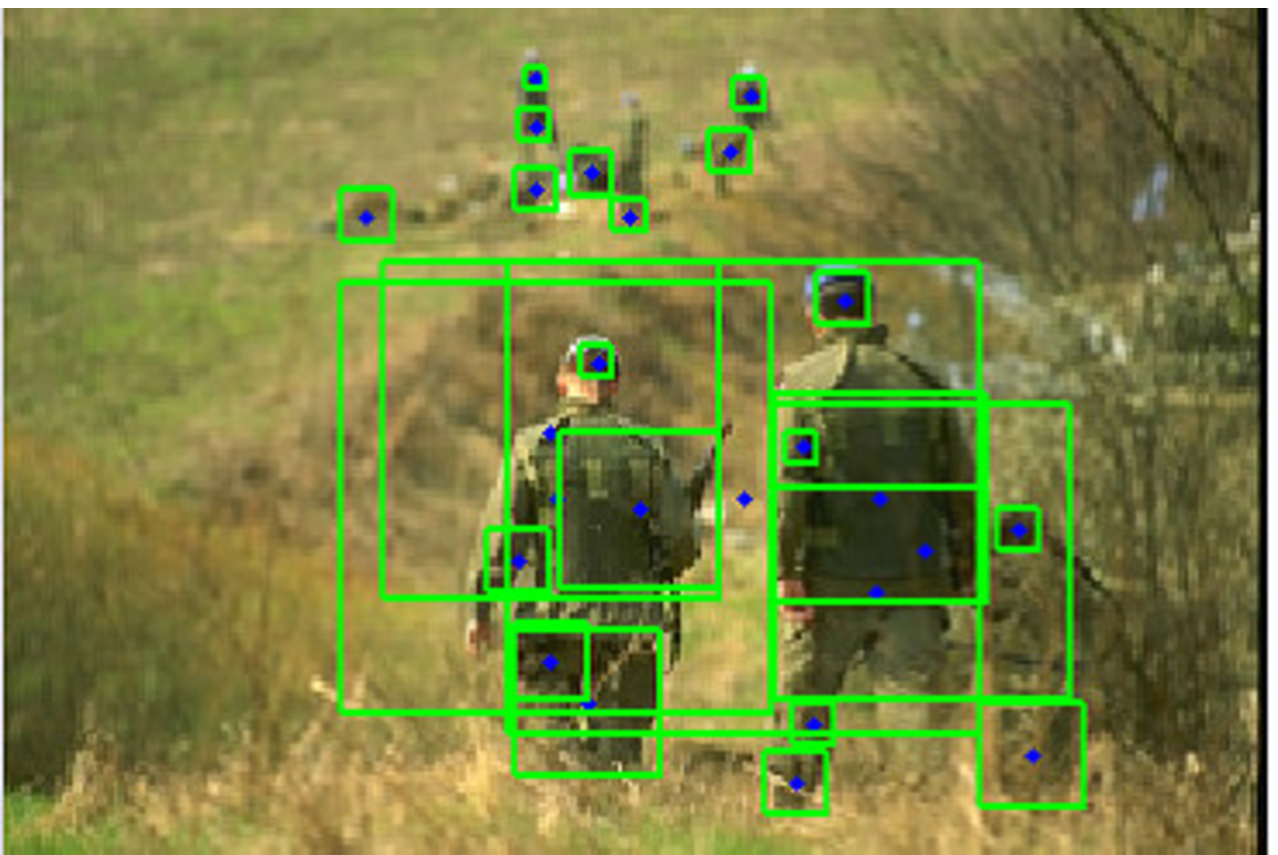}
      \end{center}
      \end{minipage}&

      \begin{minipage}{0.2\columnwidth}
      \begin{center}
      \includegraphics[width=1\columnwidth,keepaspectratio=true,clip]
      {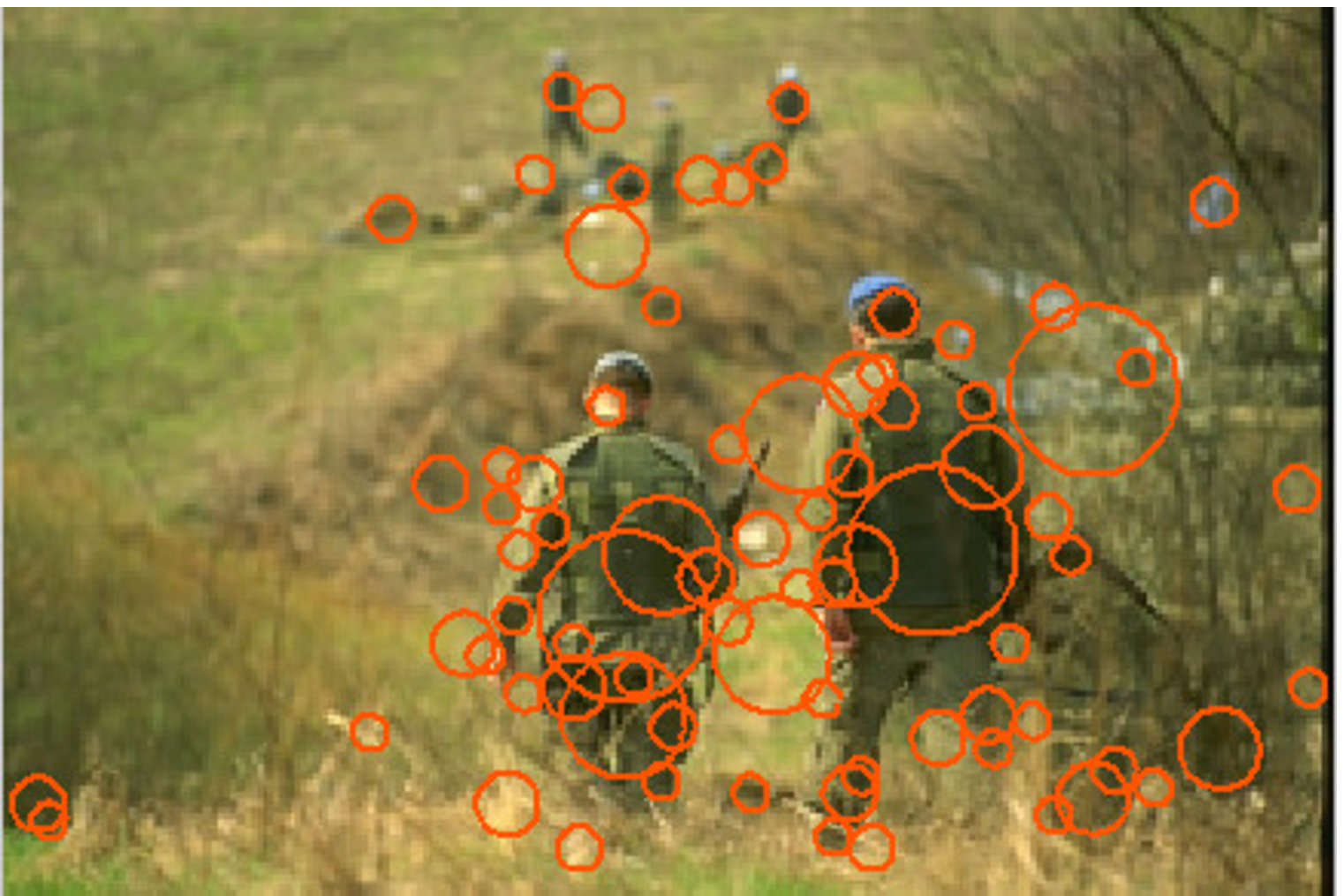}
      \end{center}
      \end{minipage}\\

      (a) & (b) & (c) & (d) & (e)

    \end{tabular}
  \end{center}
\caption{Torque value maps and extrema in torque.
(a) Original test images. 
(b) Torque value maps.
(c) and (d) Local maxima and local minima in torque volume, respectively. Dots indicate the location of extrema in image space, and the corresponding squares indicate  the size of the patch producing the extrema.
(e) Blob detection \cite{Lindeberg1998,Kokkinos2006} as reference.
\label{fig:Torque Value Maps and Extrema in Torque}}
\end{figure}

Figure~\ref{fig:Torque Value Maps and Extrema in Torque} illustrates the application of the torque operator on  real images from the Berkeley dataset \cite{Martin2001}.
Here square patches were used with their sizes  varying from 3 to 91 pixels, and the images were downsized by a factor of two  to $161 \times 241$. Thus the largest image patch covers about 21\% of the  image area. Figures~\ref{fig:Torque Value Maps and Extrema in Torque} (a)- (d) show the test images, torque value maps, and location of the extrema and corresponding patch sizes (as green squares for minima and yellow squares for maxima).
For each test image, the 25 maxima and 25 minima of largest absolute torque values are shown (in \ref{fig:Torque Value Maps and Extrema in Torque} (c) and (d), respectively).
As can be seen from column (b), the negative torque regions match well object regions for these images. We should note  that the sign of the torque  depends on the relative brightness of an object to the background, and in all these examples the objects are darker than their surroundings (as reflected in the torque  minima in column (d)).
An interesting property of the torque operator is that the extrema are  not located simply at the dense edges, but at locations surrounded by edges.
For example, in  the image of the pheasant, the inner part of the pheasant doesn't have clear edges. Nevertheless, local minima are found in the torque volume, and  the location of structured, surrounding boundaries in these parts are detected.
A second interesting property of the torque operator is that the extrema indicate the size of structured edges.
Referring to the figure,
it can be seen that the patches associated with local minima (shown in green) cover most of the object regions.
These properties of indicating roughly the location and size of structured edges are useful for further image processes, such as visual attention and object segmentation.

Fig.~\ref{fig:Torque Value Maps and Extrema in Torque}(e) provides a comparison to the blob detection of \cite{Lindeberg1998} using the implementation in \cite{Kokkinos2006}. The blob detector locates the points in scale space where the square of a normalized Laplacian assumes maxima with respect to space and scale. The extrema are thus contrast invariant (i.e, both dark blobs on light background and light blobs on dark background). The torque extrema also amount to local extrema with respect to space and scale, but we separate them according to contrast.
Comparing visually the  blob detection to the  torque minima, we can see that  the latter tends to have extrema representing  whole objects. We can see that the regions of the  eagle, the pheasant, the horses, and the persons are detected by negative torque extrema in the test images. This indicates  that the torque detector has greater flexibility to shape variation.  The blob detector is maximally tuned to annular-like image patches (like the Laplacian kernel). The torque detector, if implemented with square or disk windows, also is maximally tuned to circular structures, but it also responds  to  other shapes, especially closed convex shapes, as it adds up all the edgels in the window. Furthermore, we see the torque mechanism as  a more general concept, and it can be extended in various ways. For example, we can use other window shapes, as in section \ref{sec:rec}, or  we can also tune it to favor  certain shapes.
Furthermore, besides the extrema, we can also utilize the  torque values and scales, as in Section \ref{sec:rec}.

\subsection{Scale Selection}
\label{sec:Scale Selection}

In the previous section we proposed a two-dimensional representation of the torque volume as torque value map and scale map. The  torque value map and scale map are generated from the torque volume by selecting at each pixel the scale corresponding to the largest absolute torque value over scales.
Scale selection in general is an important topic in computer vision \cite{Lindeberg1994,Lindeberg1998}, and selecting the appropriate scale can lead to  better performance  in most image processing tasks \cite{Galun2007,XuY2012}.

Our main focus in this paper is detection of object-like structures, and thus for most images we want  to  avoid attention to smaller structures.
One strategy to handle this within the torque framework, is to first reduce the noise in the torque volume due to edge fragments and then select the scale.
Standard techniques for noise reduction such as filtering and optimization can be applied to the torque volume.

\begin{figure}[htbp]
  \begin{center}
    \begin{tabular}{@{}c@{}c@{}c@{}c@{}}

      Test Image & $\alpha=1.0$ & $\alpha=1.5$ & $\alpha=2.0$\\

	  \begin{minipage}{0.25\columnwidth}
      \begin{center}
      \includegraphics[width=1\columnwidth,keepaspectratio=true,clip]
      {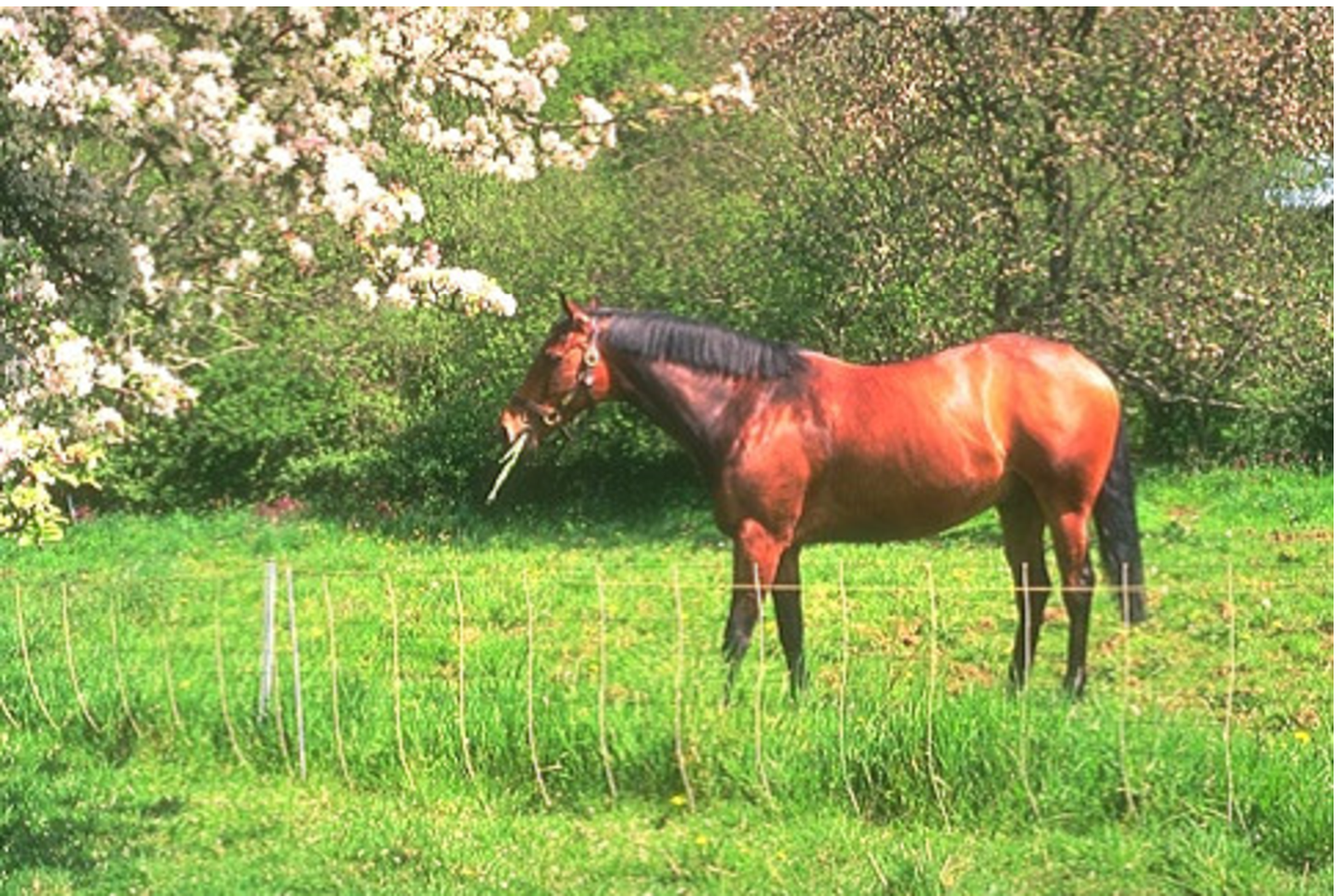}
      \end{center}
      \end{minipage}&

	  \begin{minipage}{0.25\columnwidth}
      \begin{center}
      \includegraphics[width=1\columnwidth,keepaspectratio=true,clip]
      {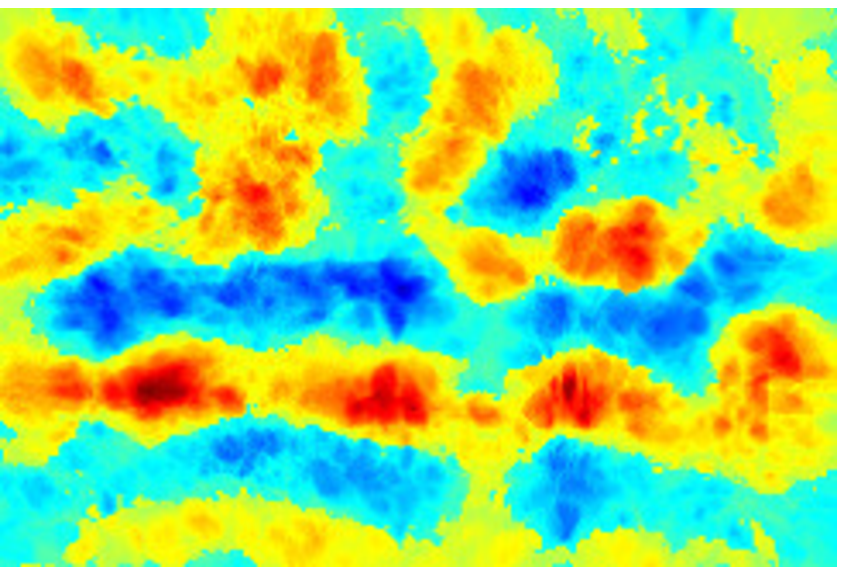}
      \end{center}
      \end{minipage}&

	  \begin{minipage}{0.25\columnwidth}
      \begin{center}
      \includegraphics[width=1\columnwidth,keepaspectratio=true,clip]
      {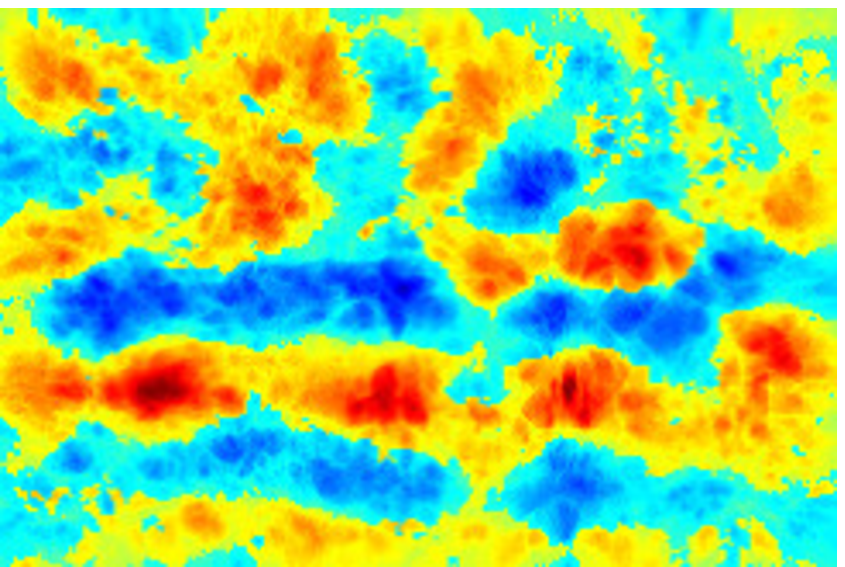}
      \end{center}
      \end{minipage}&

	  \begin{minipage}{0.25\columnwidth}
      \begin{center}
      \includegraphics[width=1\columnwidth,keepaspectratio=true,clip]
      {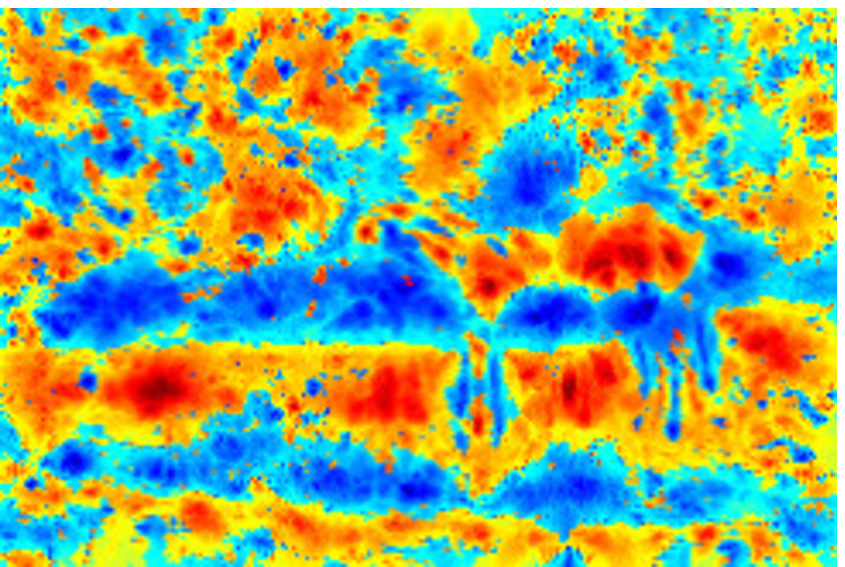}
      \end{center}
      \end{minipage}\\

      \begin{minipage}{0.25\columnwidth}
      \begin{center}
      \includegraphics[width=1\columnwidth,keepaspectratio=true,clip]
      {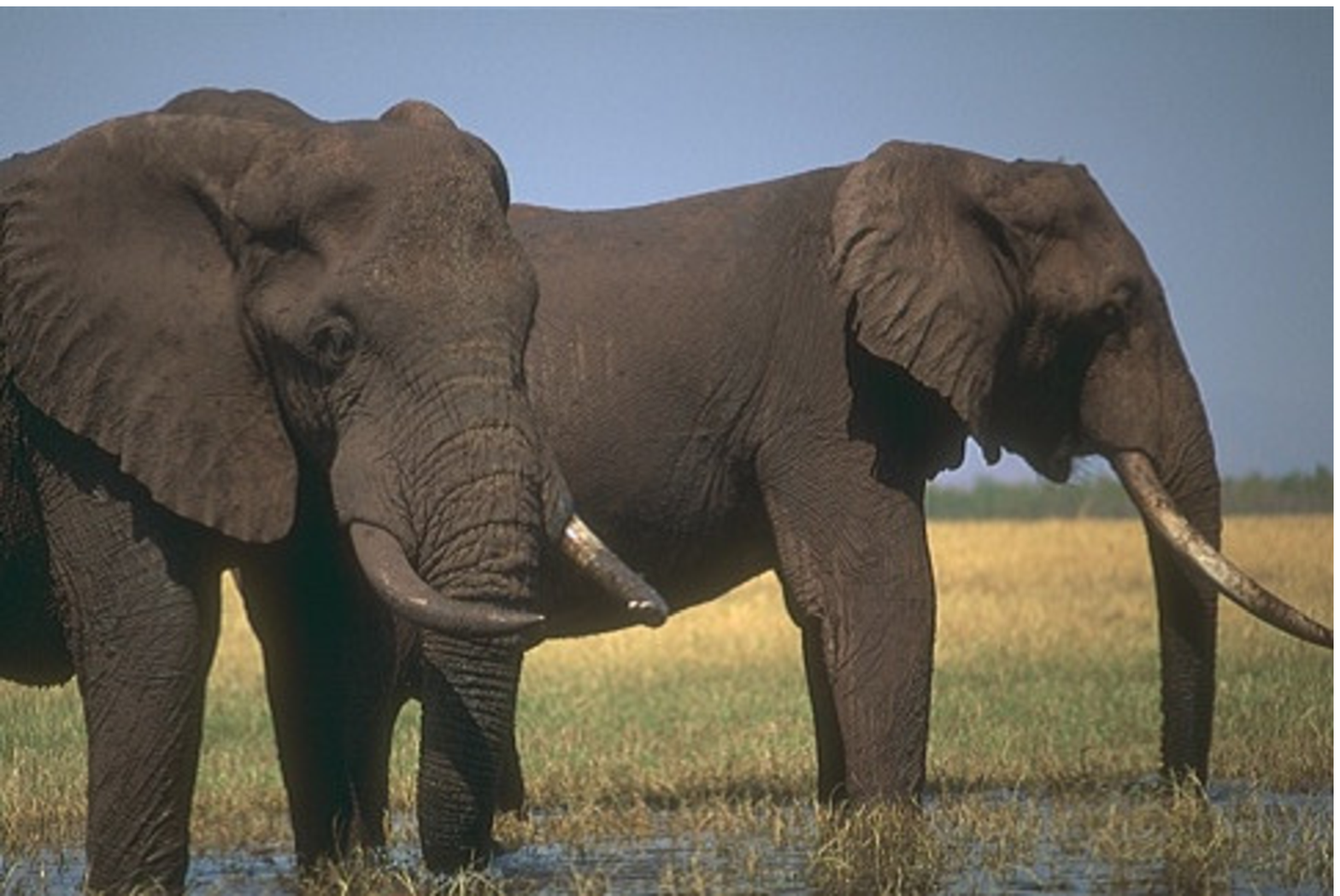}
      \end{center}
      \end{minipage}&

      \begin{minipage}{0.25\columnwidth}
      \begin{center}
      \includegraphics[width=1\columnwidth,keepaspectratio=true,clip]
      {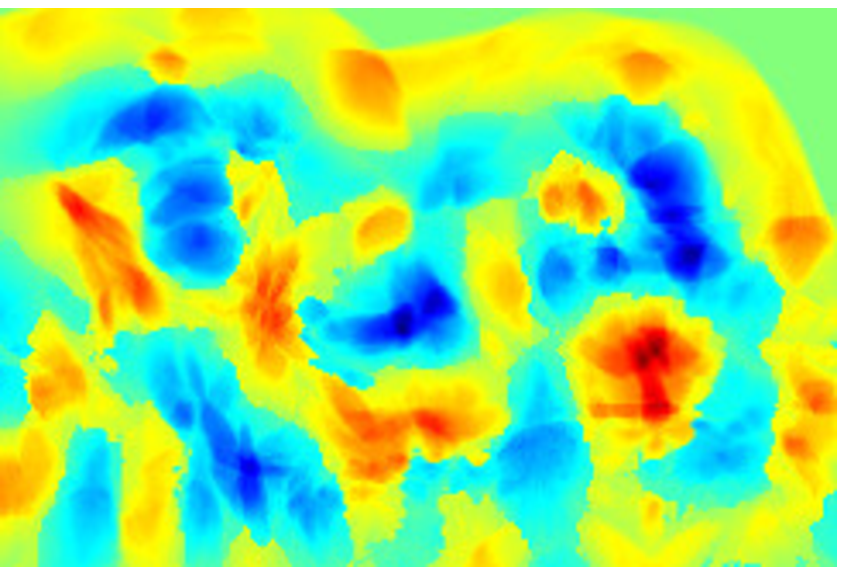}
      \end{center}
      \end{minipage}&

      \begin{minipage}{0.25\columnwidth}
      \begin{center}
      \includegraphics[width=1\columnwidth,keepaspectratio=true,clip]
      {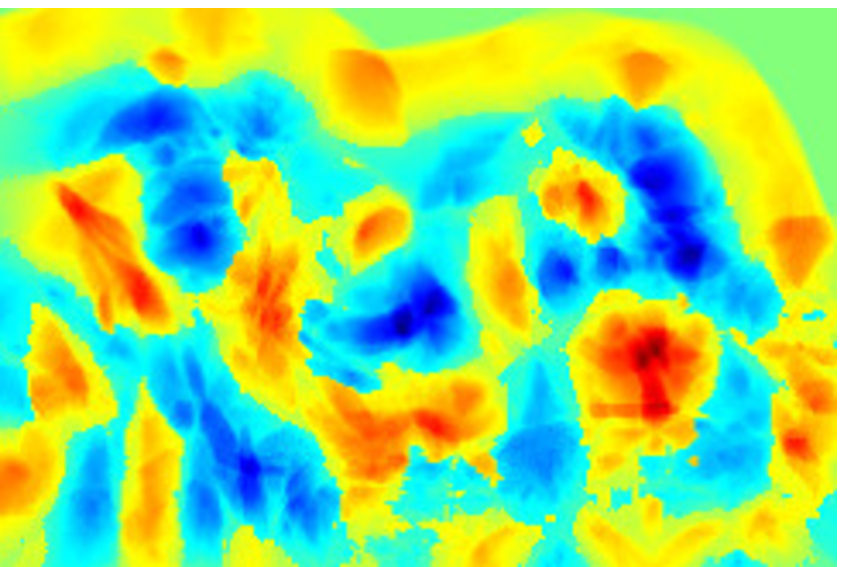}
      \end{center}
      \end{minipage}&

      \begin{minipage}{0.25\columnwidth}
      \begin{center}
      \includegraphics[width=1\columnwidth,keepaspectratio=true,clip]
      {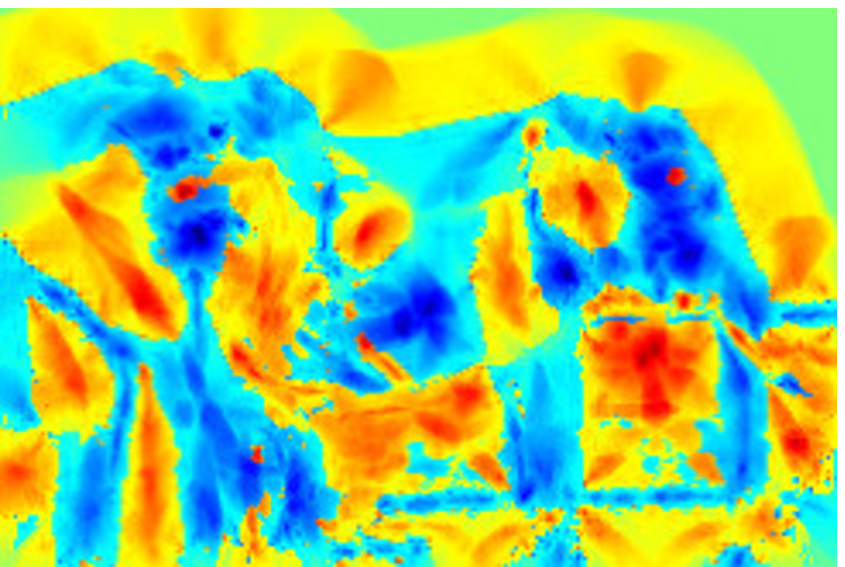}
      \end{center}
      \end{minipage}\\

	  (a) & (b) & (c) & (d)
    \end{tabular}
  \end{center}
\caption{Torque value maps with different normalization factors.
(a)  Test images;
(b)-(d)  torque value maps corresponding to the normalization factor $\alpha$ shown at the top of each column.
As  $\alpha$ gets larger, the number of positive and negative regions in the torque value map increases.
\label{fig:Torque value maps with different normalization factor}}
\end{figure}

Another way to modify the scale selection is by modifying the normalization factor. This is motivated by the work of  Galun et al.~\cite{Galun2007}, who studied the ideal threshold for multi-scale edge detection under Gaussian noise assumptions.
The torque for a patch is defined (from eq. \ref{eq:torque for patch pixel-wise}) as:
\begin{align}
\tau_{P,p} &= \frac{1}{2 Z}\sum_{q\in E\left(P\right)}\tau_{pq},
\end{align}
where  $Z$ is the normalization factor, and it was set to $|P|$, i.e. the area of the patch in eq.~(\ref{eq:torque for patch pixel-wise}). For a disk patch of radius $R$ it amounts
to $|P|=\pi R^2$.
We now vary the normalization factor by introducing a parameter $\alpha$  as follows:
\begin{align}
Z &= \pi R^\alpha.
\end{align}
Fig.~\ref{fig:Torque value maps with different normalization factor} shows torque value maps  for different  $\alpha$ values.
As can be seen from the figure, the torque value maps  become   smoother as $\alpha$ gets smaller.
 This is because  image patches of larger size now have a  smaller normalization factor, and therefore they tend to produce relatively higher torque values. As a result larger edge structures get favored.
Thus, a smaller $\alpha$ gives the effect of smoothing the torque value maps. An $\alpha=1$  corresponds to a normalization by the square root of the patch area, which is also the ideal  normalization of the area term derived  by \cite{Galun2007} for edge multi-scale detection.
In general, the amount of smoothing preferred will depend on the application, and by adjusting  the normalization factor, $\alpha$, one can easily adapt the torque operator.

\subsection{Efficient Torque Computation}

The computation of torque would be time consuming with a straightforward implementation.
However, the torque can be computed efficiently  using the concept of integral images, which will be explained next.  \ref{sec:gradient} then discusses a slightly modified torque definition, which allows for a very efficient computation.

First, let us quickly review the concept. An integral image (or summed area table) is a data-structure and algorithm to generate the sum of values in rectangular areas of an image. Let $k(x,y)$  be some image quantity. The  value  $K(x,y)$ for the  region of pixels in the range $[0 \dots x, 0 \dots y]$ amounts to
\begin{equation}\label{eq:k}
    K(x,y) =\sum_{u\leq x} \sum_{v\leq y} k(x,y),
\end{equation}
and it can be computed in a single pass over the image
as
\begin{equation}\label{eq:K_(x,y)}
   K(x,y) = k(x,y) + K(x-1,y) + K(x, y-1) - K(x-1,y-1).
\end{equation}
Once the table has been created, the sum of values in any rectangular region $[a \dots b, c \dots d]$
 can be computed  in constant time
with only one addition and two subtractions
as:
\begin{eqnarray}\label{eq:sum}
\sum_{u=a}^{b} \sum_{v=c}^{d}k(u,v)&=& K(a,c) + K(b, d) -K(a, d) - K(b, c).
\end{eqnarray}

Let us now explain  how to use these concepts  to compute the torque.
Let $o$ be the origin of the image coordinate system in the left top corner of the image,
and let $p$ be the center of a patch $P$, and  $\overrightarrow {F_q}$  the edge vector at $q$.
 We emphasize that now we need to change the  center  with respect to which  we compute the torque, which is denoted as  subscript in our notation.
Using a vector notation, the torque of the patch $P$ with respect to center $p$ without normalization can be written as:
\begin{eqnarray}\label{eqn:torque_integral}
\vec{\tau}_{P,p} \cdot (2\left|P\right|)
& = & \sum\limits_{q \in P} \overrightarrow {pq} \times  \overrightarrow {F_q} \notag \\
& = & \sum\limits_{q \in P} (\overrightarrow {po} + \overrightarrow {oq}) \times  \overrightarrow {F_q} \notag  \\
& = & -\overrightarrow {op} \times \sum_{q \in P} {\overrightarrow {F_q}} + \sum_{q \in P} {\overrightarrow{oq}} \times {\overrightarrow{F_q}}.
\end{eqnarray}
The first term in eq. (\ref{eqn:torque_integral}) amounts to the cross-product of the vector from the origin to $p$ and the vector of the sum of edges in the patch. The second term is the torque computed with respect to the origin of the image. If we express the vectors in terms of their components to denote $\overrightarrow {op}=(x,y)$, $\overrightarrow {oq}=(u,v)$ and   $F_q = (F^x(u,v), F^y(u,v))$, we can rewrite eq. (\ref{eqn:torque_integral})
as
\begin{align}
\vec{\tau}_{P,p} \cdot (2\left|P\right|) &=\nonumber\\
&\left. - x \sum_{\left(u,v\right)\in P} F^y\left(u,v\right) + y \sum_{\left(u,v\right)\in P} F^x\left(u,v\right)\right.\nonumber\\
&\left. + \sum_{\left(u,v\right)\in P} u F^y\left(u,v\right) - \sum_{\left(u,v\right)\in P} v F^x\left(u,v\right)\right.
\label{eq:torque summation}
\end{align}
Now it becomes clear that for each of the terms $F^x$, $F^y$,$u F^y$ and $v F^x$  we can pre-compute summed area tables with respect to the origin, and then compute the values  for any patch in linear time to derive the torque of the patch.

Further efficiency in the torque computation can be obtained by approximating the edge orientation to one of equally  divided eight directions:
\begin{align}
\theta_i = \left(i-1\right)\frac{2\pi}{8},
\end{align}
where $i\in \left\{1,2,3,4,5,6,7,8\right\}$ is the index to the orientation.
An edge vector of orientation $\theta_i$  is represented by a unit vector
$\left(\cos\theta_i,\sin\theta_i\right)$.
The torque at a point $q = (x,y)$ with respect to center $o$ for an edge vector in orientation $i$ then amounts to
\begin{align}
\tau_{oq}(i) &= (x \sin \theta_i-y \cos\theta_i )\cdot \delta\left(q,i\right), \label{eq:torque for edge point approximated}
\end{align}
where $\delta\left(q,i\right)\in \left\{0,1\right\}$ is a binary indicator for the existence of an edge at pixel $q$ in the  orientation indicated by $i$, and
the torque at  point $q$ is:
\begin{align}
\tau_{oq}  &= \sum^{8}_{i=1} \tau_{oq}(i). \label{eq:torque approximate}
\end{align}

With this approximation, the torque  in an  image patch before normalization in eq. (\ref{eqn:torque_integral}) becomes
\begin{align}
\vec{\tau}_{P,p} \cdot (2\left|P\right|)=& \nonumber\\
&\left. \sum^{8}_{i=1} \left((- x \sin \theta_i + y \cos \theta_i) \sum_{\left(u,v\right)\in P} \delta\left((u,v),i\right) \right
) \right.\nonumber\\
&\left. + \sum^{8}_{i=1} \sum_{\left(u,v\right)\in P} \tau_o((u,v),i) \right.
\label{eq:torque summation2}
\end{align}
We then implement the torque by keeping two three-dimensional arrays, one
 to store the edges, represented as $\delta\left(x,y,i\right)$, and  one to store the torque values $\tau\left(x,y,i\right)$, and we compute summed area tables for each $\delta\left(x,y,i\right)$ and  $\tau\left(x,y,i\right)$.
The computational cost is independent of  the patch size. For any given patch size, computing the torque for all patches in the  image is linear in the  number of pixels $N$, i.e. $O\left(N\right)$.

Finally, let us  look at the computational efficiency of the Torque operator  and compare it to common approaches.
The well-known SIFT algorithm involves  two steps: detection of  key points and generation of a description.
The key point search in SIFT, which is like  blob detection, has  similarity to the extrema detection in the torque volume as both are  multi-scale  localization procedures of interest points.
As analyzed in \cite{Grabner2006}, the highest costs in the key point search in  SIFT  are due to the multiple convolutions with the Gaussian operator to compute extrema in DoG space. Although it is common to down-sample the  image to avoid  computational costs due to increasing operator size, the convolutions  require time linear in the  size of image, $N$, times the size of Gaussian operator, $M$, for each scale, i.e. $O\left(N M\right)$. On the other hand, because of the use of integral images, the computational complexity deriving a torque map from edge responses, is linear in the image size.
The torque operator requires as input edges, whose computational cost depends on the sophistication of the edge detection algorithm. Assuming a simple edge detection, based on differences only, as  in the code provided, the computation of the torque is more efficient than standard interest point detection. However, we should note that alternative more efficient  approaches have been proposed for keypoint detection, such as the use of the box filter \cite{Grabner2006} or the Harris detector \cite{Suzuki2013} with integral images, instead of Laplacian filtering.

\section{Application}
\label{sec:app}

\begin{figure}[htbp]
\begin{center}
  \includegraphics[width=1.0\columnwidth,keepaspectratio=true,clip]
  {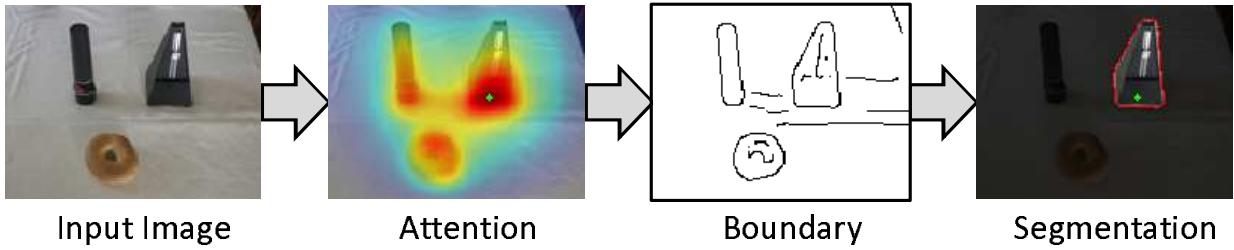}
\end{center}
\caption{Visual processing using the image torque operator.
\label{fig:Visual Processing using Torque}}
\end{figure}

Next we demonstrate the usefulness of the torque in a number of applications.
We focus on the   visual  processes for locating objects in the scene: visual attention, boundary detection, and foreground segmentation, as depicted in Fig.~\ref{fig:Visual Processing using Torque}.
Although there could be different approaches to object detection and localization when one considers single images, an approach involving  the above  three modules is necessary in mobile robot applications: First the
attention mechanism  focuses the processing to a conspicuous region - the region of interest. Then contours are extracted and the object in the region of interest is segmented.
We evaluate how much the proposed torque operator could improve these three processes by comparing against other methods in standard database settings. Finally, we demonstrate  the torque mechanism in a contour based object detection and recognition scheme. 

\subsection{Visual Attention}

As  discussed in section \ref{sec:Extrema in Torque}
the torque operator tends to produce extrema  at  points surrounded by boundary edges, and for  patch sizes corresponding to  object size.
This property of the torque operator is expected to be useful as a cue for bottom-up visual attention.
In the following experiment, we computed two torque-based saliency maps. One is generated as  mixture of Gaussians with the  Gaussian distributions centered at the  extrema in the torque volume, and the other is a weighted sum of the generated saliency map and the graph-based visual saliency (GBVS) by Harel \etal \cite{Harel2006}.
We used weights of 0.3 for the torque-based saliency map and 0.7 for the GBVS,
which were found empirically from  evaluations on subsets of the dataset.

We used the eye tracking data by Judd \etal. \cite{Judd2009} to generate ground truth saliency maps.
Fixation points were extracted from the data, and saliency maps were generated as mixture of Gaussian distributions centered at the fixation points, and the generated saliency maps were normalized in the range $[0,1]$. The ground truth saliency maps were  binarized by a threshold (we used 0.5) in the quantitative comparison.

We resized the test images such that the shorter side of the images was  150 pixels, in order to reduce computational time and standardize the image size.
The standard deviation of the Gaussian distributions used to generate the ground truth and torque-based saliency maps were both set to 25 pixels.

\begin{figure}[htbp]
  \begin{center}
    \begin{tabular}[t]{@{}c@{}c@{}}
      \begin{minipage}{0.50\columnwidth}
      \begin{center}
      \includegraphics[width=1\textwidth,keepaspectratio=true,clip]
      {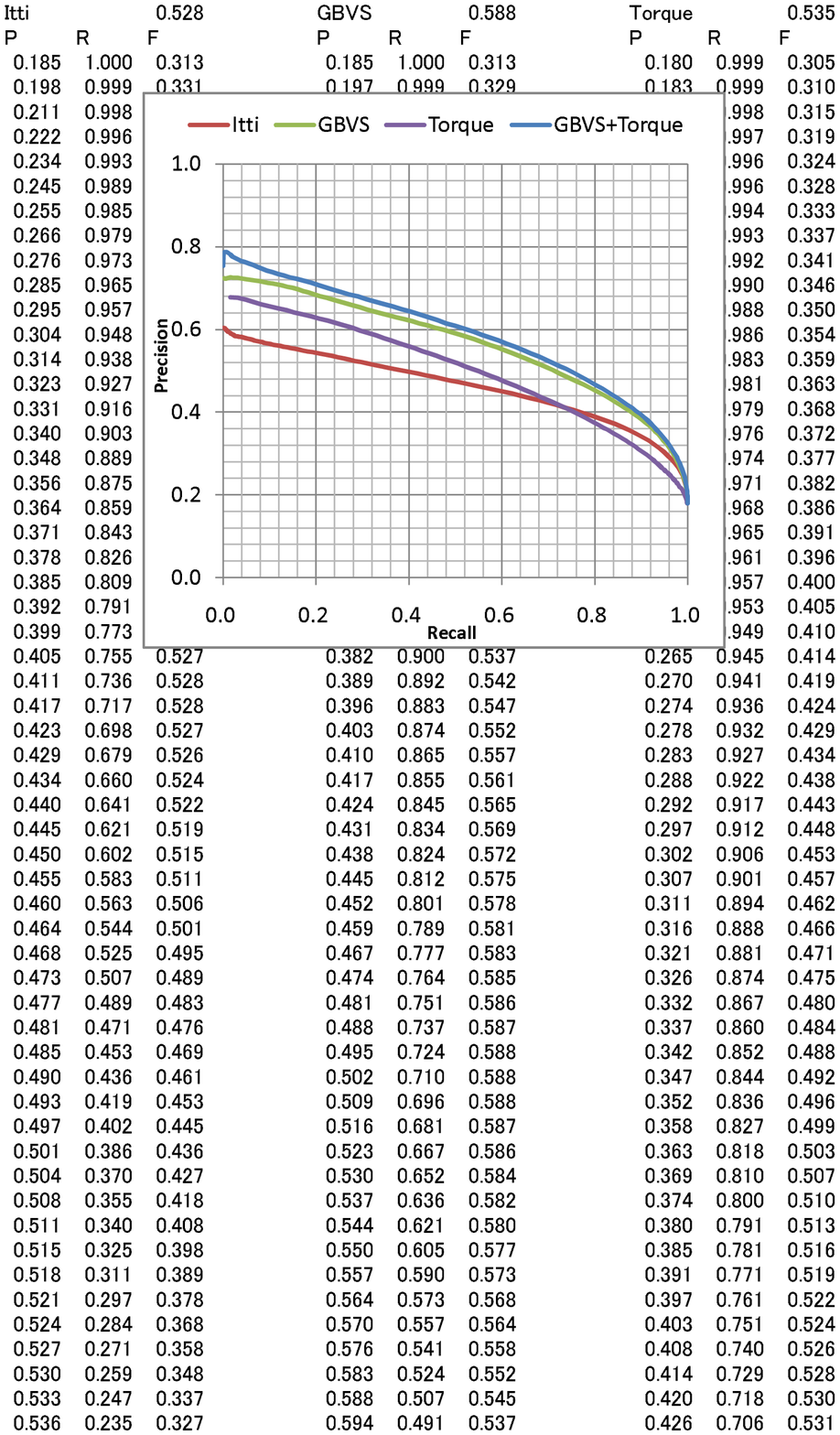}
      \end{center}
      \end{minipage}&

      \begin{minipage}{0.45\columnwidth}
      \begin{center}
      \begin{tabular}{|c|c|}
      \hline
	   method & F-measure \\
	  \hline
	  Itti & 0.528 \\
	  GBVS & 0.588 \\
	  Torque & 0.538 \\
	  GBVS+Torque & {\bfseries 0.599}\\
	  \hline
      \end{tabular}
      \end{center}
      \end{minipage}\\

      (a) & (b)
    \end{tabular}
  \end{center}
\caption{Evaluation of attention models. (a) ROC curves. (b) F-measure scores.
\label{fig:Evaluation of Attention Models}}
\end{figure}

\begin{figure}[htbp]
  \begin{center}
    \begin{tabular}{|@{}c@{}|@{}c@{}|@{}c@{}|@{}c@{}|@{}c@{}|}

	  \hline
	
	  \begin{minipage}{0.2\columnwidth}
      \begin{center}
	  \footnotesize{Itti \etal}
      \end{center}
      \end{minipage}&

	  \begin{minipage}{0.2\columnwidth}
      \begin{center}
	  \footnotesize{GBVS}
      \end{center}
      \end{minipage}&

	  \begin{minipage}{0.2\columnwidth}
      \begin{center}
	  \footnotesize{Troque}
      \end{center}
      \end{minipage}&

	  \begin{minipage}{0.2\columnwidth}
      \begin{center}
	  \scriptsize{GBVS+Torque}
      \end{center}
      \end{minipage}&

	  \begin{minipage}{0.2\columnwidth}
      \begin{center}
	  \footnotesize{Ground truth}
      \end{center}
      \end{minipage}\\
	
      \hline

	  \begin{minipage}{0.2\columnwidth}
      \begin{center}
      \includegraphics[width=1\columnwidth,keepaspectratio=true,clip]
      {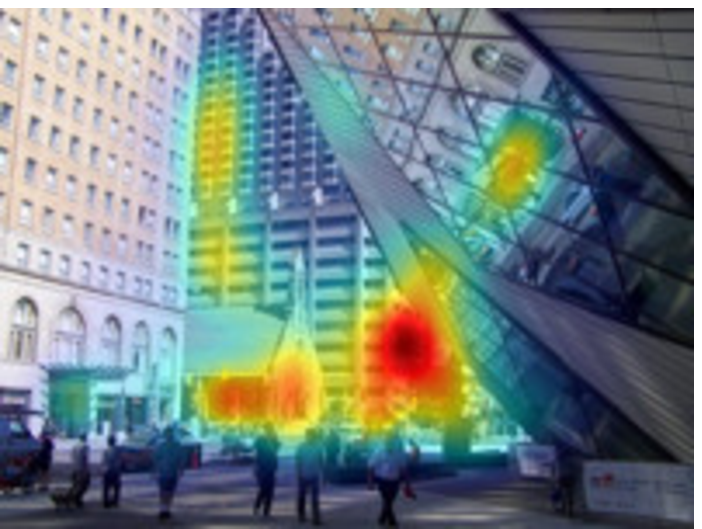}
      \end{center}
      \end{minipage}&
	
      \begin{minipage}{0.2\columnwidth}
      \begin{center}
      \includegraphics[width=1\columnwidth,keepaspectratio=true,clip]
      {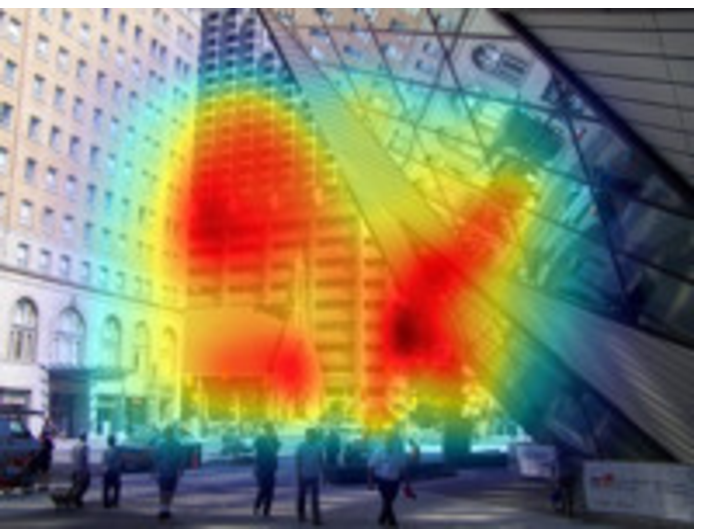}
      \end{center}
      \end{minipage}&
	
      \begin{minipage}{0.2\columnwidth}
      \begin{center}
      \includegraphics[width=1\columnwidth,keepaspectratio=true,clip]
      {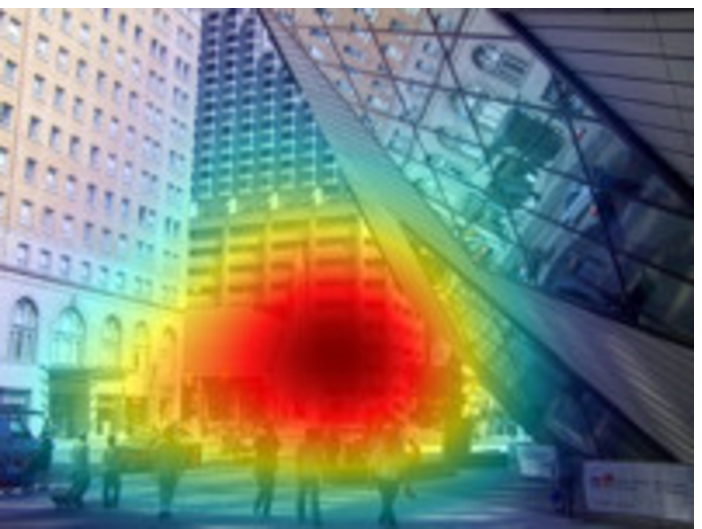}
      \end{center}
      \end{minipage}&

      \begin{minipage}{0.2\columnwidth}
      \begin{center}
      \includegraphics[width=1\columnwidth,keepaspectratio=true,clip]
      {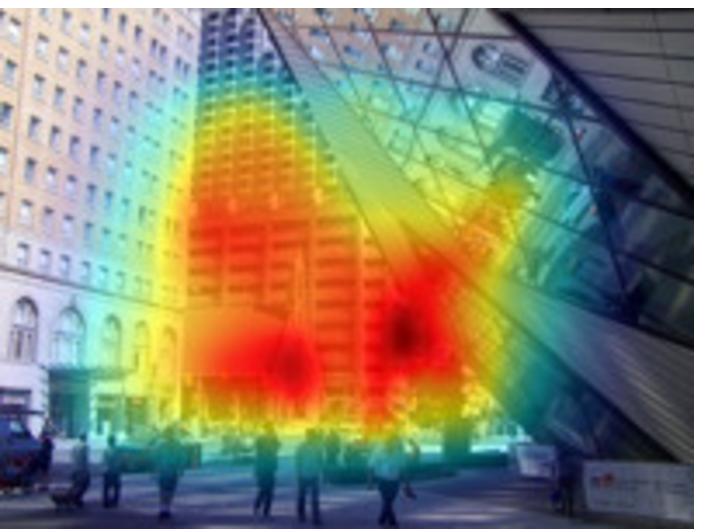}
      \end{center}
      \end{minipage}&

      \begin{minipage}{0.2\columnwidth}
      \begin{center}
      \includegraphics[width=1\columnwidth,keepaspectratio=true,clip]
      {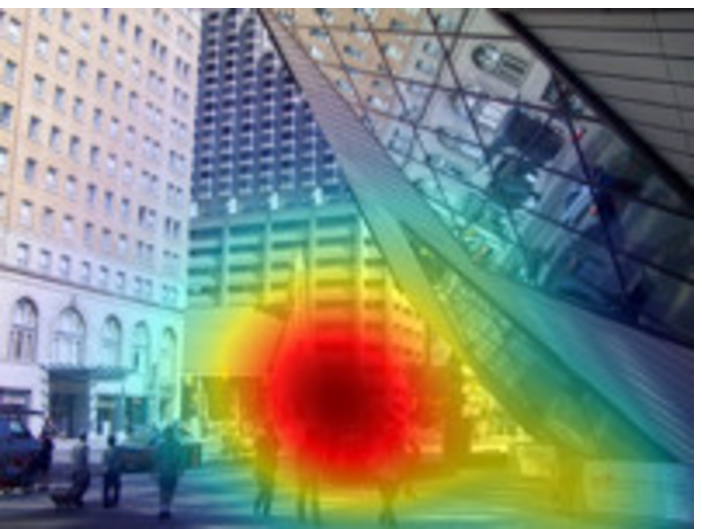}
      \end{center}
      \end{minipage}\\

      0.608 & 0.486 & {\bf 0.687} & 0.544 & \\

      \hline

	  \begin{minipage}{0.2\columnwidth}
      \begin{center}
      \includegraphics[width=1\columnwidth,keepaspectratio=true,clip]
      {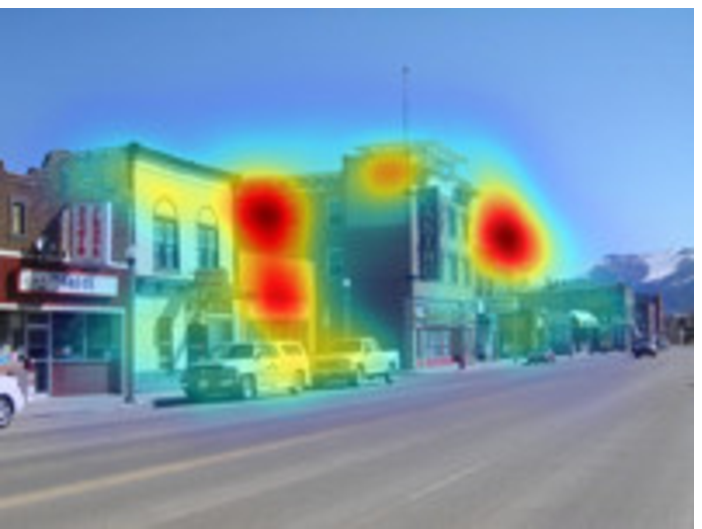}
      \end{center}
      \end{minipage}&
	
      \begin{minipage}{0.2\columnwidth}
      \begin{center}
      \includegraphics[width=1\columnwidth,keepaspectratio=true,clip]
      {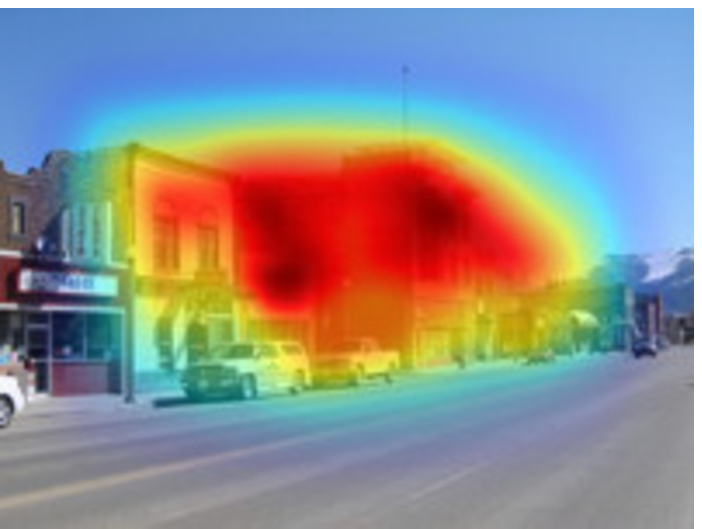}
      \end{center}
      \end{minipage}&
	
      \begin{minipage}{0.2\columnwidth}
      \begin{center}
      \includegraphics[width=1\columnwidth,keepaspectratio=true,clip]
      {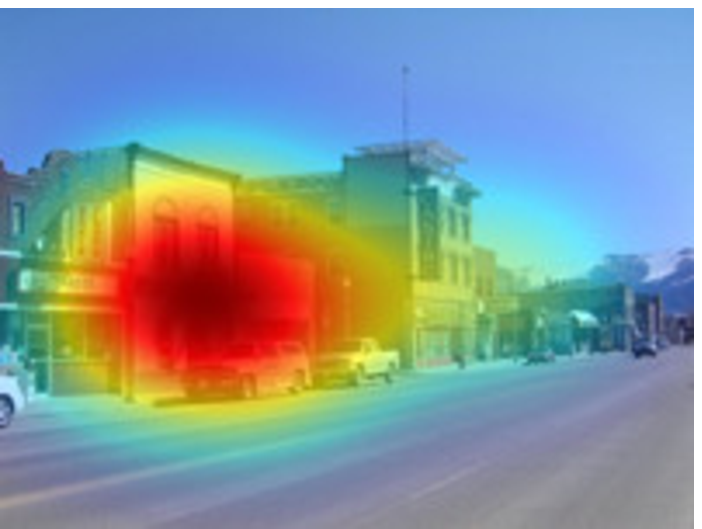}
      \end{center}
      \end{minipage}&

      \begin{minipage}{0.2\columnwidth}
      \begin{center}
      \includegraphics[width=1\columnwidth,keepaspectratio=true,clip]
      {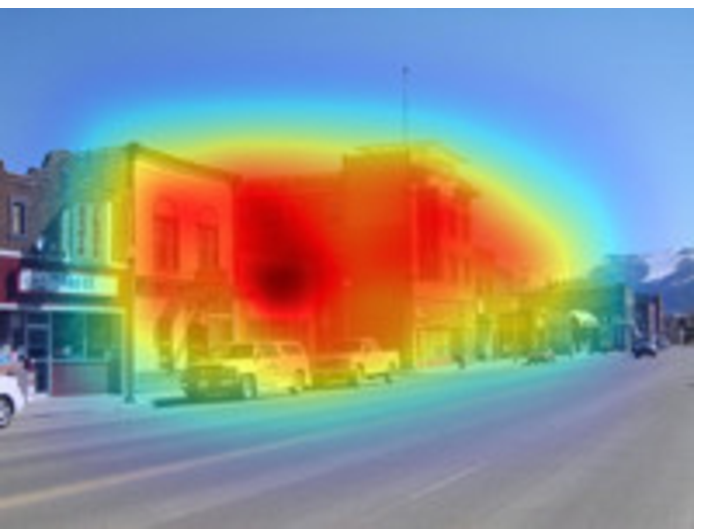}
      \end{center}
      \end{minipage}&

      \begin{minipage}{0.2\columnwidth}
      \begin{center}
      \includegraphics[width=1\columnwidth,keepaspectratio=true,clip]
      {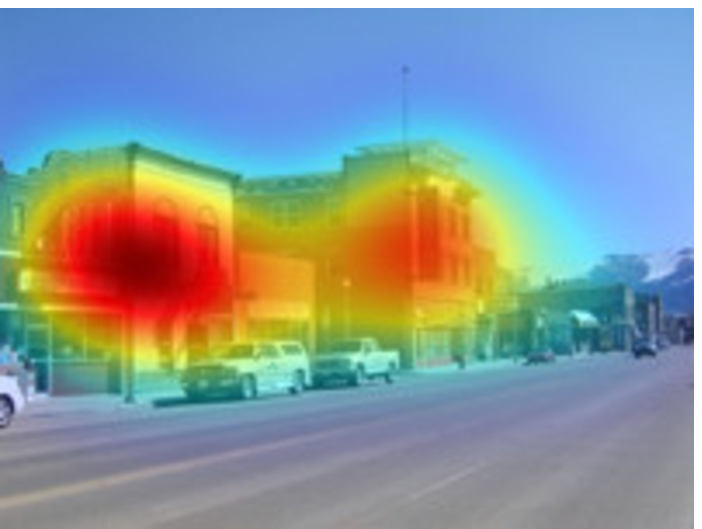}
      \end{center}
      \end{minipage}\\

	  0.700 & 0.718 & 0.789 & {\bf 0.794} & \\

      \hline

	  \begin{minipage}{0.2\columnwidth}
      \begin{center}
      \includegraphics[width=1\columnwidth,keepaspectratio=true,clip]
      {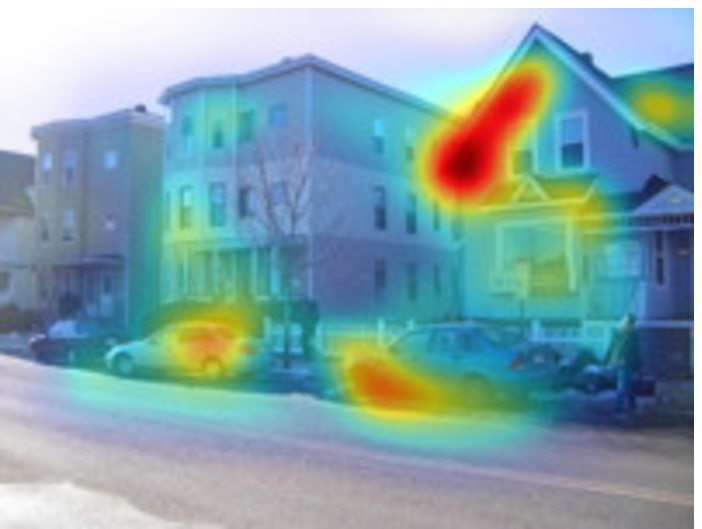}
      \end{center}
      \end{minipage}&
	
      \begin{minipage}{0.2\columnwidth}
      \begin{center}
      \includegraphics[width=1\columnwidth,keepaspectratio=true,clip]
      {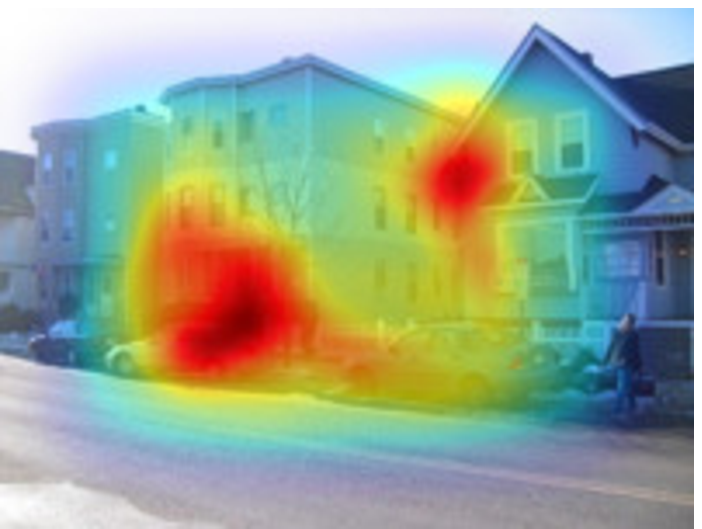}
      \end{center}
      \end{minipage}&
	
      \begin{minipage}{0.2\columnwidth}
      \begin{center}
      \includegraphics[width=1\columnwidth,keepaspectratio=true,clip]
      {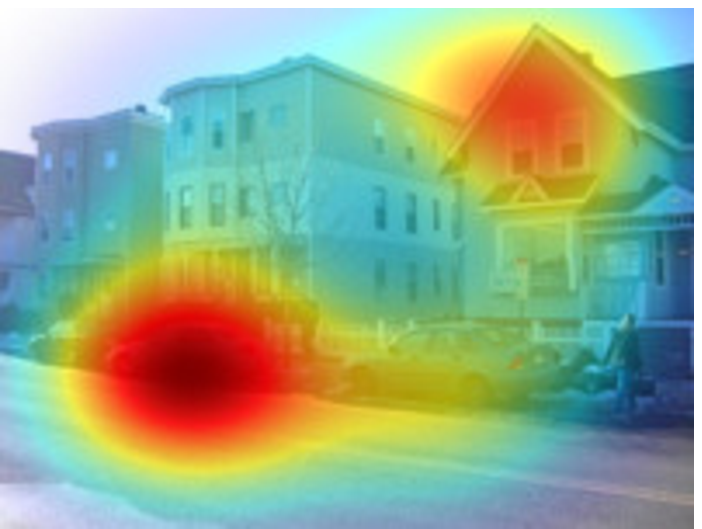}
      \end{center}
      \end{minipage}&

      \begin{minipage}{0.2\columnwidth}
      \begin{center}
      \includegraphics[width=1\columnwidth,keepaspectratio=true,clip]
      {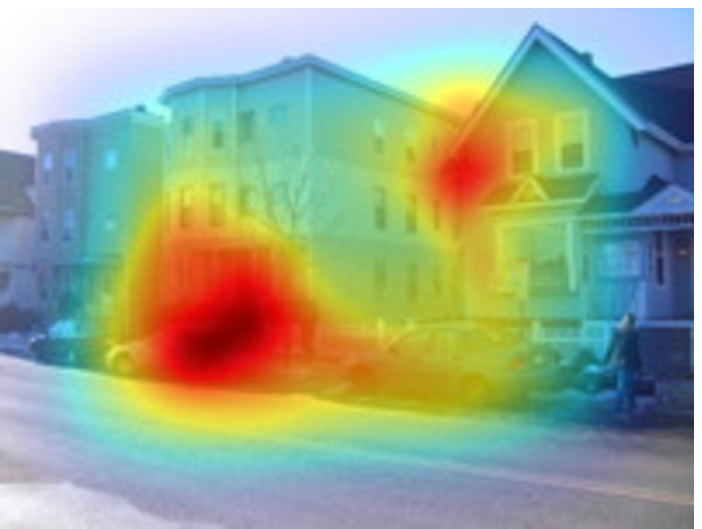}
      \end{center}
      \end{minipage}&

      \begin{minipage}{0.2\columnwidth}
      \begin{center}
      \includegraphics[width=1\columnwidth,keepaspectratio=true,clip]
      {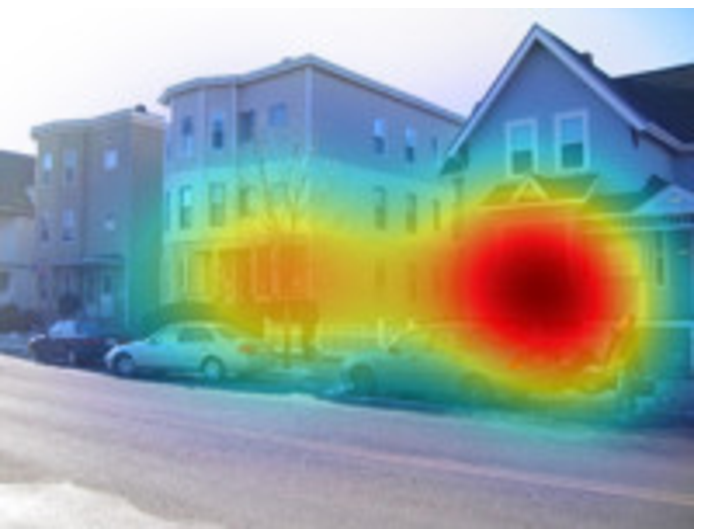}
      \end{center}
      \end{minipage}\\

      0.558 & {\bf 0.604} & 0.481 & 0.585 & \\

      \hline

	  \begin{minipage}{0.2\columnwidth}
      \begin{center}
      \includegraphics[width=1\columnwidth,keepaspectratio=true,clip]
      {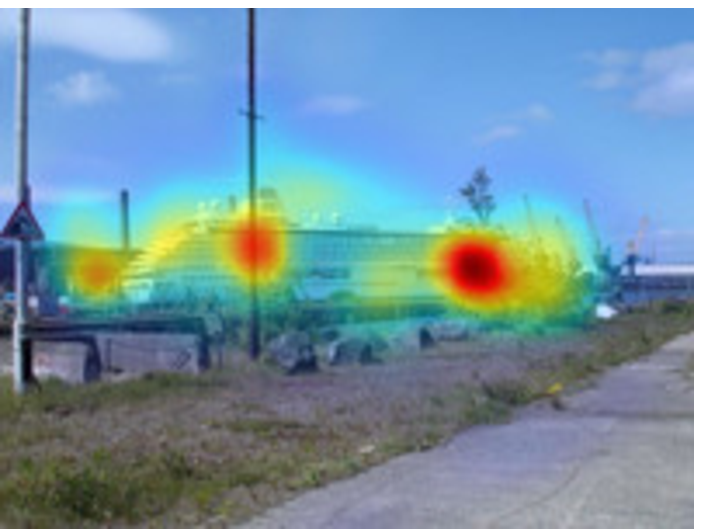}
      \end{center}
      \end{minipage}&
	
      \begin{minipage}{0.2\columnwidth}
      \begin{center}
      \includegraphics[width=1\columnwidth,keepaspectratio=true,clip]
      {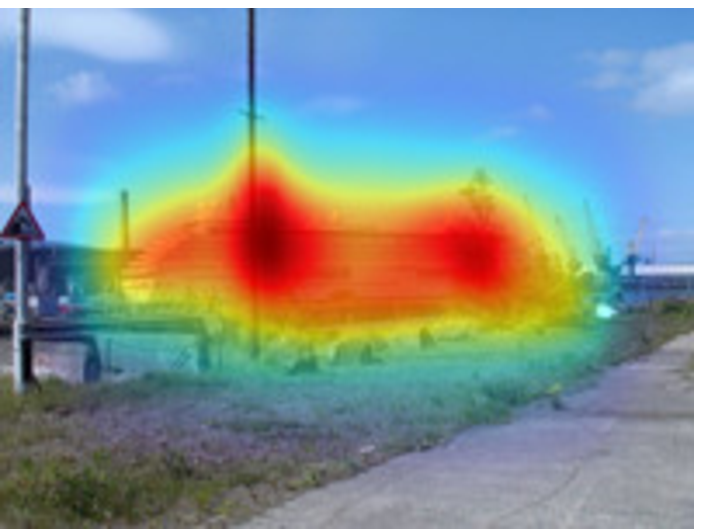}
      \end{center}
      \end{minipage}&
	
      \begin{minipage}{0.2\columnwidth}
      \begin{center}
      \includegraphics[width=1\columnwidth,keepaspectratio=true,clip]
      {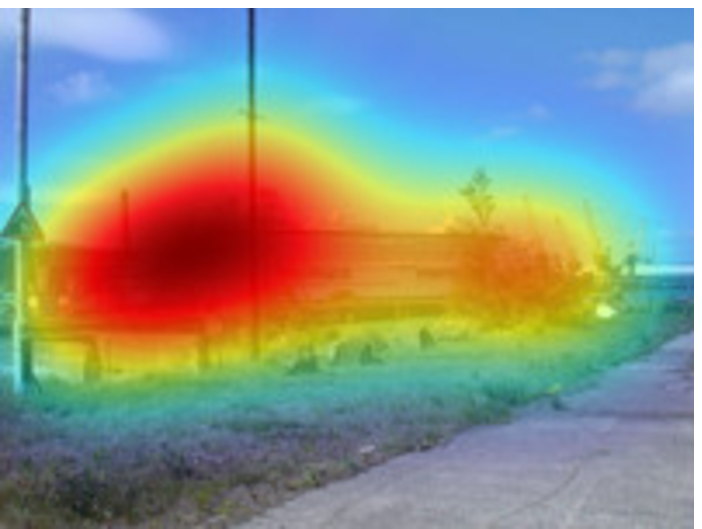}
      \end{center}
      \end{minipage}&

      \begin{minipage}{0.2\columnwidth}
      \begin{center}
      \includegraphics[width=1\columnwidth,keepaspectratio=true,clip]
      {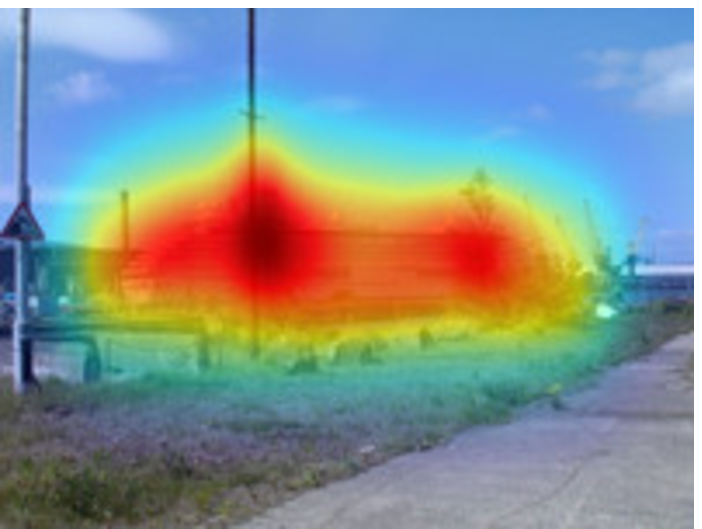}
      \end{center}
      \end{minipage}&

      \begin{minipage}{0.2\columnwidth}
      \begin{center}
      \includegraphics[width=1\columnwidth,keepaspectratio=true,clip]
      {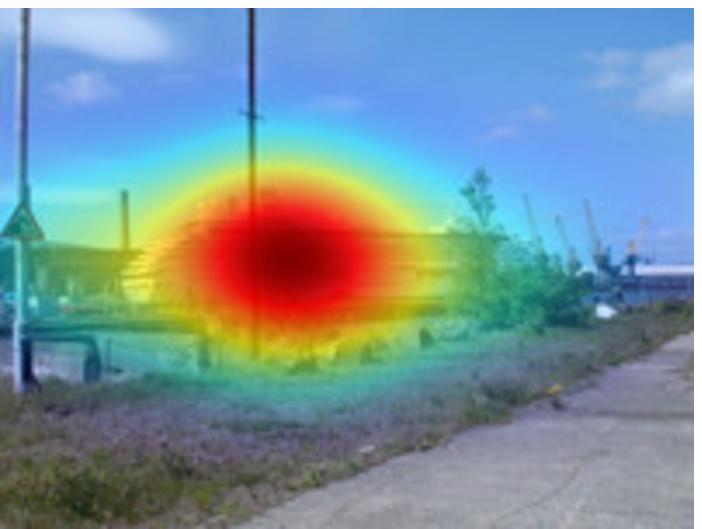}
      \end{center}
      \end{minipage}\\

      0.775 & 0.795 & 0.774 & {\bf 0.811} & \\

      \hline

	  \begin{minipage}{0.2\columnwidth}
      \begin{center}
      \includegraphics[width=1\columnwidth,keepaspectratio=true,clip]
      {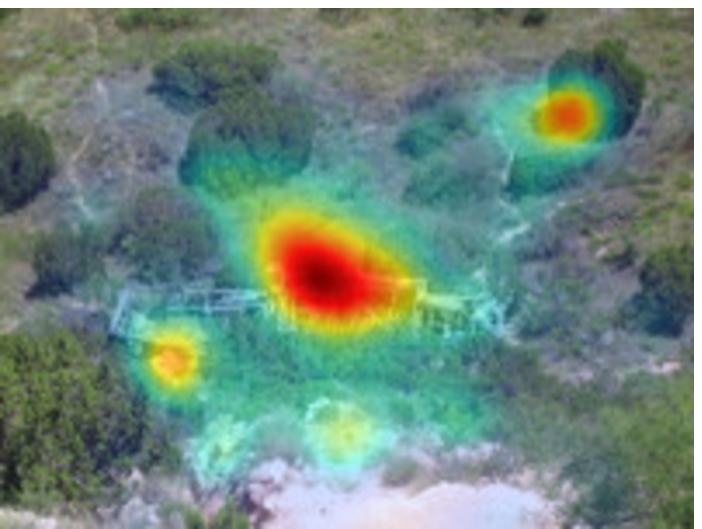}
      \end{center}
      \end{minipage}&
	
      \begin{minipage}{0.2\columnwidth}
      \begin{center}
      \includegraphics[width=1\columnwidth,keepaspectratio=true,clip]
      {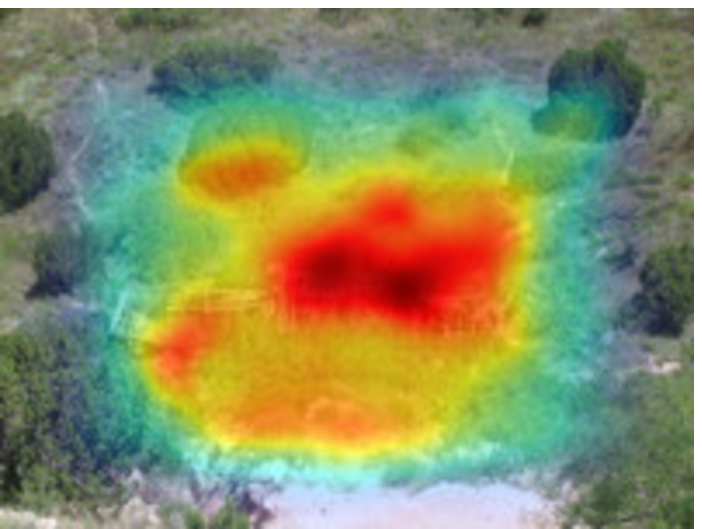}
      \end{center}
      \end{minipage}&
	
      \begin{minipage}{0.2\columnwidth}
      \begin{center}
      \includegraphics[width=1\columnwidth,keepaspectratio=true,clip]
      {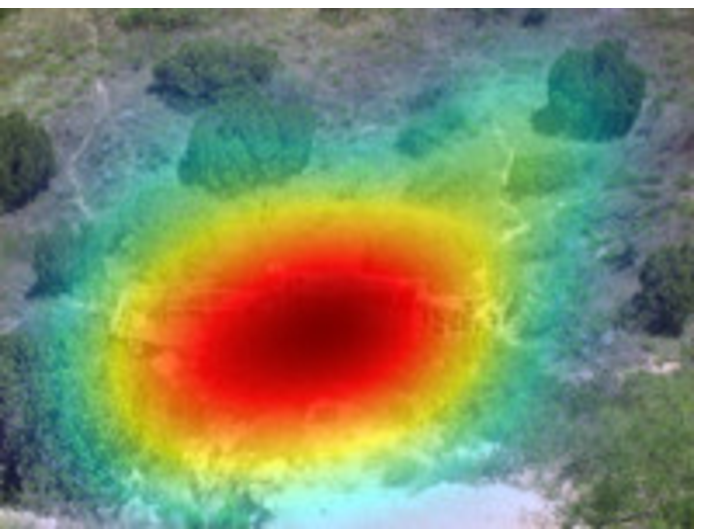}
      \end{center}
      \end{minipage}&

      \begin{minipage}{0.2\columnwidth}
      \begin{center}
      \includegraphics[width=1\columnwidth,keepaspectratio=true,clip]
      {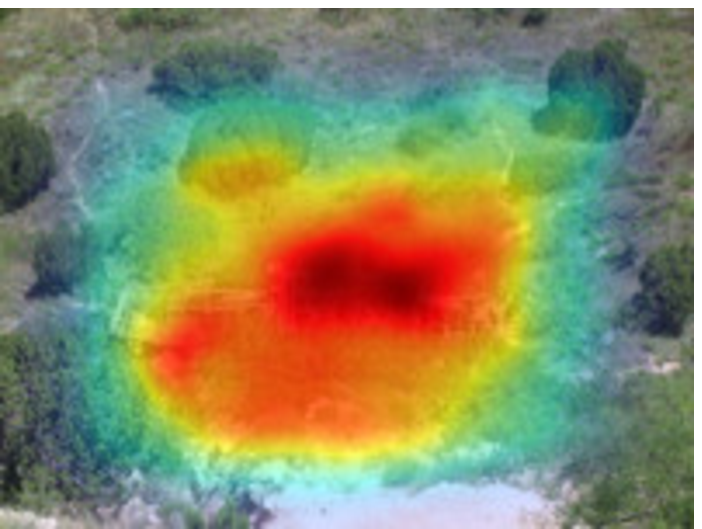}
      \end{center}
      \end{minipage}&

      \begin{minipage}{0.2\columnwidth}
      \begin{center}
      \includegraphics[width=1\columnwidth,keepaspectratio=true,clip]
      {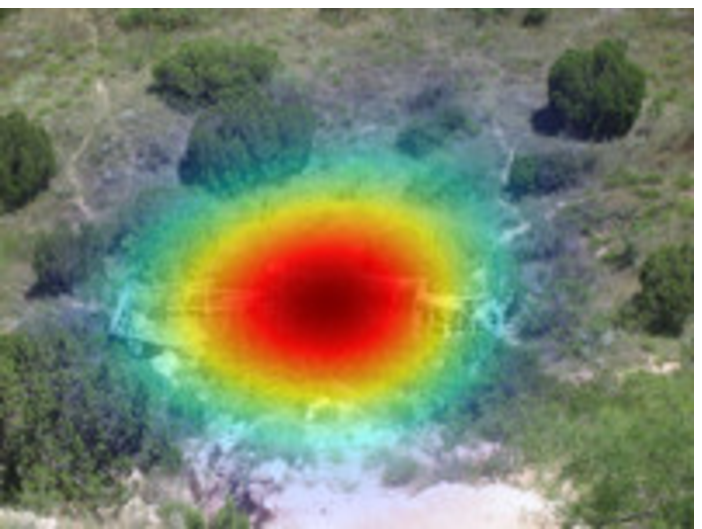}
      \end{center}
      \end{minipage}\\

      0.690 & 0.760 & 0.815 & {\bf 0.818} & \\

      \hline

    \end{tabular}
  \end{center}
\caption{Examples of Visual Attention.
Saliency maps computed by four different methods and the ground-truth saliency map  are overlayed on the respective test images. The maximum F-measure is shown below each saliency map.
\label{fig:Visual Attention Models Comparison}}
\end{figure}

The torque-based saliency maps were quantitatively compared with the saliency maps of \cite{Itti1998} and \cite{Harel2006}.
We binarized the computed saliency maps  for a set of threshold values  equally distant in $[0,1]$, and  evaluated precision and recall of the binarized saliency maps as follows:
\begin{align}
P = \frac{TP}{TP+FP}, \quad R = \frac{TP}{TP+FN},\quad\\
TP = \left|S \cap \mathcal{G}\right|,
FP = \left|S \cap \overline{\mathcal{G}}\right|,
FN = \left|\overline{S} \cap \mathcal{G}\right|,
\end{align}
where $S$ is the binarized  saliency map, and
$\mathcal{G}$ is the binarized ground truth saliency map.
$P$ and $R$ denote precision and recall respectively.
$TP$, $FP$, $FN$ are true positive, false positive, and false negative, respectively.
Fig.~\ref{fig:Evaluation of Attention Models} shows the
ROC curves and maximum F-measures computed from the  898 test images in the dataset.
Examples of computed saliency maps along  with the ground-truth are shown in Fig.~\ref{fig:Visual Attention Models Comparison}.
The quantitative comparison shows that  the attention map  based only on torque does not outperform GBVS.
However, the torque measure as additional mid-level visual cue improves the quality of GBVS.

\subsection{Boundary Detection}

\begin{figure}[htbp]
  \begin{center}
    \begin{tabular}[t]{@{}c@{\,}c@{\,}c@{}}
      \begin{minipage}{0.25\columnwidth}
      \begin{center}
      \includegraphics[width=1\textwidth,keepaspectratio=true,clip]
      {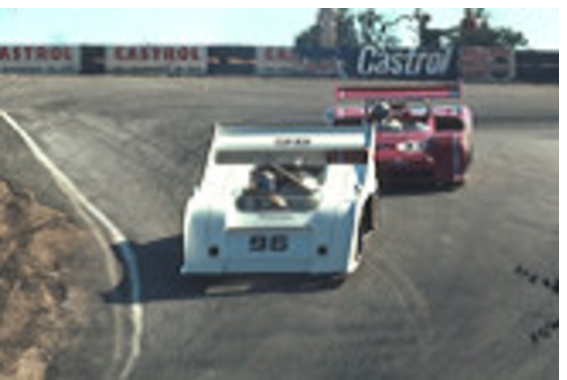}
      \end{center}
      \end{minipage}&

      \begin{minipage}{0.25\columnwidth}
      \begin{center}
      \includegraphics[width=1\textwidth,keepaspectratio=true,clip]
      {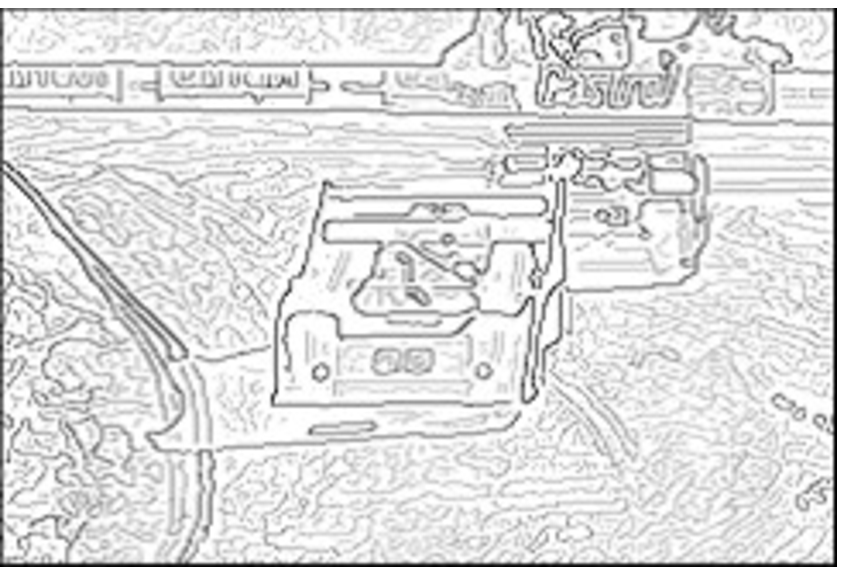}
      \end{center}
      \end{minipage}&

      \begin{minipage}{0.25\columnwidth}
      \begin{center}
      \includegraphics[width=1\textwidth,keepaspectratio=true,clip]
      {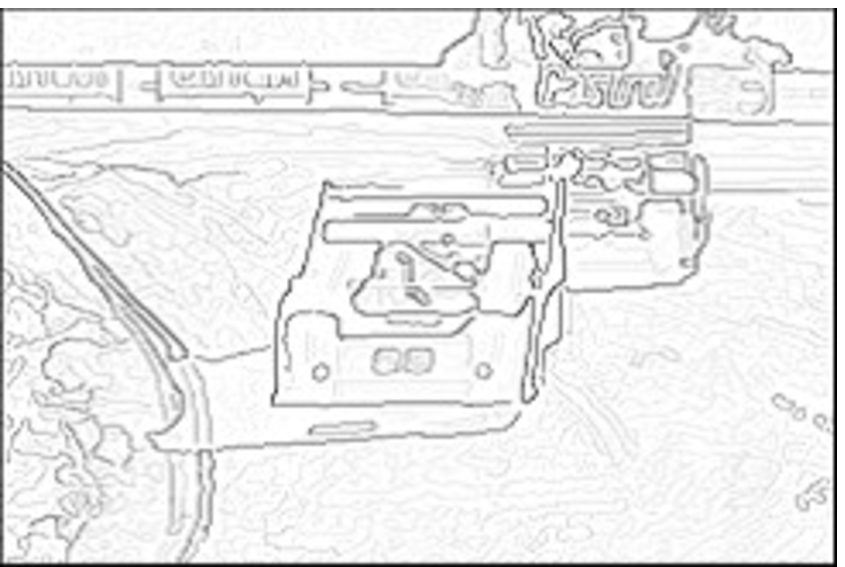}
      \end{center}
      \end{minipage}\\

      \begin{minipage}{0.25\columnwidth}
      \begin{center}
      \includegraphics[width=1\textwidth,keepaspectratio=true,clip]
      {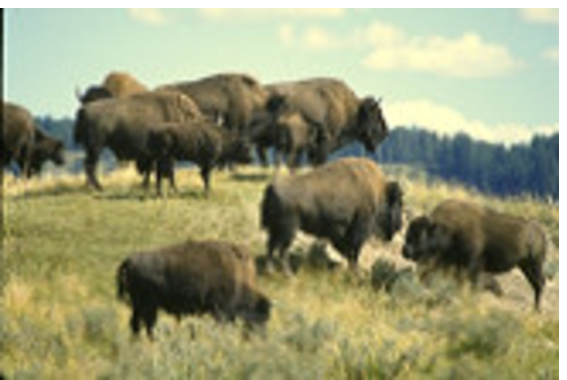}
      \end{center}
      \end{minipage}&

      \begin{minipage}{0.25\columnwidth}
      \begin{center}
      \includegraphics[width=1\textwidth,keepaspectratio=true,clip]
      {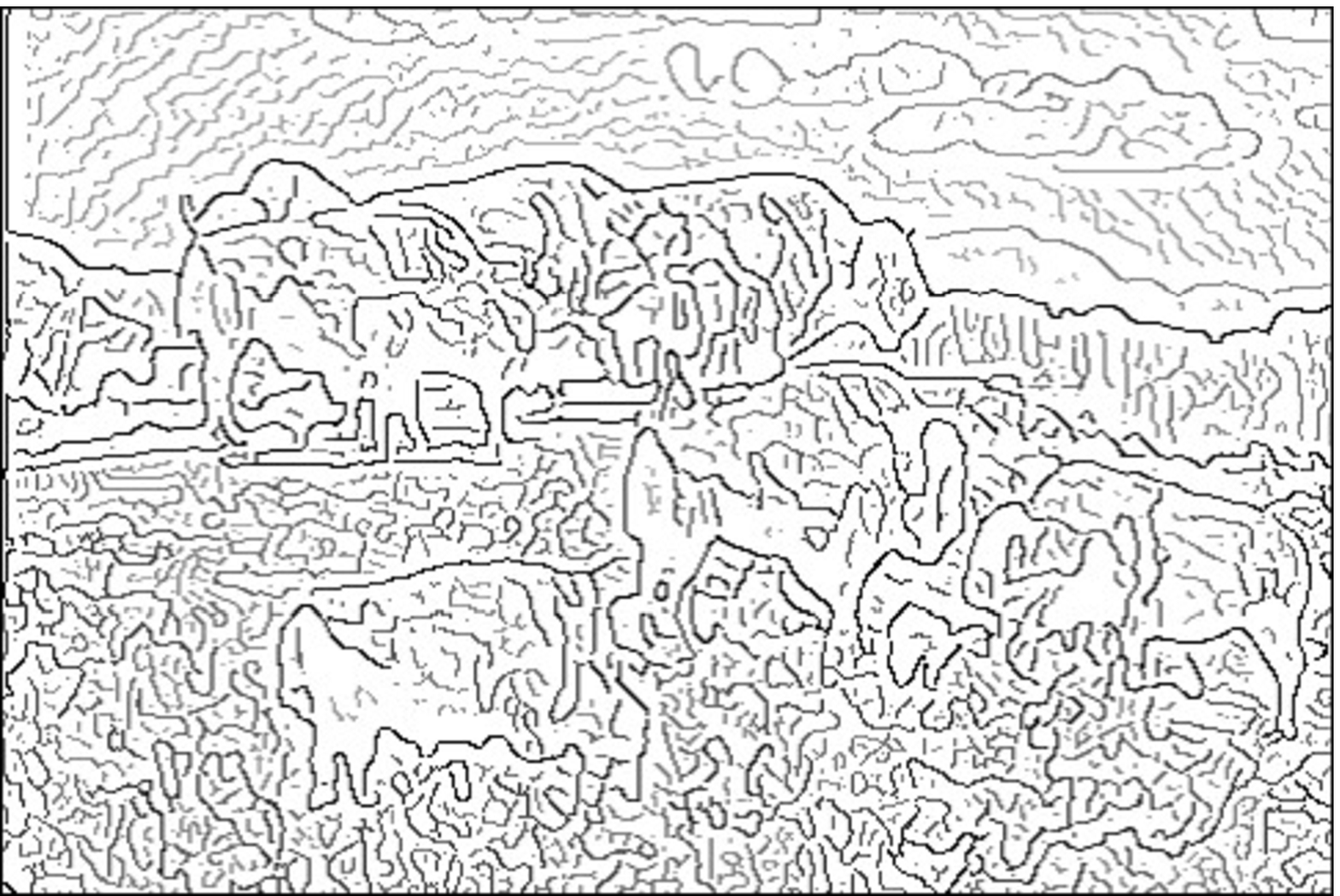}
      \end{center}
      \end{minipage}&

      \begin{minipage}{0.25\columnwidth}
      \begin{center}
      \includegraphics[width=1\textwidth,keepaspectratio=true,clip]
      {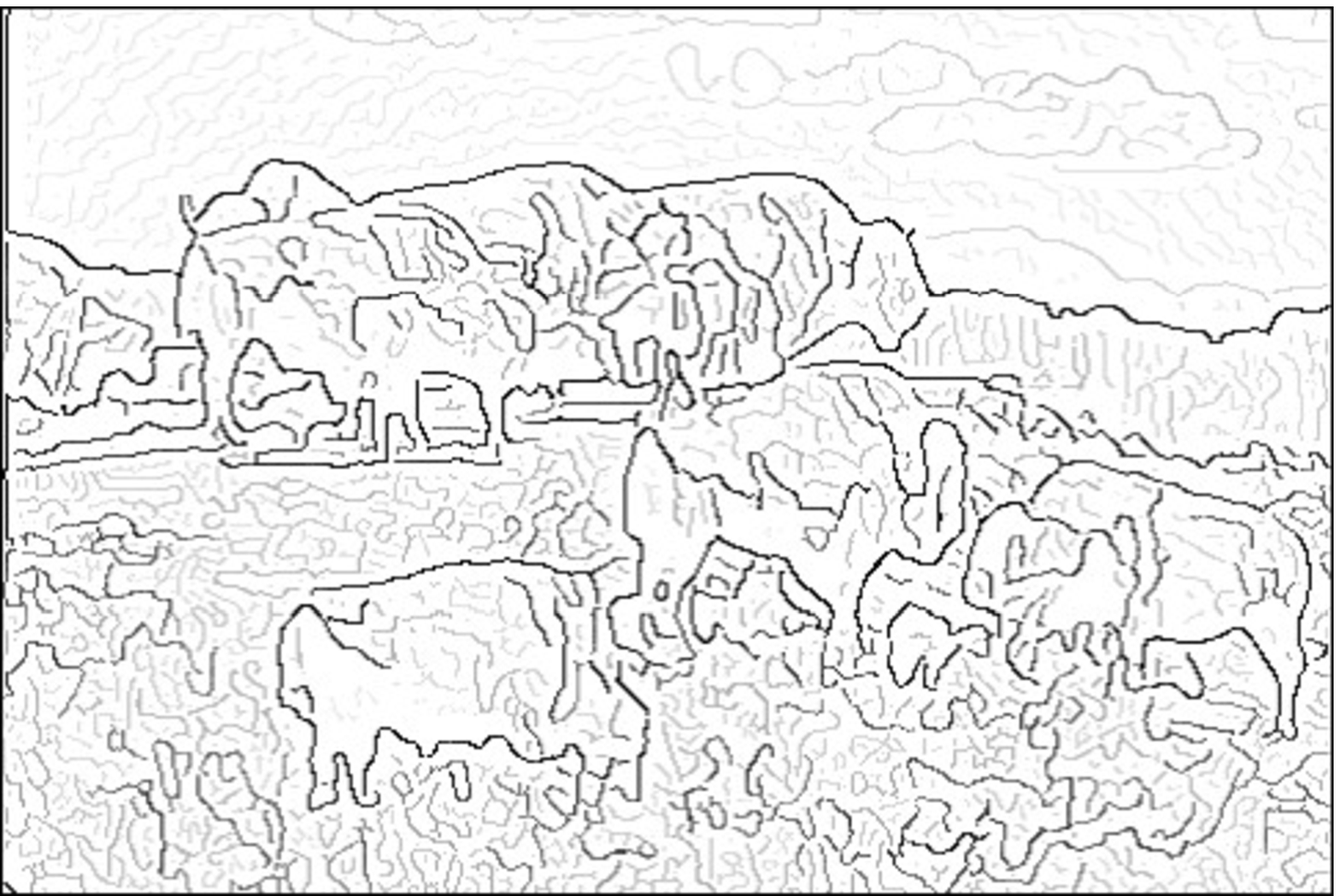}
      \end{center}
      \end{minipage}\\

      \begin{minipage}{0.25\columnwidth}
      \begin{center}
      \includegraphics[width=1\textwidth,keepaspectratio=true,clip]
      {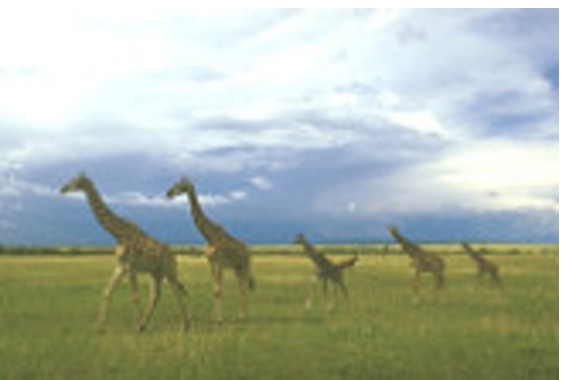}
      \end{center}
      \end{minipage}&

      \begin{minipage}{0.25\columnwidth}
      \begin{center}
      \includegraphics[width=1\textwidth,keepaspectratio=true,clip]
      {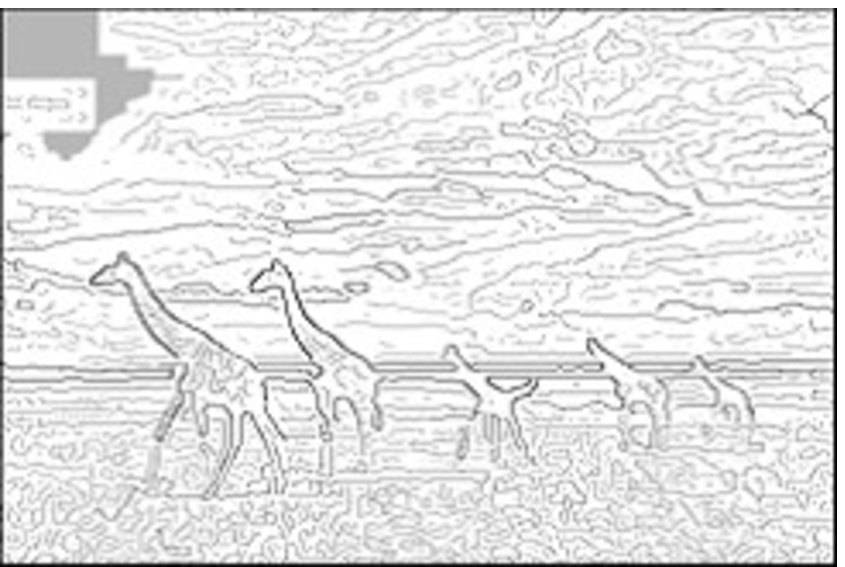}
      \end{center}
      \end{minipage}&

      \begin{minipage}{0.25\columnwidth}
      \begin{center}
      \includegraphics[width=1\textwidth,keepaspectratio=true,clip]
      {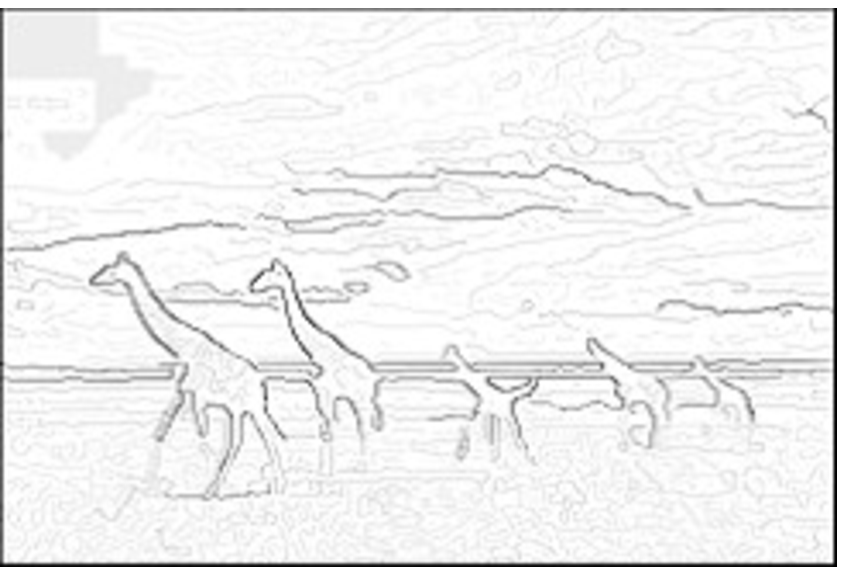}
      \end{center}
      \end{minipage}\\

      \begin{minipage}{0.25\columnwidth}
      \begin{center}
      \includegraphics[width=1\textwidth,keepaspectratio=true,clip]
      {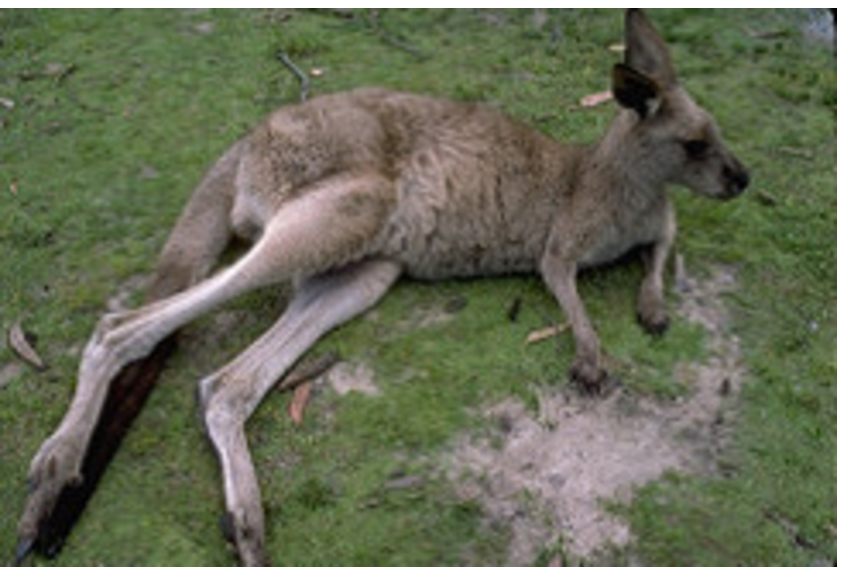}
      \end{center}
      \end{minipage}&

      \begin{minipage}{0.25\columnwidth}
      \begin{center}
      \includegraphics[width=1\textwidth,keepaspectratio=true,clip]
      {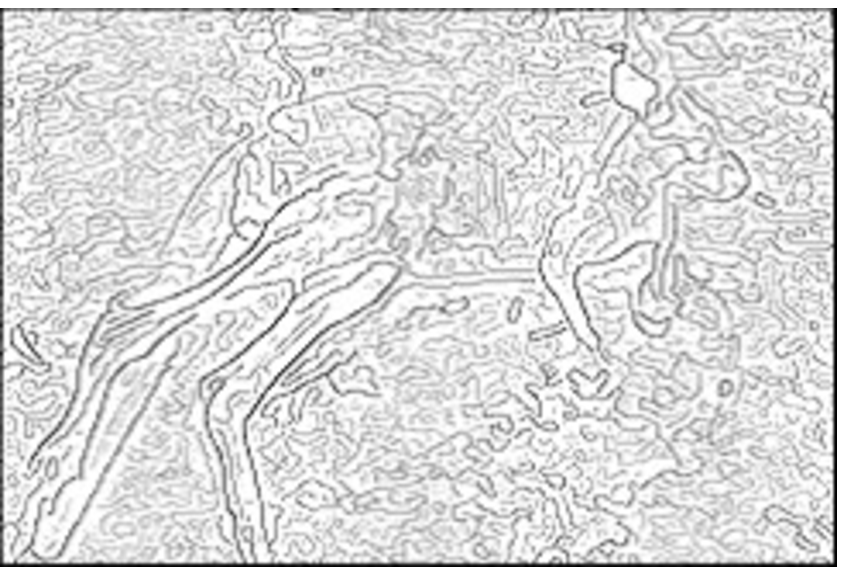}
      \end{center}
      \end{minipage}&

      \begin{minipage}{0.25\columnwidth}
      \begin{center}
      \includegraphics[width=1\textwidth,keepaspectratio=true,clip]
      {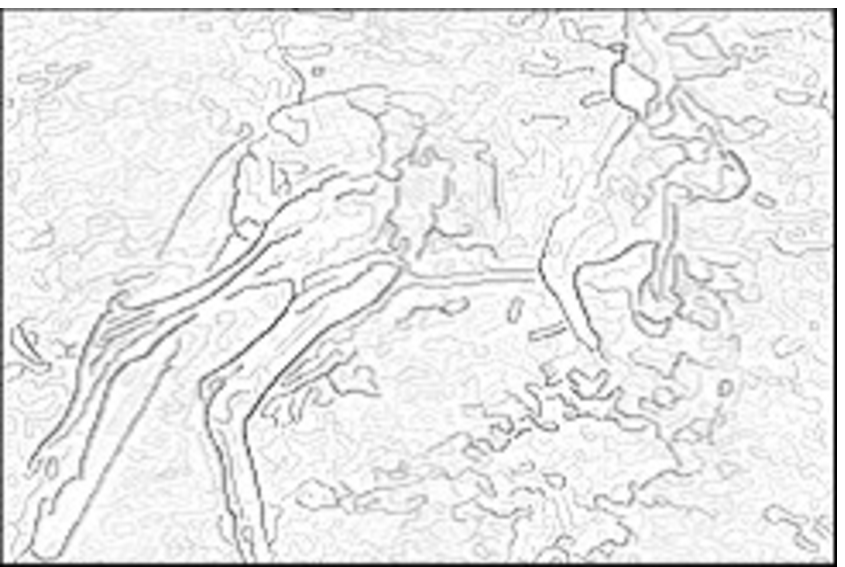}
      \end{center}
      \end{minipage}\\

      \begin{minipage}{0.25\columnwidth}
      \begin{center}
      \includegraphics[width=1\textwidth,keepaspectratio=true,clip]
      {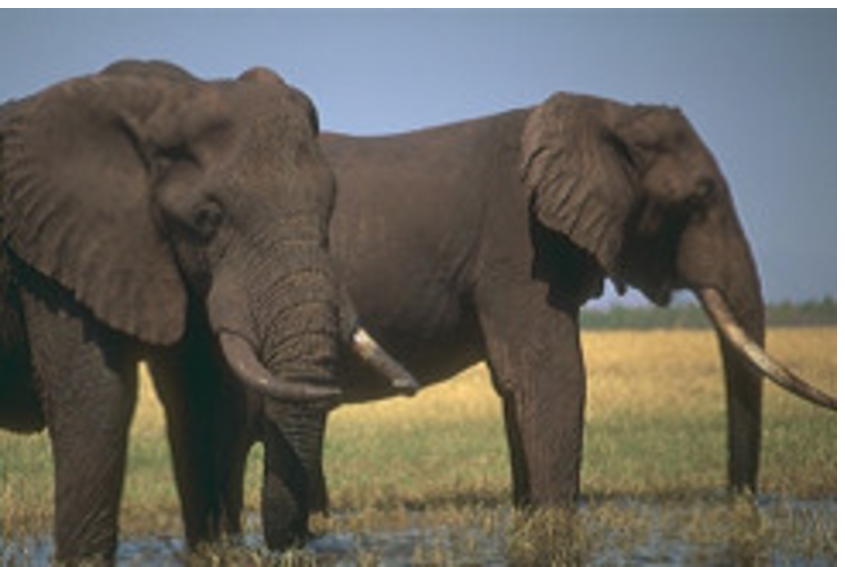}
      \end{center}
      \end{minipage}&

      \begin{minipage}{0.25\columnwidth}
      \begin{center}
      \includegraphics[width=1\textwidth,keepaspectratio=true,clip]
      {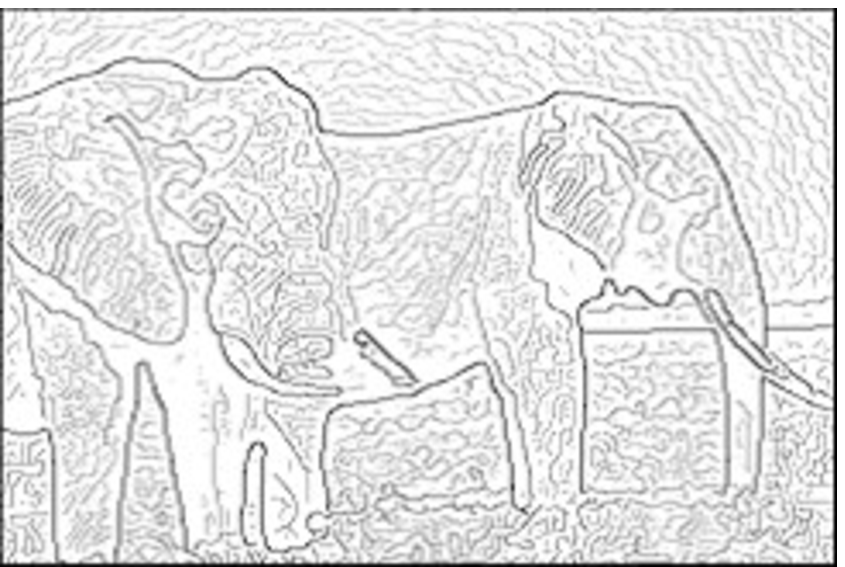}
      \end{center}
      \end{minipage}&

      \begin{minipage}{0.25\columnwidth}
      \begin{center}
      \includegraphics[width=1\textwidth,keepaspectratio=true,clip]
      {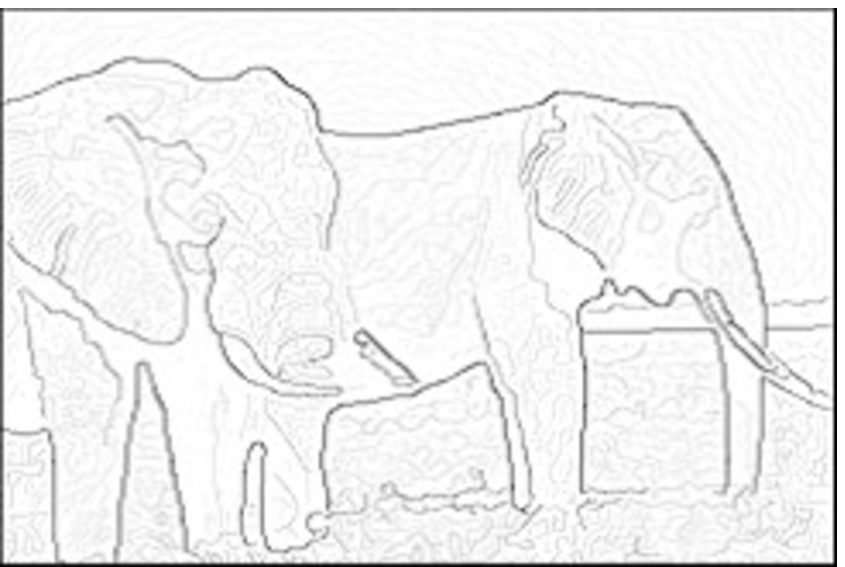}
      \end{center}
      \end{minipage}\\

      \begin{minipage}{0.25\columnwidth}
      \begin{center}
      \includegraphics[width=1\textwidth,keepaspectratio=true,clip]
      {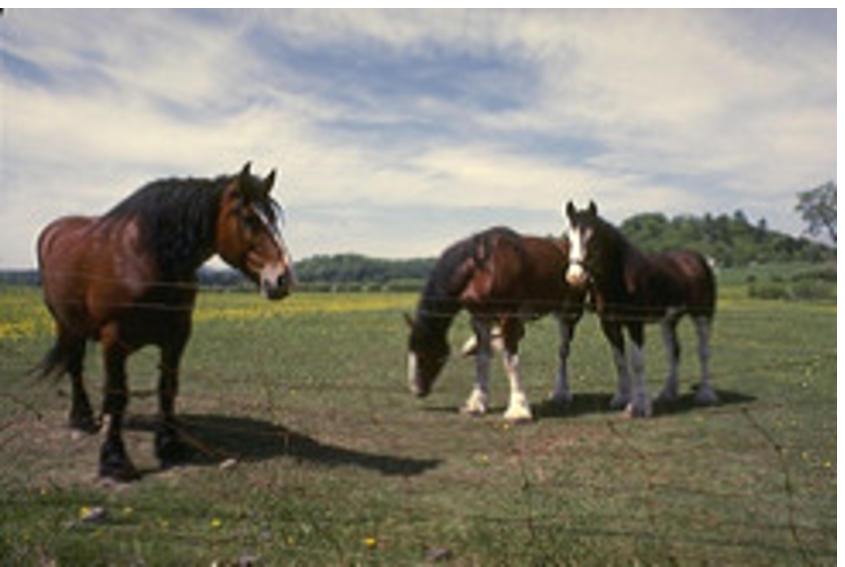}
      \end{center}
      \end{minipage}&

      \begin{minipage}{0.25\columnwidth}
      \begin{center}
      \includegraphics[width=1\textwidth,keepaspectratio=true,clip]
      {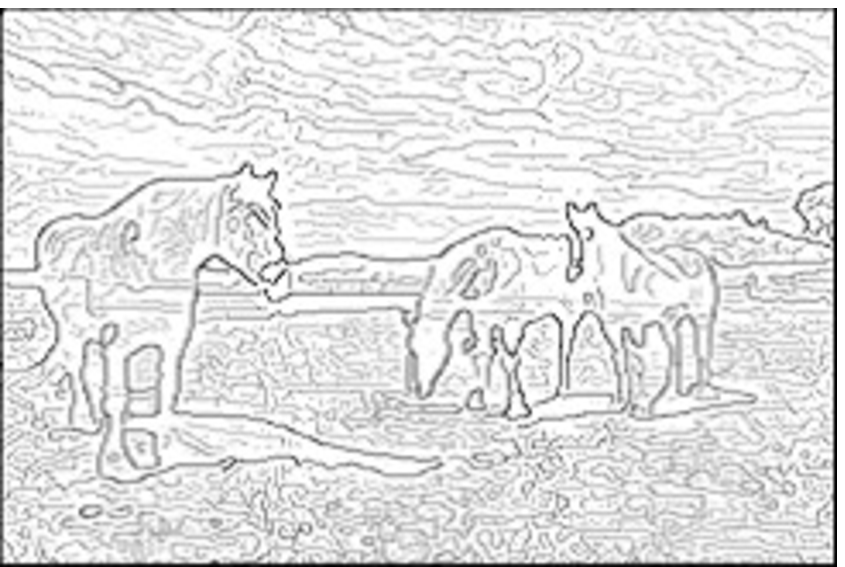}
      \end{center}
      \end{minipage}&

      \begin{minipage}{0.25\columnwidth}
      \begin{center}
      \includegraphics[width=1\textwidth,keepaspectratio=true,clip]
      {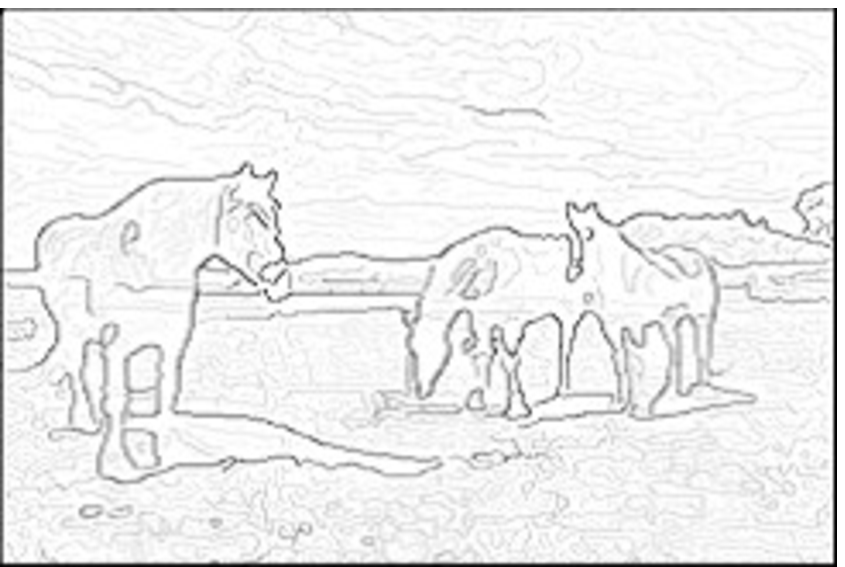}
      \end{center}
      \end{minipage}\\

      (a) & (b) & (c)
    \end{tabular}
  \end{center}
\caption{Examples of Strengthened edges. (a) Test images. (b) Canny edges. (c) Strengthened edges.
The Canny edges shown in (b) are used to compute the Strengthened edges in (c).
\label{fig:Strengthened Edges}}
\end{figure}

In our boundary detection we use the torque volume to reweigh edges according to their contribution to large torque values.
The contribution of an edge point to  torque values of different patches can be obtained  as:
\begin{eqnarray}
\upsilon_q = \sum_{\left\{P|q\in P\right\}}\tau_{p\left(P\right)q},\label{eq:edge contribution}
\end{eqnarray}
where $p\left(P\right)$ indicates the center of the patch $P$, and
$q$ is an edge point.
The  idea is the same as for the  attention mechanism. Extrema in the torque volume indicate the existence of edges surrounding the center of a patch.
Therefore, edges on object boundaries  should have a large contribution to  extrema (eq.~\ref{eq:edge contribution}).
Edges are strengthened by combining the original edge with the value of this contribution  as follows:
\begin{align}
d_s = \frac{1}{1+e^{-\left(c_0 + c_1 d_o + c_2 d_\tau \right)}},\label{eq:Edge Strengthing}
\end{align}
where $d_o$ is the original edge intensity, and $d_\tau$ is the normalized torque contribution.
The computed edge's contribution to torque is normalized into $[0,1]$.
$c_0$, $c_1$, and $c_2$ are constants.
We call the edges reweighted by the value $d_s$ in eq.~(\ref{eq:Edge Strengthing}) \emph{Strengthened edges}.
Examples of such strengthened edges are shown in Fig.~\ref{fig:Strengthened Edges}.
In this experiment,   the constant parameters $c_0$, $c_1$ and $c_2$ were set to  -2.54, 1.86 and 2.69, respectively, and  these parameter were learned using training  images.
Canny edges were used to compute the torque as shown in (b). As can be seen from (c),
the strengthened edges tend to be stronger at boundary edges of objects, while weaker at texture edges.

We used the  Berkeley dataset \cite{Martin2001} to quantitatively
 evaluate the improvement of boundary detection.
While the Canny edge method scored 0.57, the torque-based strengthened edge method using the Canny edges increased the score to 0.59 in the F-measure of the Berkeley benchmark.

\begin{figure}[htbp]
  \begin{center}
    \begin{tabular}{@{}c@{}c@{}c@{}}
      \begin{minipage}{0.33\columnwidth}
      \begin{center}
      \includegraphics[width=1\textwidth,keepaspectratio=true,clip]
      {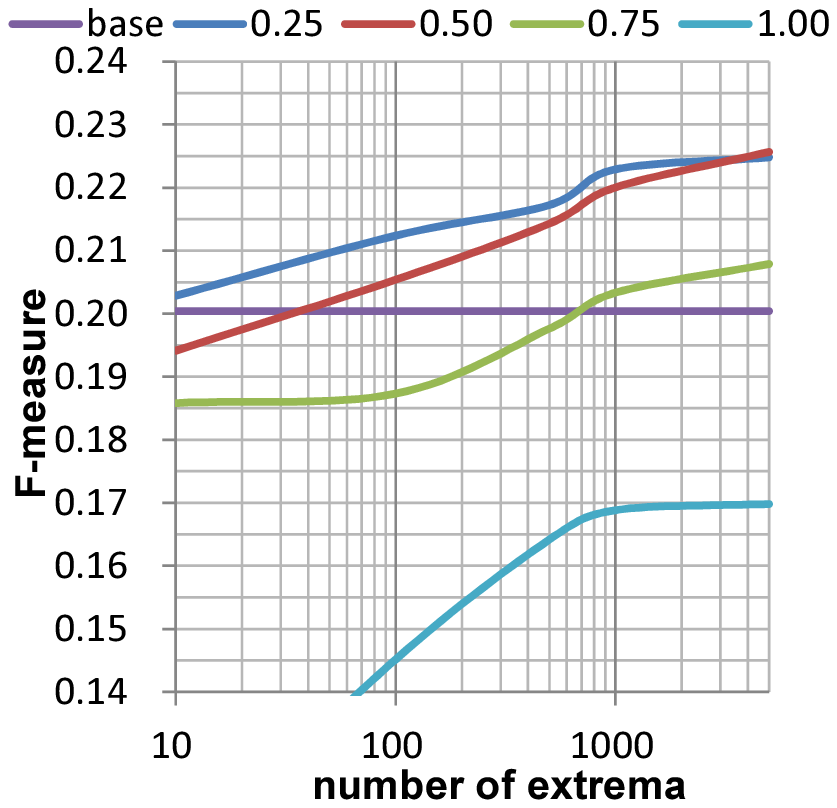}
      \end{center}
      \end{minipage}&

      \begin{minipage}{0.33\columnwidth}
      \begin{center}
      \includegraphics[width=1\textwidth,keepaspectratio=true,clip]
      {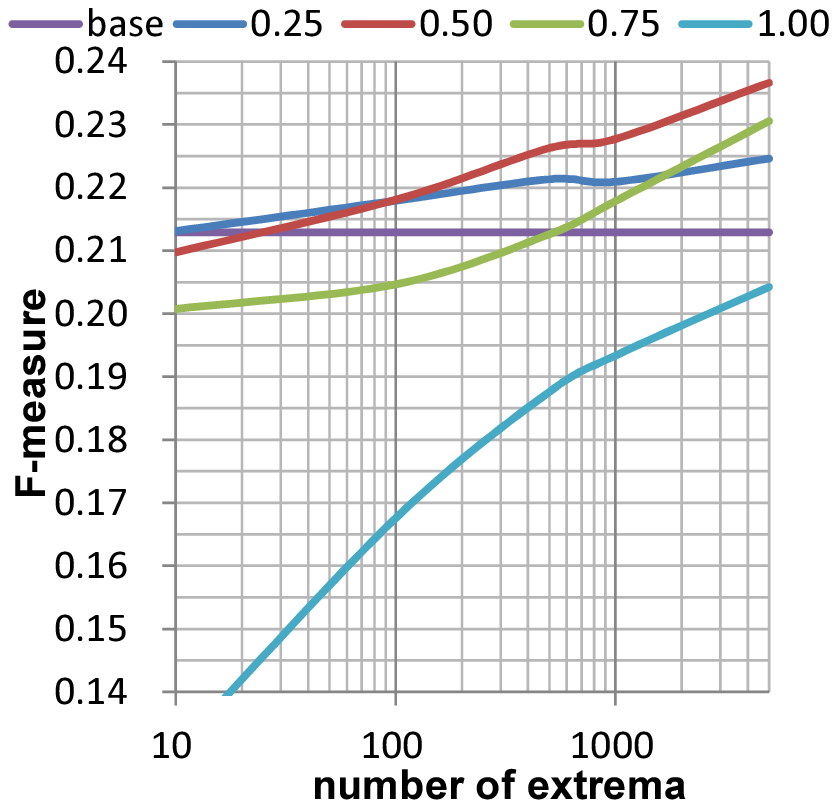}
      \end{center}
      \end{minipage}&

      \begin{minipage}{0.33\columnwidth}
      \begin{center}
      \includegraphics[width=1\textwidth,keepaspectratio=true,clip]
      {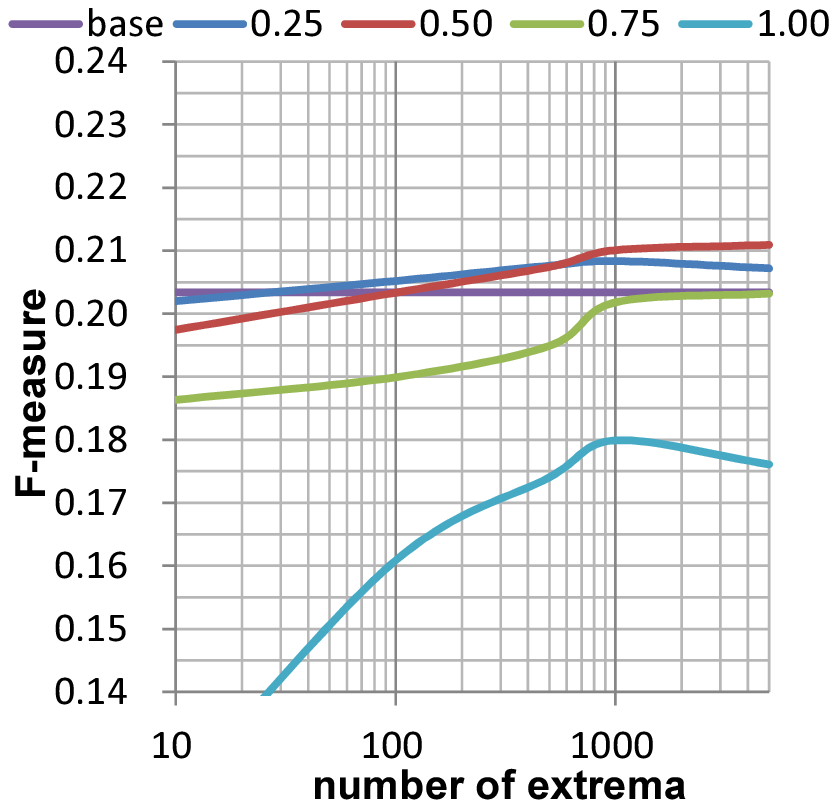}
      \end{center}
      \end{minipage}\\

      (a) Canny & (b) pb & (c) gPb
    \end{tabular}
  \end{center}
\caption{Performance of boundary detection evaluated by the maximum F-measure as a function of $\alpha$ and the number of extrema. Three edge detection methods were used to compute base edge maps: (a) Canny edges, (b) pb edges, and (c) gPb edges. The base edge maps were blended with the torque contribution map to generate strengthened edge maps.
In the legend \emph{base} refers to the base edge maps, and the numbers  indicate the weight $\alpha$ in eq.(\ref{eq:Edge Strengthing2}) used to obtain strengthened edge maps.
\label{fig:Performance of Boundary Detection and Number of Extrema}}
\end{figure}

\begin{figure}[htbp]
\caption{Precision-recall (PR) curves for boundary detection. The torque-based strengthened edge map (green ) is is compared to the base edge map (gray) for: (a) Canny edges, (b) pb edges, (c) gPb edges, and (d) Sketch tokens edges.
\label{fig:Boundary detection PR curve}}
  \begin{center}
    \begin{tabular}{cc}
      \begin{minipage}{0.4\columnwidth}
      \begin{center}
      \includegraphics[width=1\textwidth,keepaspectratio=true,clip]
      {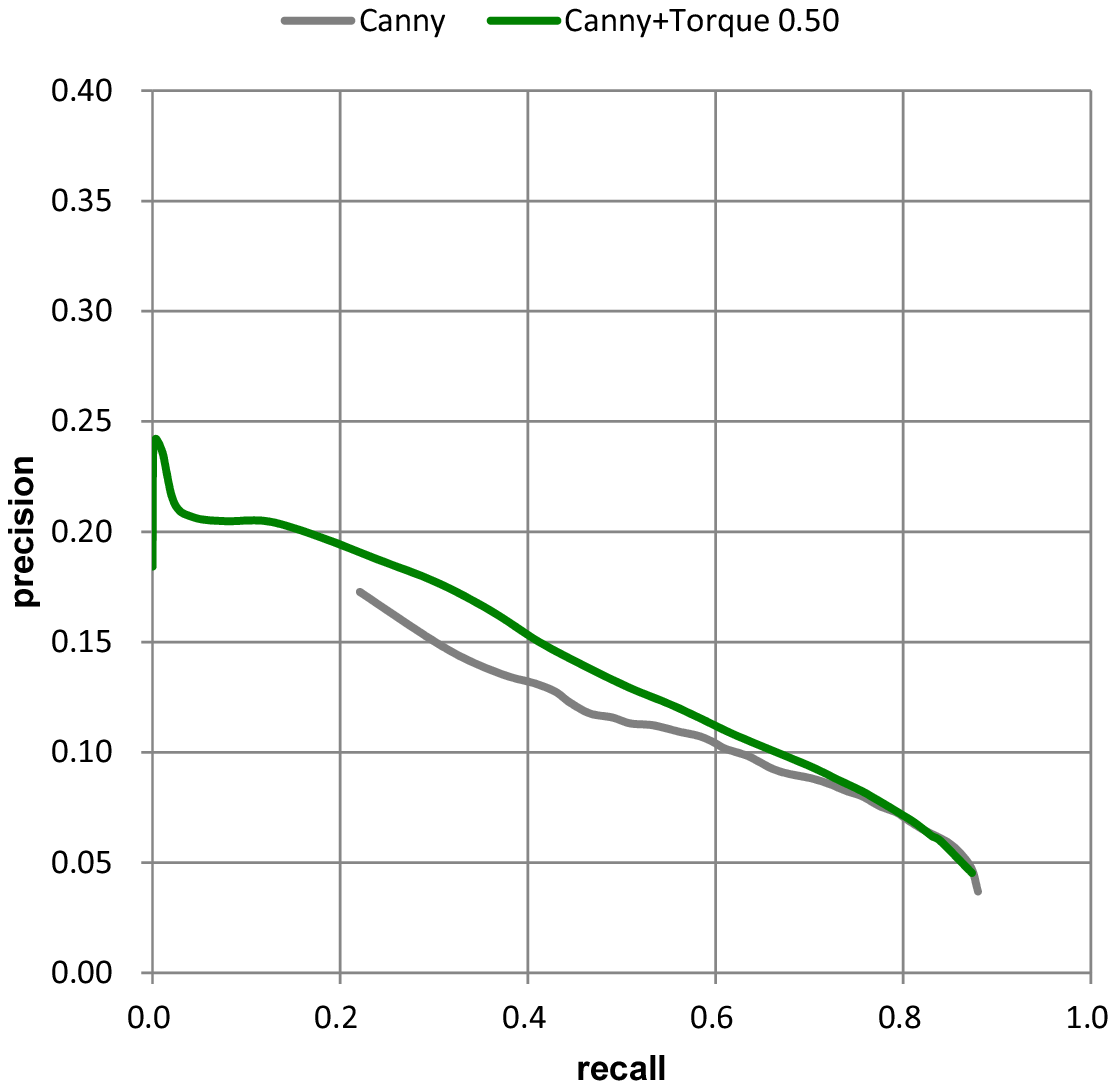}
      \end{center}
      \end{minipage}&

      \begin{minipage}{0.4\columnwidth}
      \begin{center}
      \includegraphics[width=1\textwidth,keepaspectratio=true,clip]
      {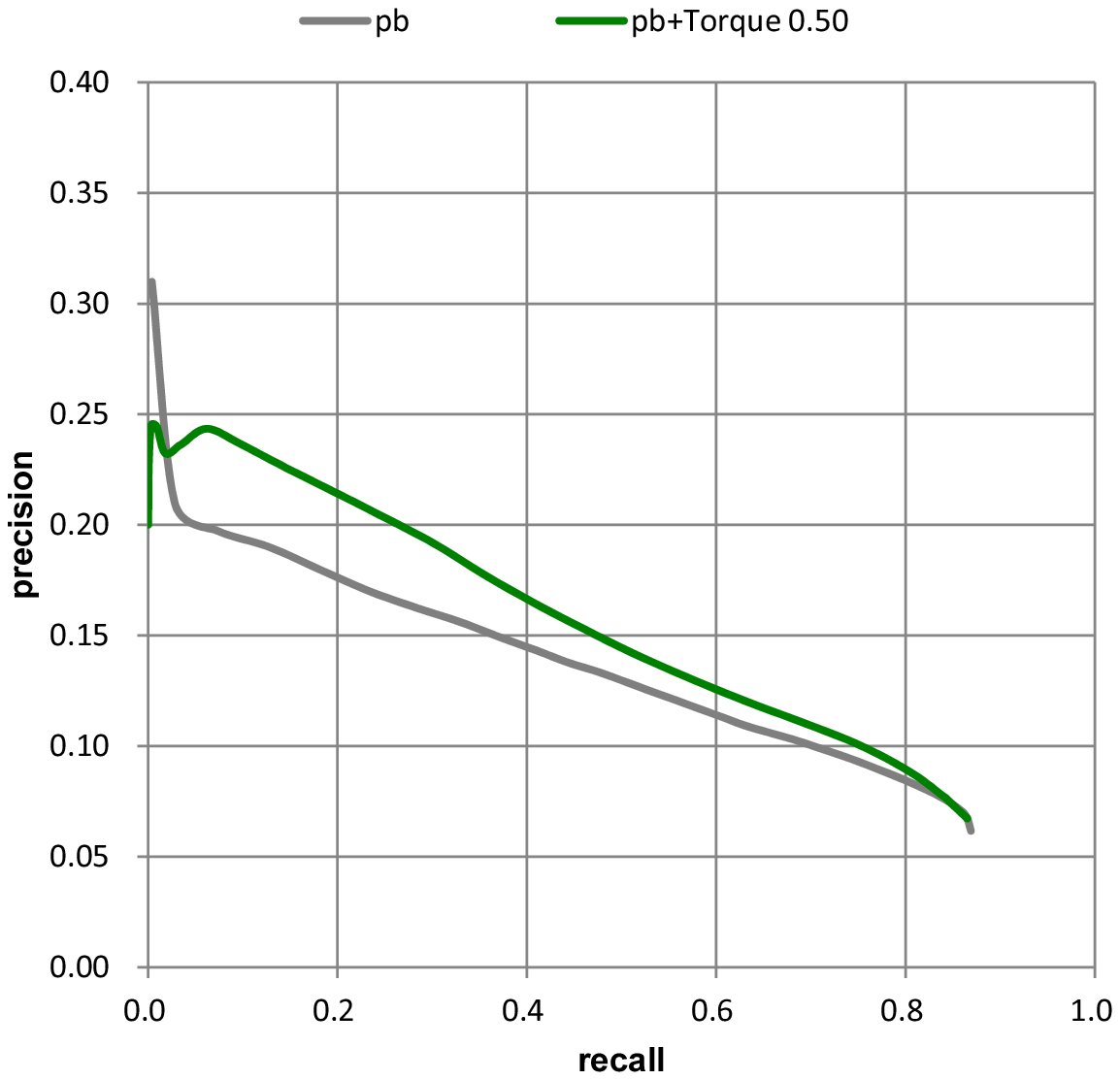}
      \end{center}
      \end{minipage}\\

      (a) Canny & (b) pb\\

      \begin{minipage}{0.4\columnwidth}
      \begin{center}
      \includegraphics[width=1\textwidth,keepaspectratio=true,clip]
      {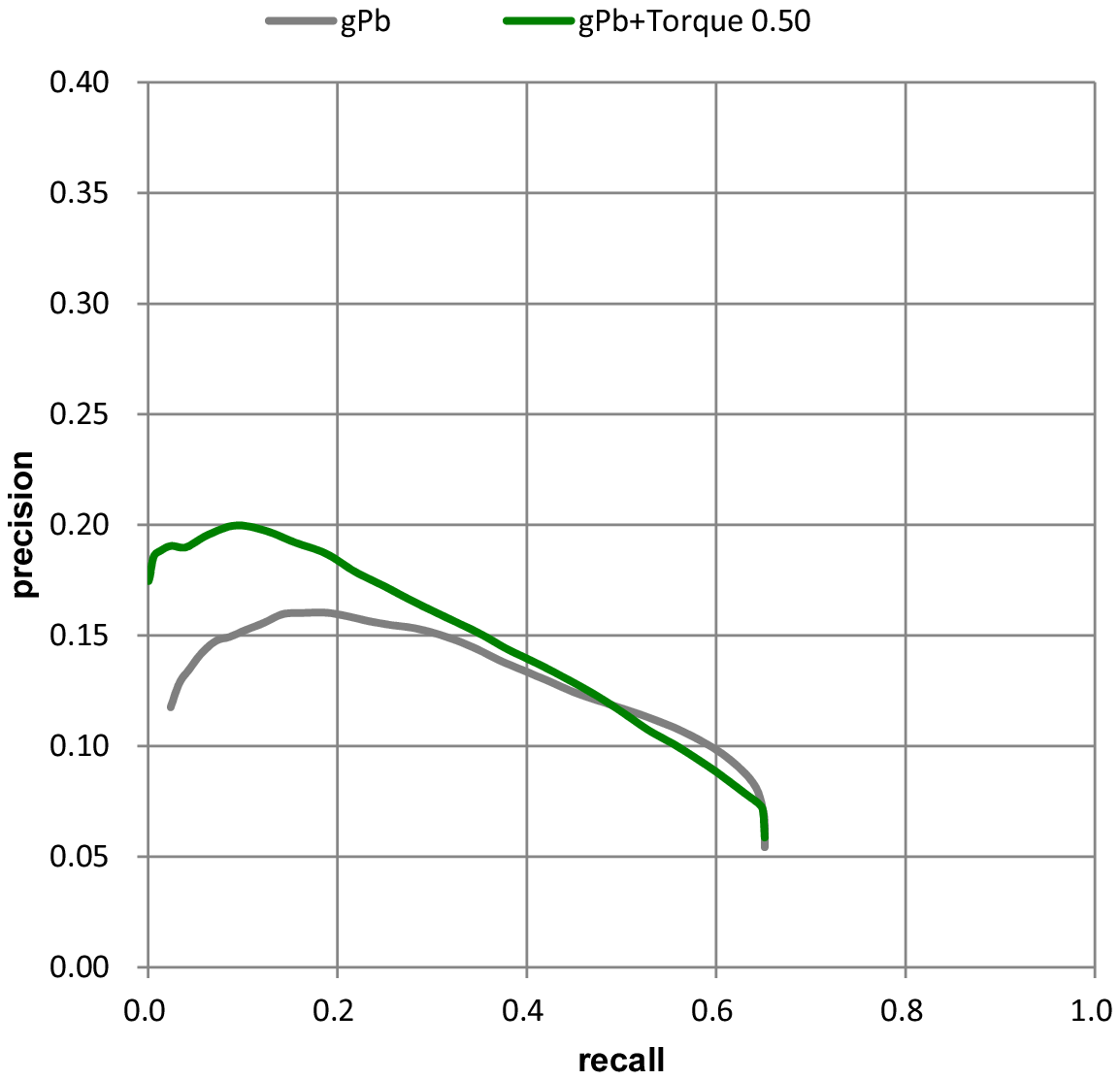}
      \end{center}
      \end{minipage}&

      \begin{minipage}{0.4\columnwidth}
      \begin{center}
      \includegraphics[width=1\textwidth,keepaspectratio=true,clip]
      {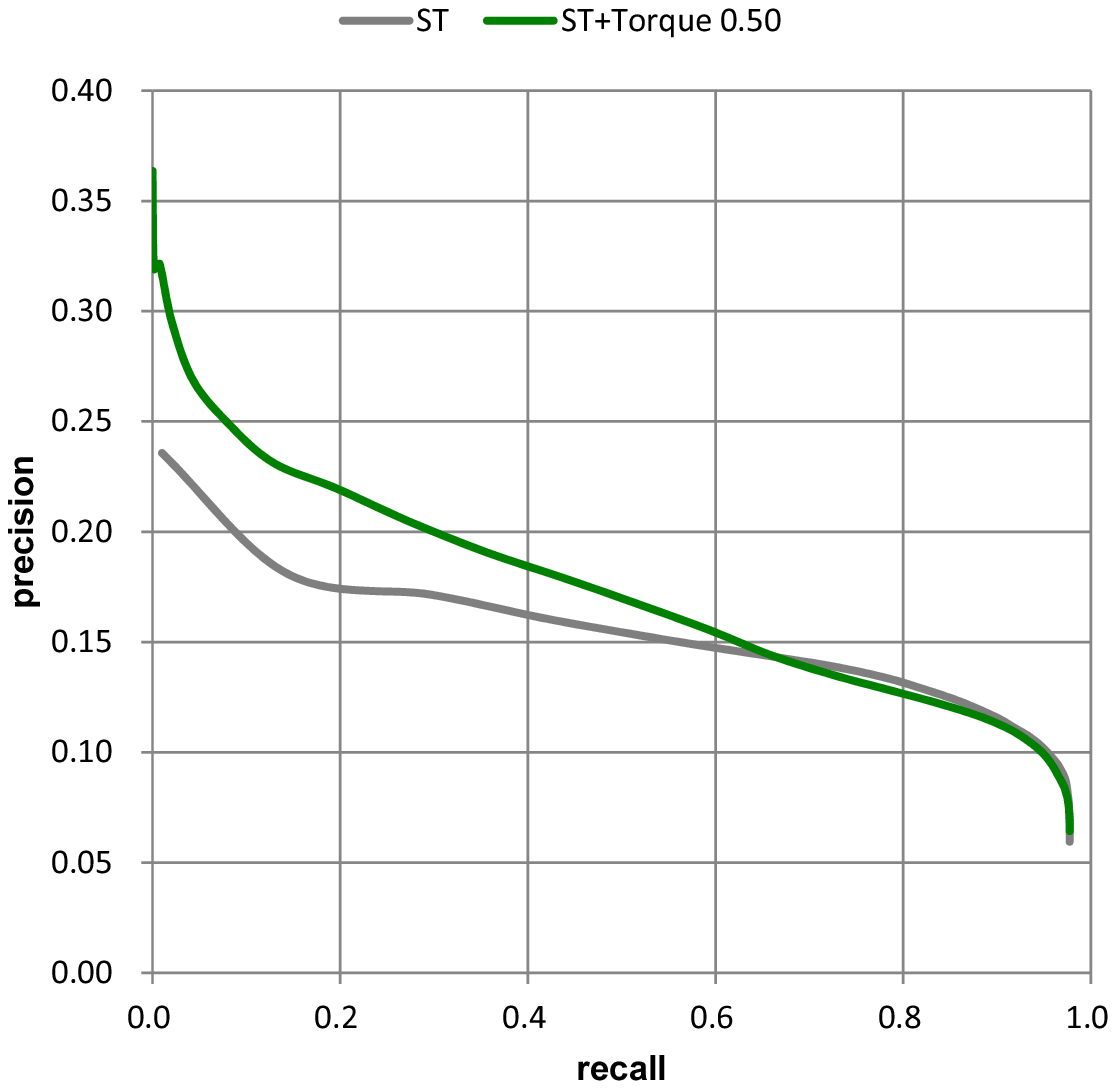}
      \end{center}
      \end{minipage}\\

      (c) gPb & (d) Sketch Tokens
    \end{tabular}
  \end{center}
\end{figure}

In a second experiment we evaluated the torque on a  dataset that is focused on objects.
We used the images and boundary annotations for the \emph{car side} category in the Caltech dataset \cite{FeiFei2006}. Edge maps were computed  using  Canny, pb boundary  \cite{Martin2001}, and gPb boundary detection \cite{Arbelaez2011}, and these edge maps were used to derive  the torque.
Then the torque-based strengthened edges were  computed and combined with the base  edge maps. Here we  simply used a weighted sum to combine the two terms as follows:
\begin{align}
d_s = (1-\alpha) d_o + \alpha d_\tau,\label{eq:Edge Strengthing2}
\end{align}
where  $d_o$ denotes the edge map, $d_\tau$  the  torque contribution map, and
$d_s$ the strengthened edge map.
We computed Precision and Recall for   the strengthened edge maps $d_s$ and the base edge maps $d_o$, and
 evaluated the maximum F-measure  as a function of parameter $\alpha$ and the number of torque extrema in the computation of the torque contribution map, as shown in Fig~\ref{fig:Performance of Boundary Detection and Number of Extrema}.
Based on this evaluation, $\alpha$ was set to  0.5.
In addition, we evaluated  the recently proposed method of Sketch Tokens \cite{Lim2013} for boundary detection and the effect of  edge strengthening on this method.
The precision-recall (PR) curve for all four methods are shown in Fig~\ref{fig:Boundary detection PR curve}.
Comparing in Fig.~\ref{fig:Performance of Boundary Detection and Number of Extrema} the maximum F-measure of the strengthened edge maps with  the base edge maps, we can verify that the torque operator  improves silhouette boundary detection. Table~\ref{tbl:F-masure comparison in boundary detection}  summarizes the results for $\alpha=0.5$ and 5000 torque extrema for all four methods.

\begin{table}[tb]

\begin{center}
\begin{tabular}{|c|c|}

	\hline
	Method & F-measure\\
	\hline
	Canny & 0.20\\
	Torque (Canny) & {\bfseries 0.23}\\
	\hline
	pb & 0.21\\
	Torque (pb) & {\bfseries 0.24}\\
	\hline
	gPb & 0.20\\
	Torque (gPb) & {\bfseries 0.21}\\
	\hline
	Sketch Tokens & 0.24\\
	Torque (Sketch Tokens) & {\bfseries 0.25}\\
	\hline

\end{tabular}
\end{center}
\caption{F-measure comparison in boundary detection between base edge maps and torque-based strengthened edge maps.
\label{tbl:F-masure comparison in boundary detection}}
\end{table}

\subsection{Segmentation}

Next the torque operator is demonstrated  for segmentation in two graphcut approaches, and a quantitative comparison to other methods is given.
The first approach is a generic multi-region segmentation, and the second is a
figure-ground segmentation.  Both approaches basically rely on the torque as attention and scale selection mechanism. The attended region is then  utilized to obtain  foreground color models or to strengthen  edges.

\subsubsection{Multi-region Segmentation}

\begin{figure}[tb]
  \begin{center}
    \begin{tabular}[t]{@{}c@{\,}c@{\,}c@{}}

%
%
%

      \begin{minipage}{0.23\columnwidth}
      \begin{center}
      \includegraphics[width=1\textwidth,keepaspectratio=true,clip]
      {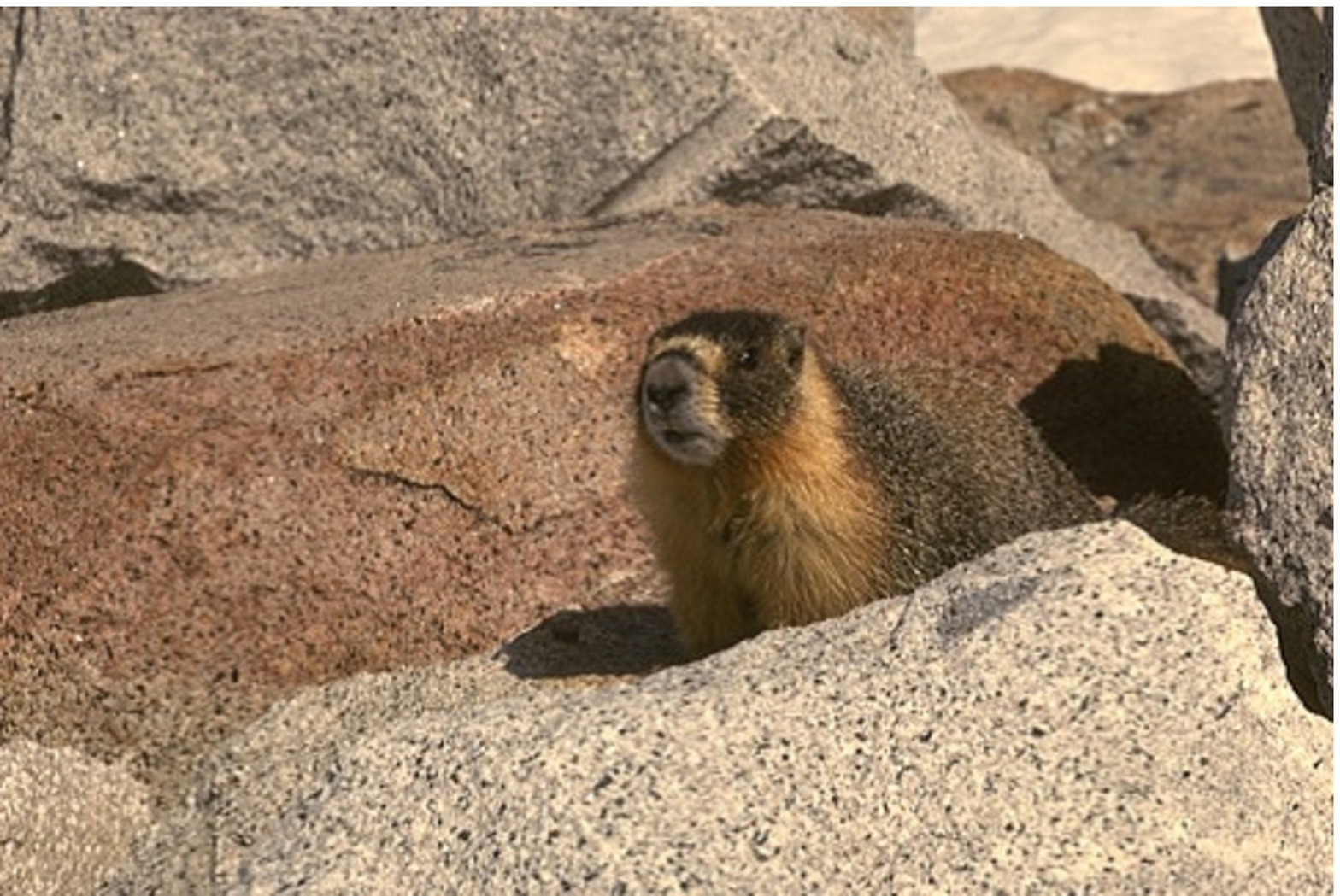}
      \end{center}
      \end{minipage}&

      \begin{minipage}{0.23\columnwidth}
      \begin{center}
      \includegraphics[width=1\textwidth,keepaspectratio=true,clip]
      {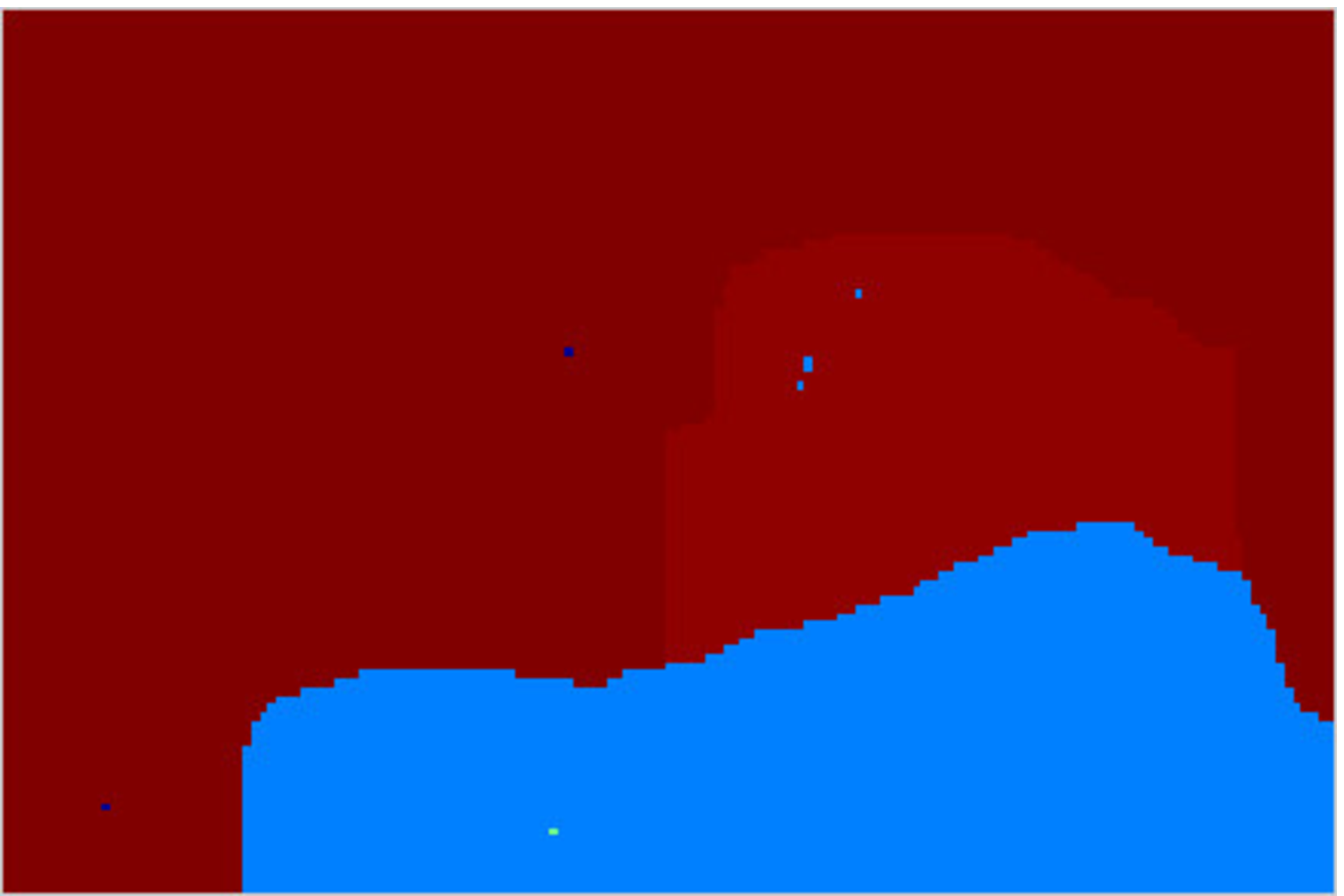}
      \end{center}
      \end{minipage}&

      \begin{minipage}{0.23\columnwidth}
      \begin{center}
      \includegraphics[width=1\textwidth,keepaspectratio=true,clip]
      {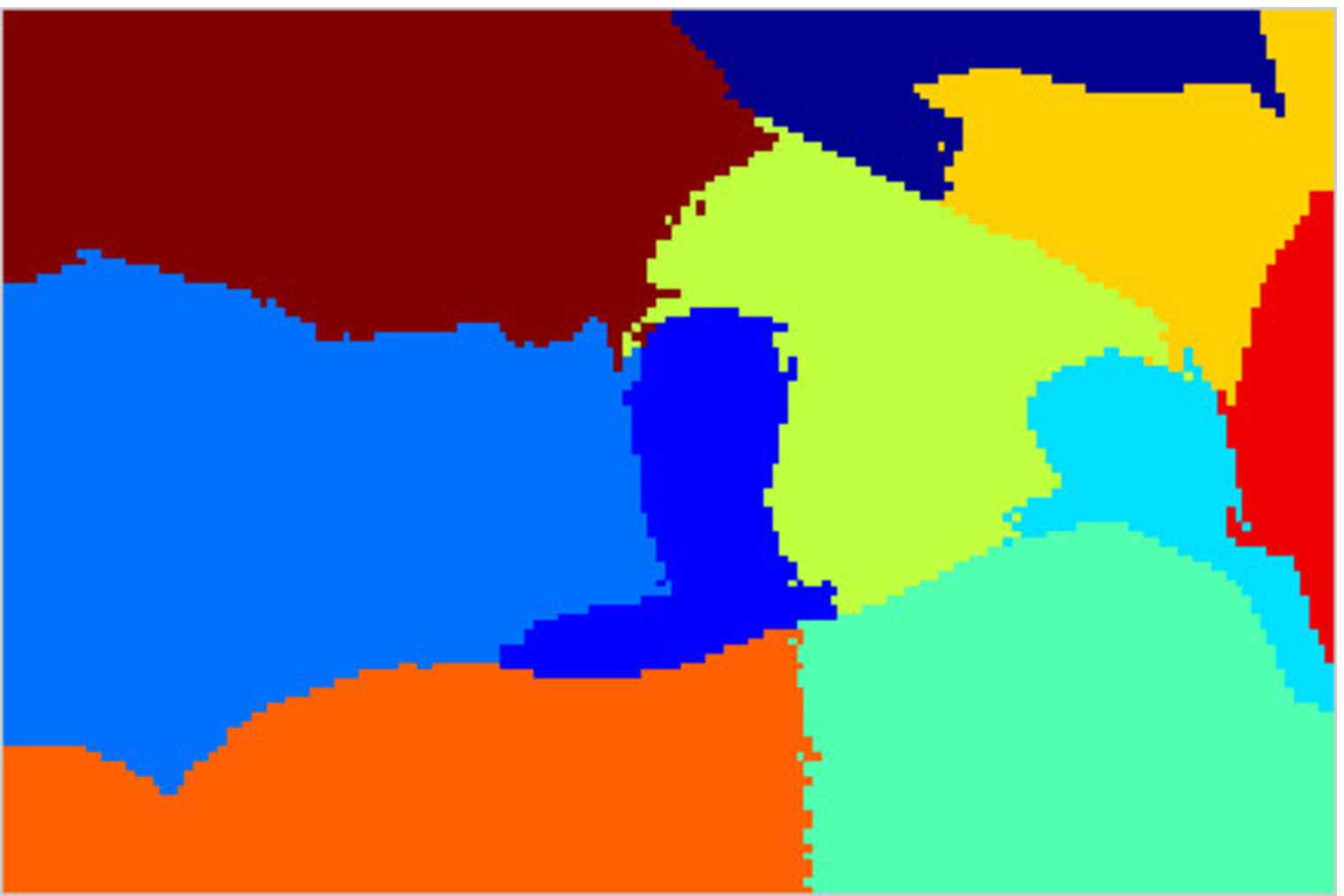}
      \end{center}
      \end{minipage}\\

      \begin{minipage}{0.23\columnwidth}
      \begin{center}
      \includegraphics[width=1\textwidth,keepaspectratio=true,clip]
      {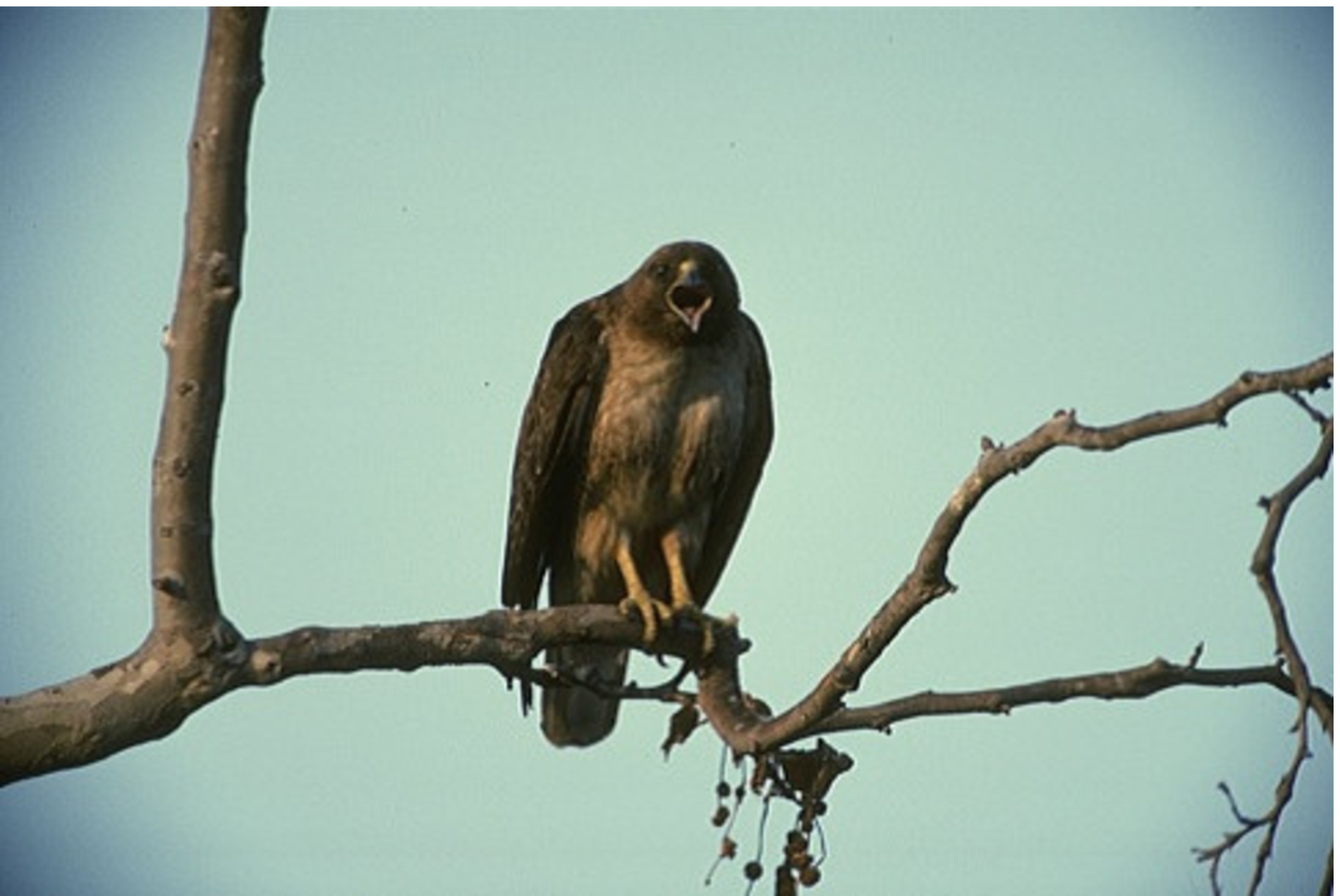}
      \end{center}
      \end{minipage}&

      \begin{minipage}{0.23\columnwidth}
      \begin{center}
      \includegraphics[width=1\textwidth,keepaspectratio=true,clip]
      {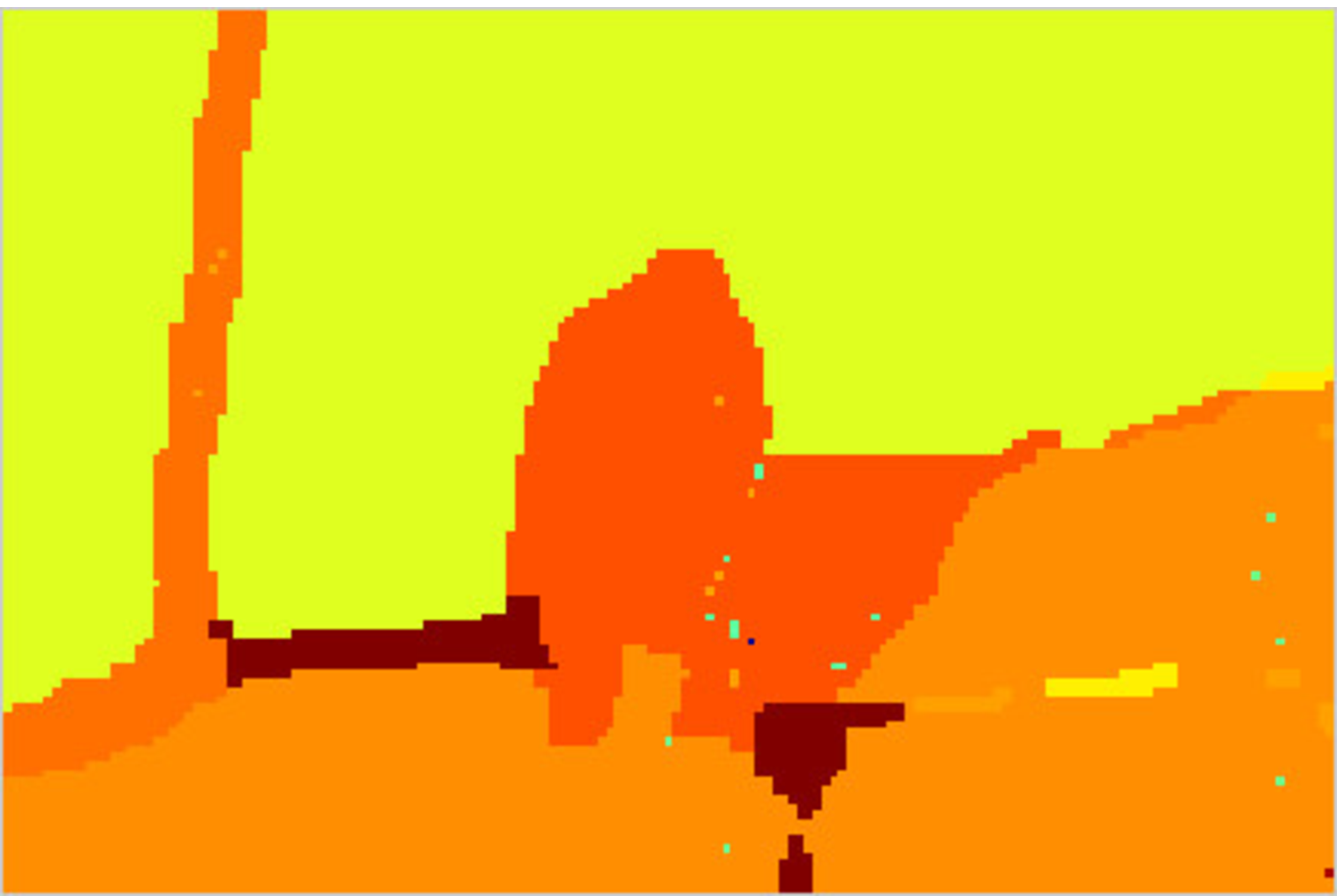}
      \end{center}
      \end{minipage}&

      \begin{minipage}{0.23\columnwidth}
      \begin{center}
      \includegraphics[width=1\textwidth,keepaspectratio=true,clip]
      {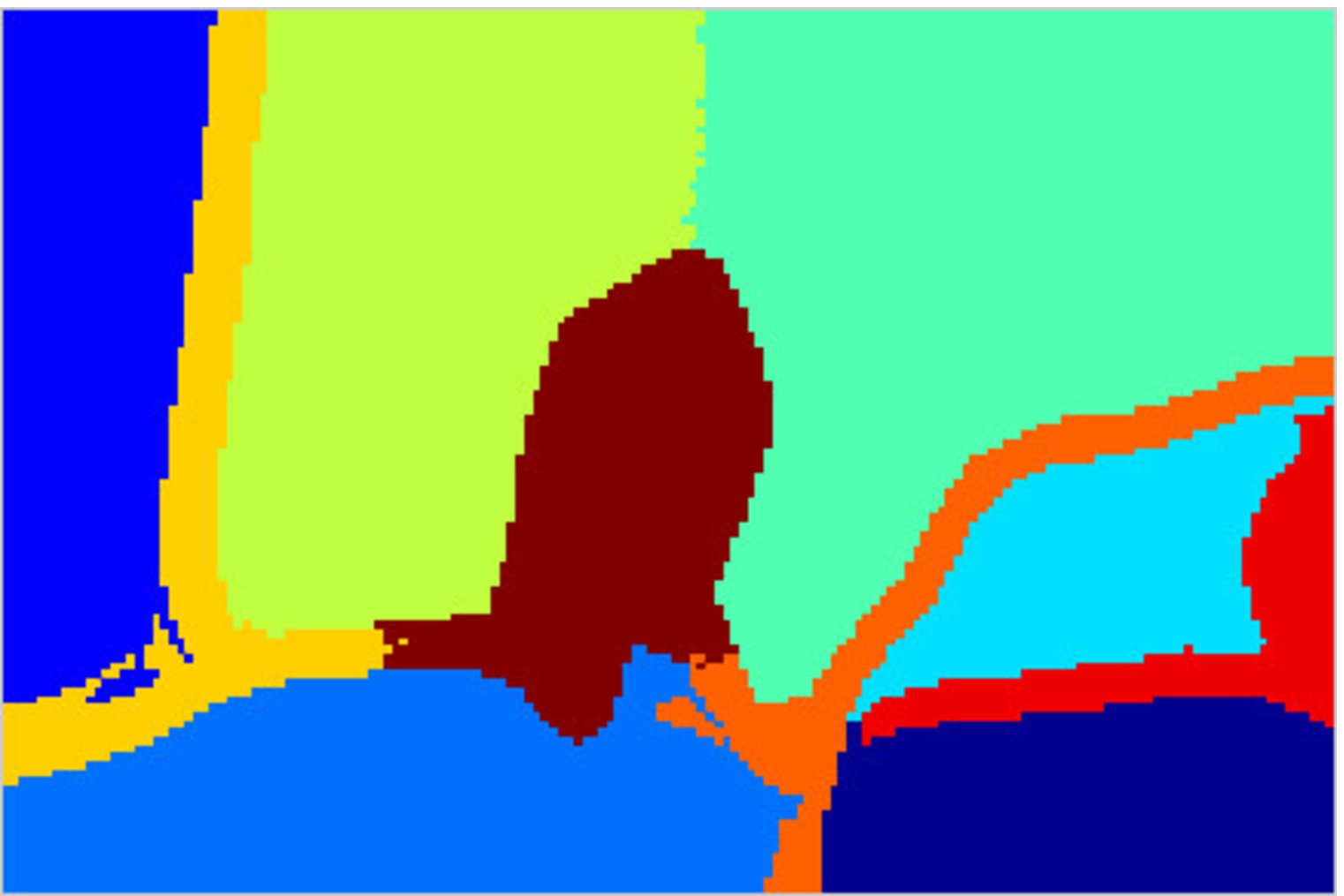}
      \end{center}
      \end{minipage}\\

%
%

      \begin{minipage}{0.23\columnwidth}
      \begin{center}
      \includegraphics[width=1\textwidth,keepaspectratio=true,clip]
      {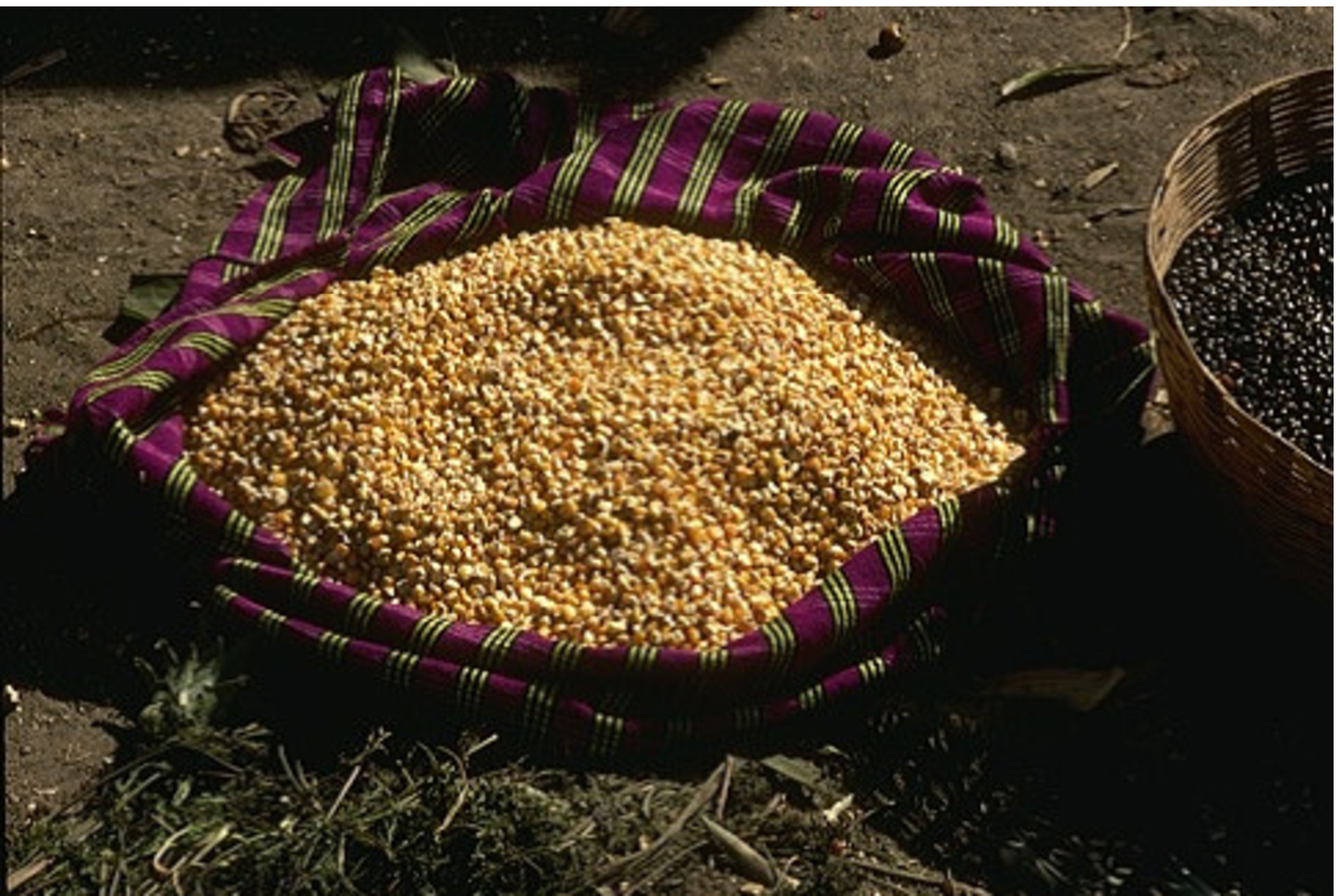}
      \end{center}
      \end{minipage}&

      \begin{minipage}{0.23\columnwidth}
      \begin{center}
      \includegraphics[width=1\textwidth,keepaspectratio=true,clip]
      {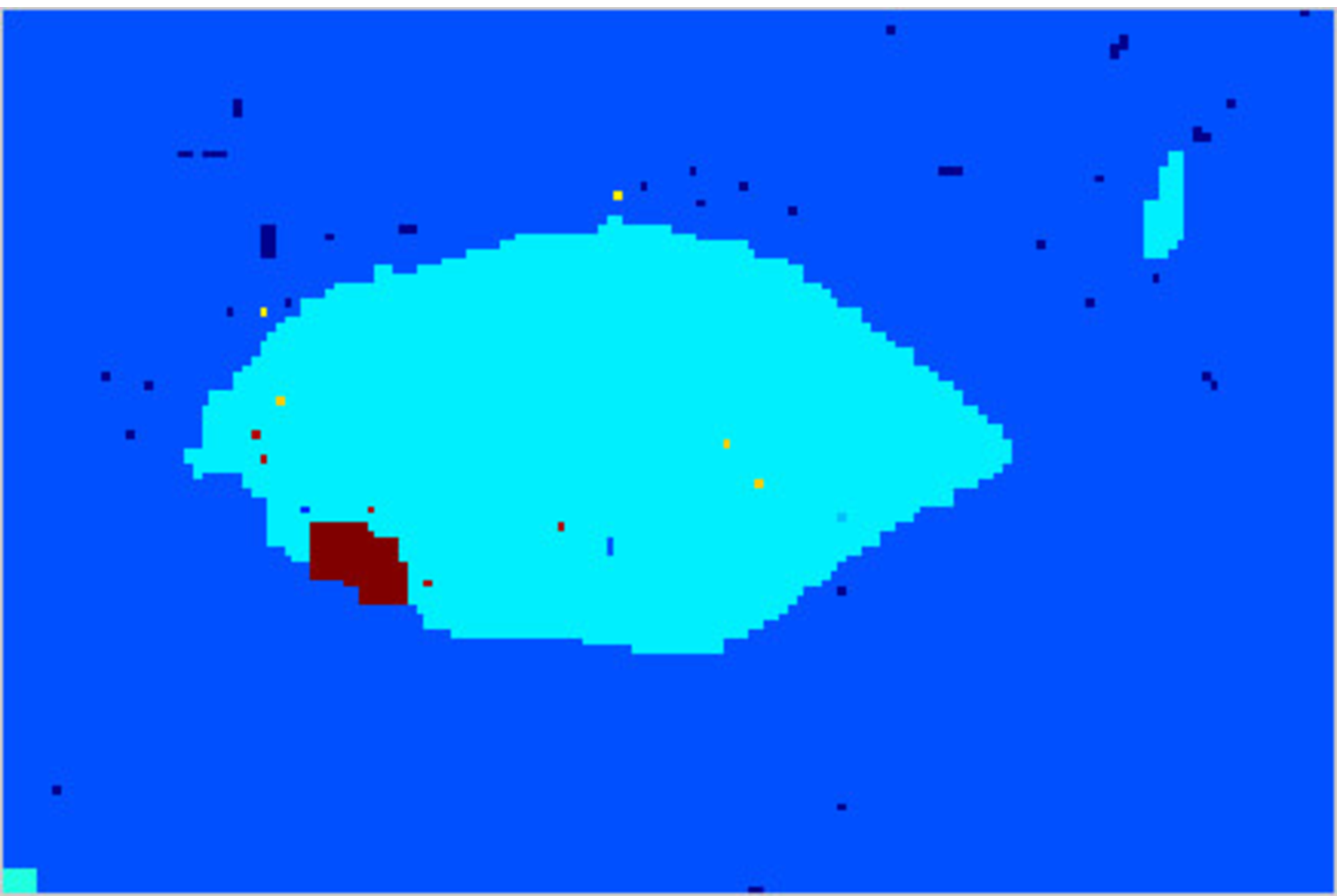}
      \end{center}
      \end{minipage}&

      \begin{minipage}{0.23\columnwidth}
      \begin{center}
      \includegraphics[width=1\textwidth,keepaspectratio=true,clip]
      {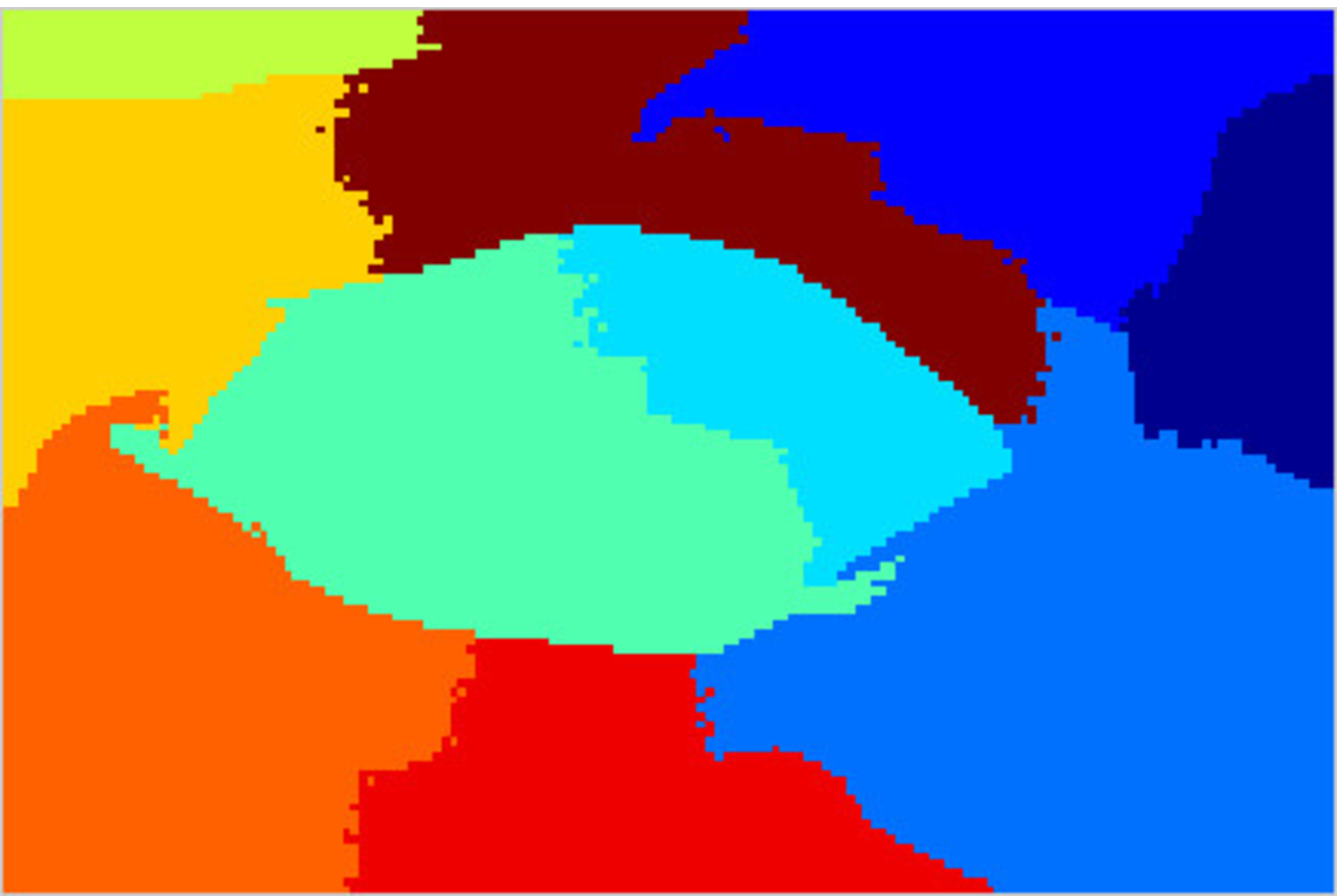}
      \end{center}
      \end{minipage}\\

      \begin{minipage}{0.23\columnwidth}
      \begin{center}
      \includegraphics[width=1\textwidth,keepaspectratio=true,clip]
      {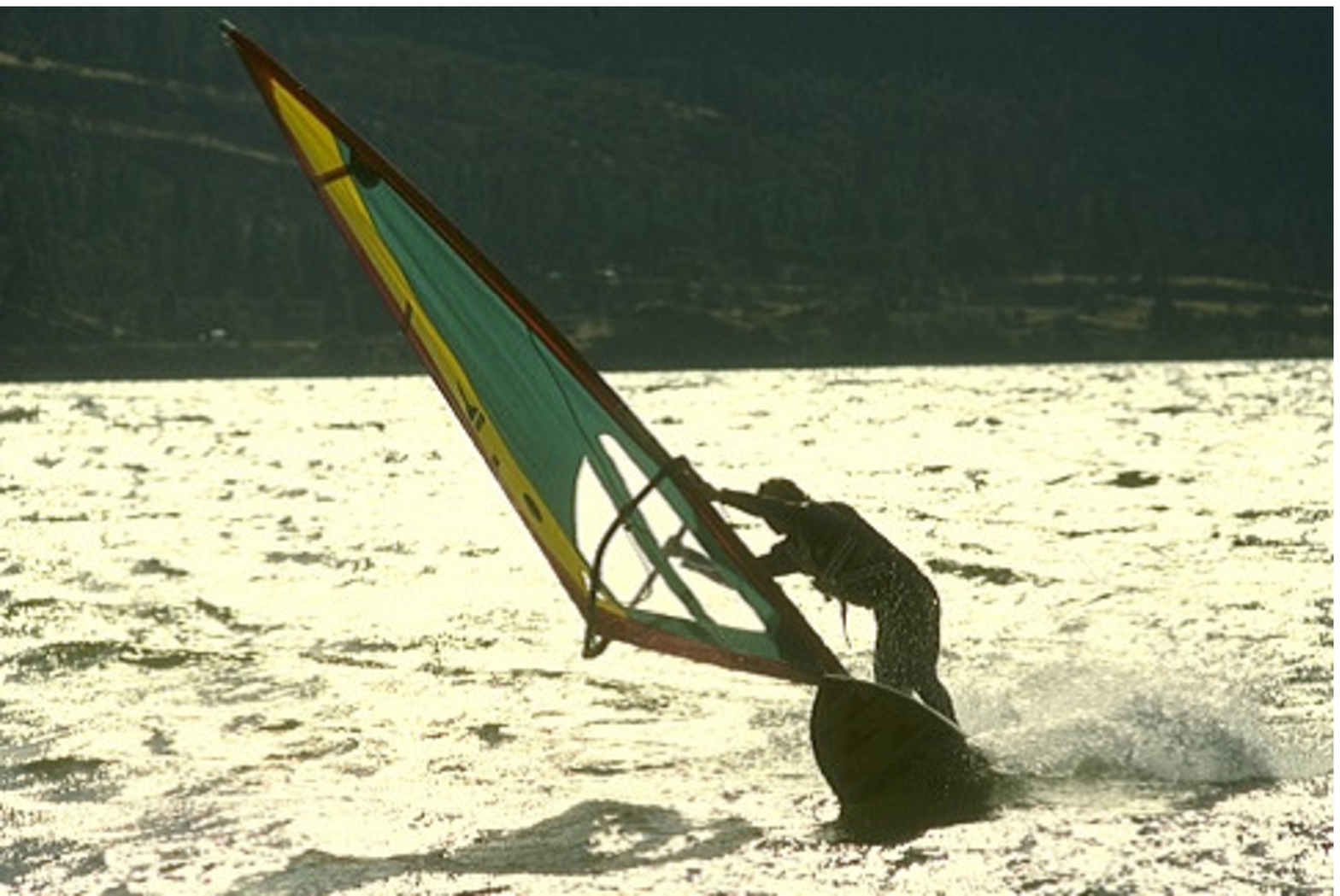}
      \end{center}
      \end{minipage}&

      \begin{minipage}{0.23\columnwidth}
      \begin{center}
      \includegraphics[width=1\textwidth,keepaspectratio=true,clip]
      {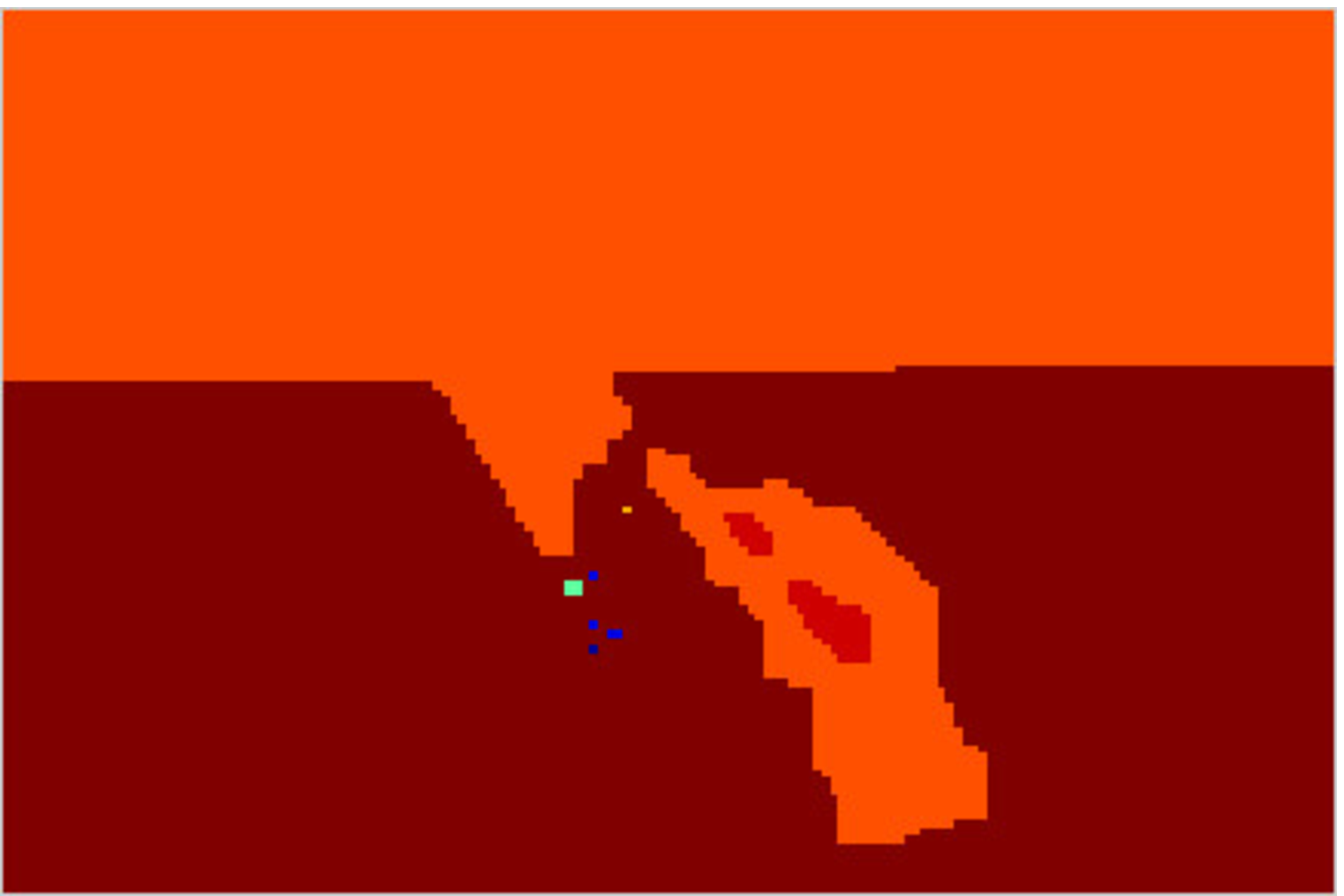}
      \end{center}
      \end{minipage}&

      \begin{minipage}{0.23\columnwidth}
      \begin{center}
      \includegraphics[width=1\textwidth,keepaspectratio=true,clip]
      {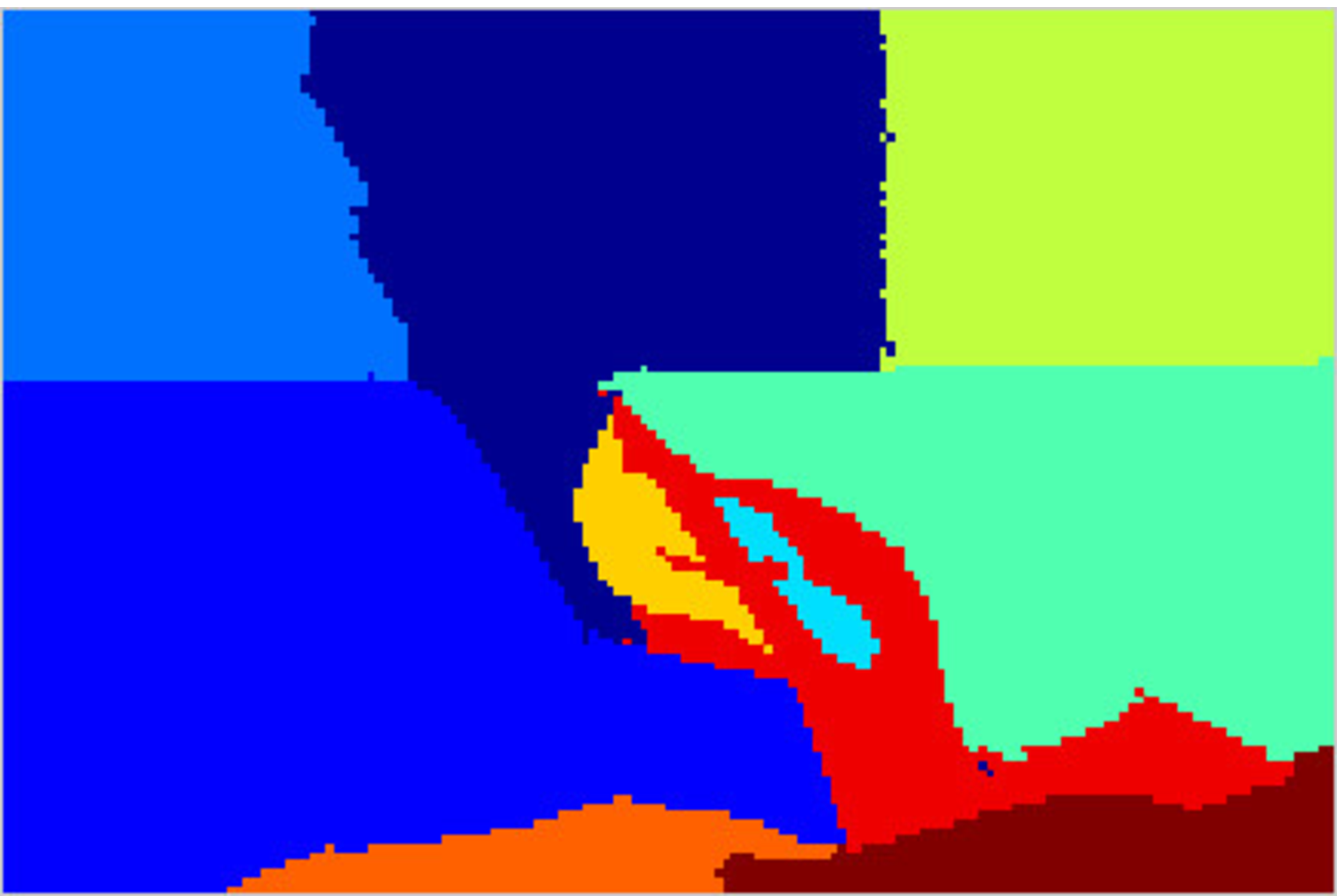}
      \end{center}
      \end{minipage}\\

      \begin{minipage}{0.23\columnwidth}
      \begin{center}
      \includegraphics[width=1\textwidth,keepaspectratio=true,clip]
      {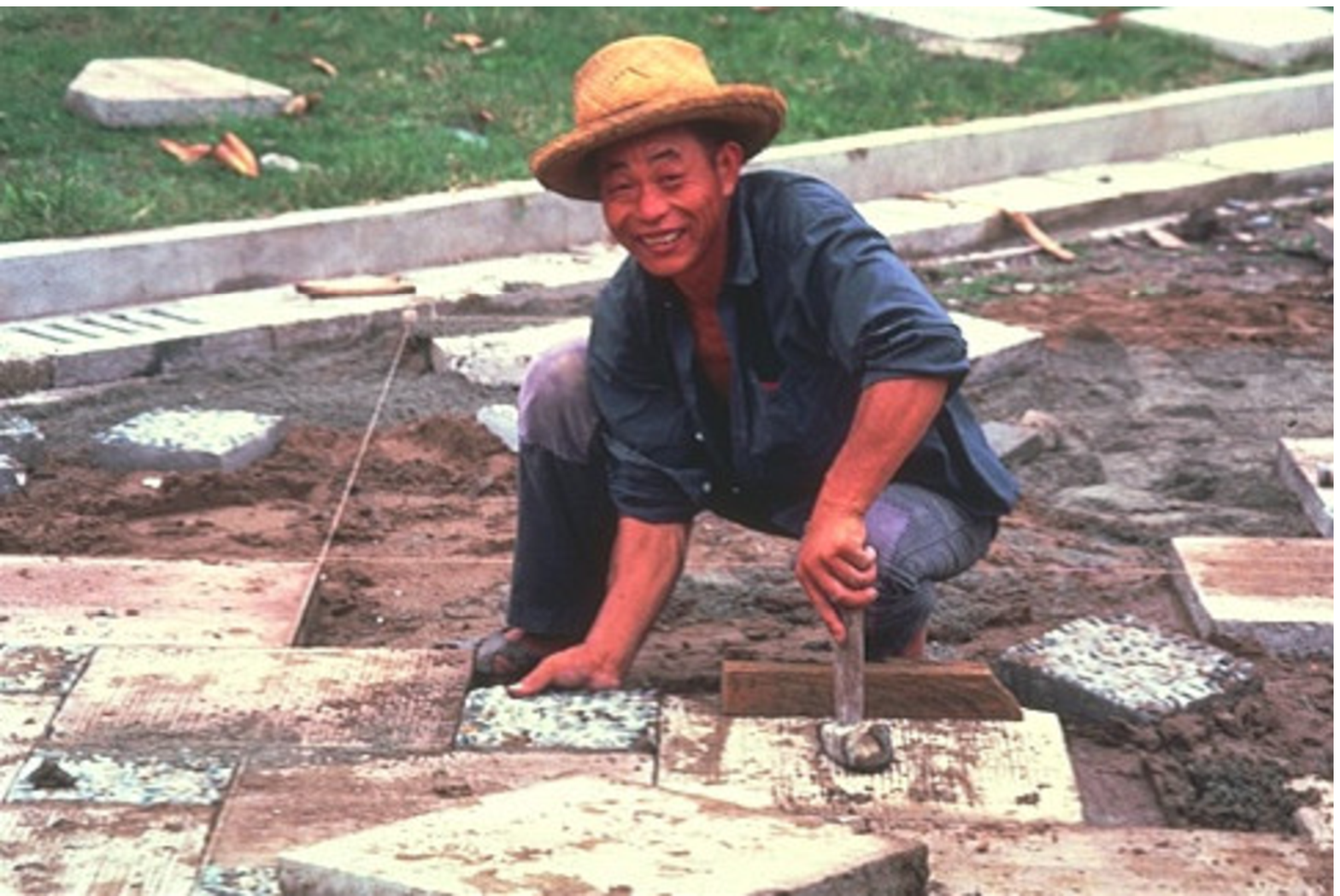}
      \end{center}
      \end{minipage}&

      \begin{minipage}{0.23\columnwidth}
      \begin{center}
      \includegraphics[width=1\textwidth,keepaspectratio=true,clip]
      {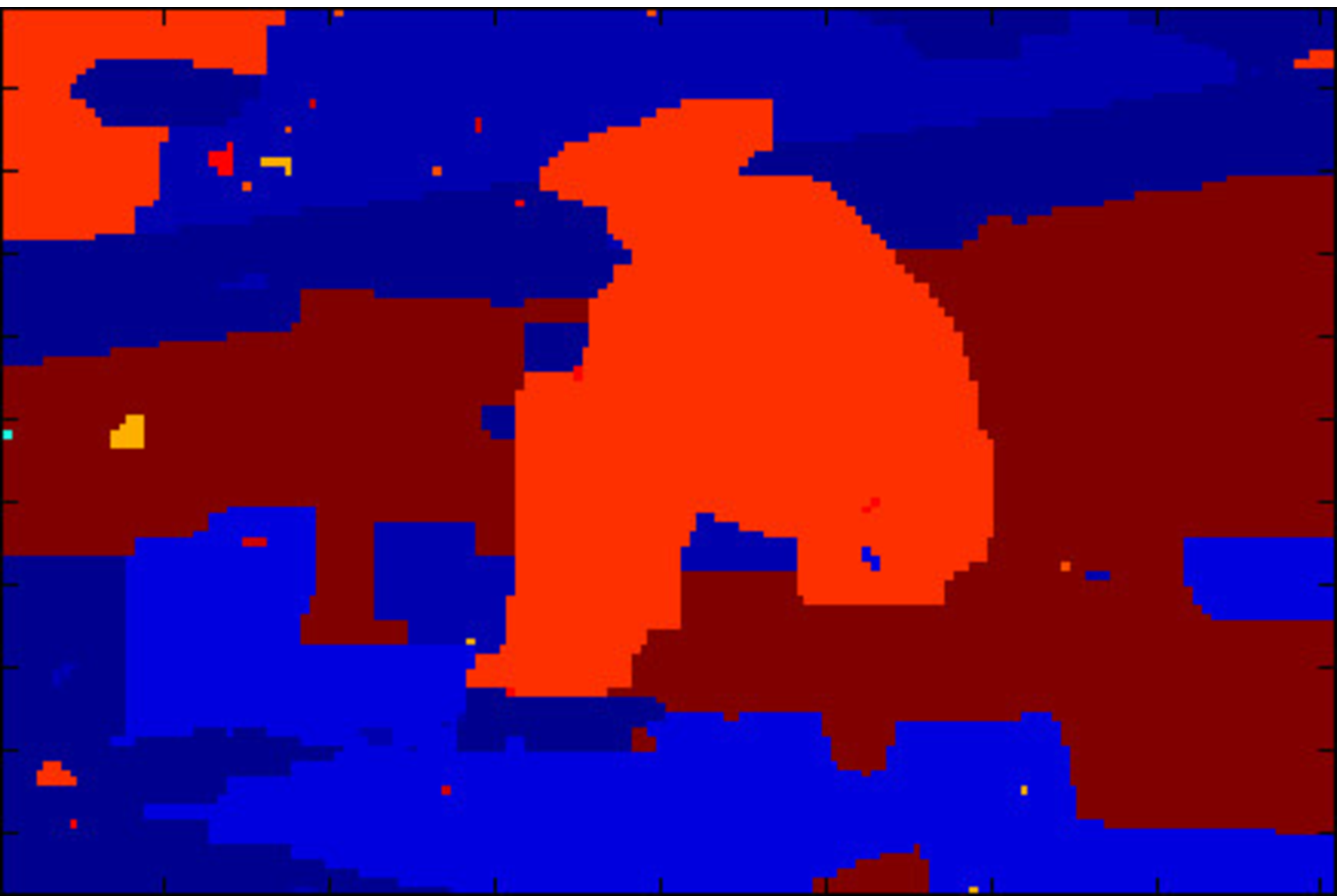}
      \end{center}
      \end{minipage}&

      \begin{minipage}{0.23\columnwidth}
      \begin{center}
      \includegraphics[width=1\textwidth,keepaspectratio=true,clip]
      {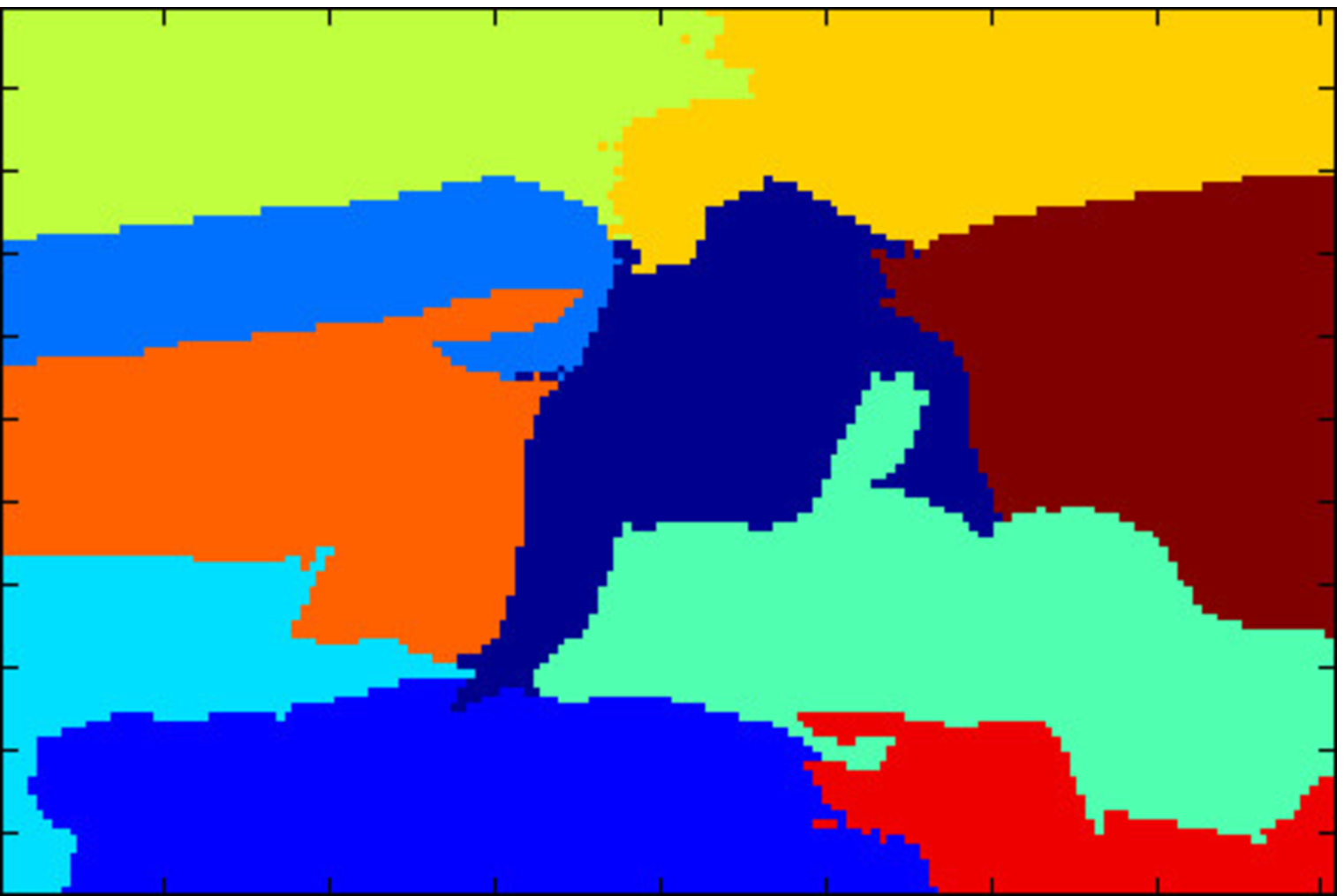}
      \end{center}
      \end{minipage}\\

      (a) & (b) & (c)
    \end{tabular}
  \end{center}
  \caption{Examples of generic multi-region segmentation: (a)  Test images. (b) Segmentation using the torque operator (c) Segmentation using the normalized cut.
\label{fig:Examples of Generic Segmentations}}
\end{figure}

We used the torque extrema and their corresponding scales to obtain  image regions  likely to correspond to interesting  elements of the image.
In this experiment, each such image region is modeled by a color histogram, and these histograms are used to create the weights in a multi-label graph-cut segmentation.
The data term in  the graph-cut is derived from  how well the color at a pixel matches  each color model, and the smoothness term is based on   color similarity of  adjacent pixels.
The segmentation method was applied to the Berkeley image data set and the quality of the segmentation was evaluated using the  covering criteria \cite{Arbelaez2009}. We compared against the normalized cut segmentation \cite{Shi2000}. 
Fig.~\ref{fig:Examples of Generic Segmentations} shows example segmentations of the two  segmentation methods. A visual evaluation shows, that the  graphcut method better segments than the normalized cut, in the sense of being able to better extract object-like regions, and  the  quantitative evaluation demonstrates  that the graphcut method clearly outperforms the normalized cut (Table 2).

\begin{table}[tb]
\caption{Comparison  of multi-region segmentations  based on the  covering criteria.}
\label{tbl:Covering of Multi-region Segmentation}
\begin{center}
\begin{tabular}{|c|c|}

    \hline
      Method&  Covering \\
    \hline
      Torque&  {\bfseries 0.429}\\
      N-Cut &  0.390\\
    \hline

\end{tabular}
\end{center}
\end{table}

\subsubsection{Figure-ground Segmentation}

The strengthened edges are expected to be useful for figure-ground segmentation because object boundaries are emphasized.
Here we demonstrate an edge-based graph cut algorithm using the strengthened edges.
For a quantitative evaluation of figure-ground segmentation, we used the dataset by Stein \etal \cite{Stein2008}, which has  ground truth segmentations for multiple foreground objects.
For each reference image in the data set we selected a single  foreground object, and used the centroid of the object  as  fixation point.
Then we applied the fixation based segmentation algorithm of \cite{Mishra2009b}. This algorithm separates foreground from background using a graphcut on a probability map of edges in a polar coordinate system.
Different visual cues were  used in  the graph-cut segmentation for comparison: the  Canny edge map, the boundary probability map (Pb)  \cite{Martin2004}, a strengthened edge map using  Canny edges, and a strengthened edge map using  Pb edges.
So we can separate the effect  of the torque measure, we used as strengthened edge directly  the normalized torque value contribution, $d_\tau$.
The quality of segmentation was evaluated by the segmentation covering \cite{Arbelaez2009}.
 In the case of foreground-background segmentation this measure amounts to the ratio of the true positive area and the union of computed segmentation and ground truth segmentation.

\begin{figure}[htbp]
  \begin{center}
    \begin{tabular}[t]{@{}c@{\,}c@{\,}c@{\,}c@{}}
      \begin{minipage}{0.23\columnwidth}
      \begin{center}
      \includegraphics[width=1\textwidth,keepaspectratio=true,clip]
      {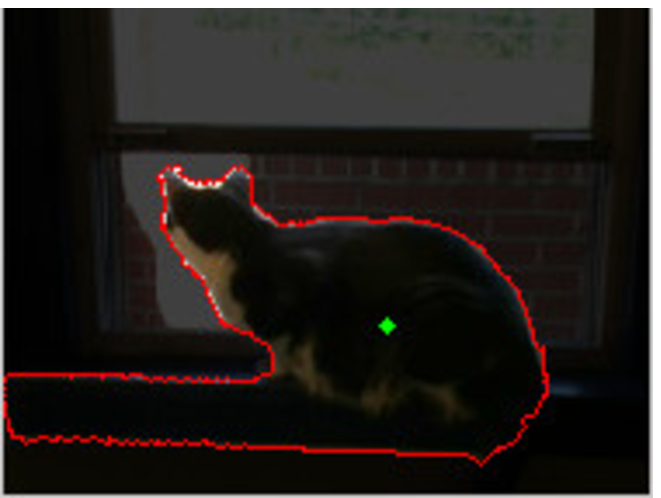}
      \end{center}
      \end{minipage}&

      \begin{minipage}{0.23\columnwidth}
      \begin{center}
      \includegraphics[width=1\textwidth,keepaspectratio=true,clip]
      {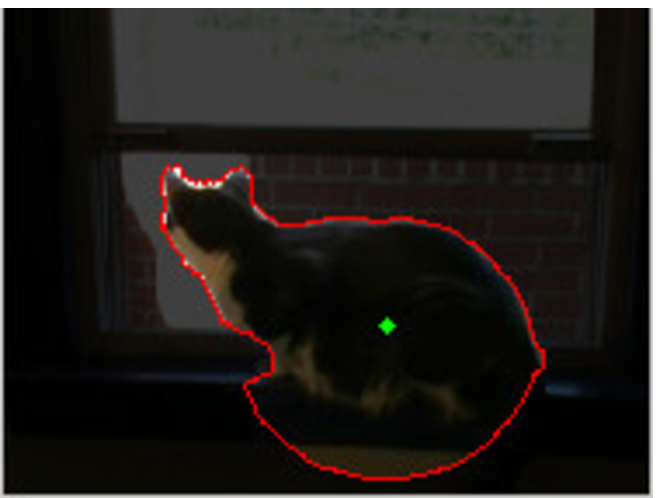}
      \end{center}
      \end{minipage}&

      \begin{minipage}{0.23\columnwidth}
      \begin{center}
      \includegraphics[width=1\textwidth,keepaspectratio=true,clip]
      {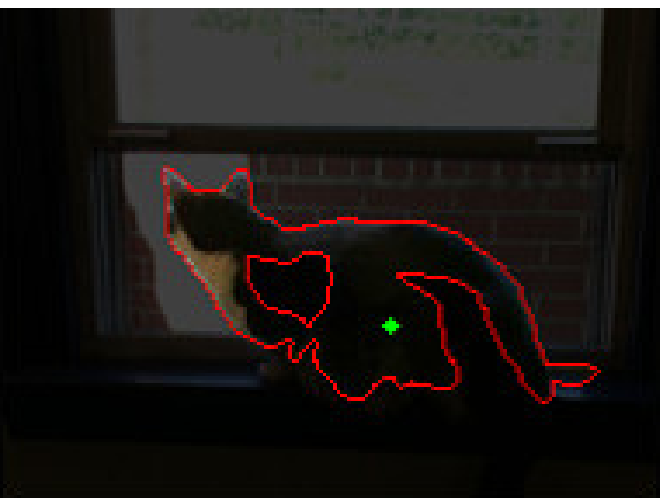}
      \end{center}
      \end{minipage}&

      \begin{minipage}{0.23\columnwidth}
      \begin{center}
      \includegraphics[width=1\textwidth,keepaspectratio=true,clip]
      {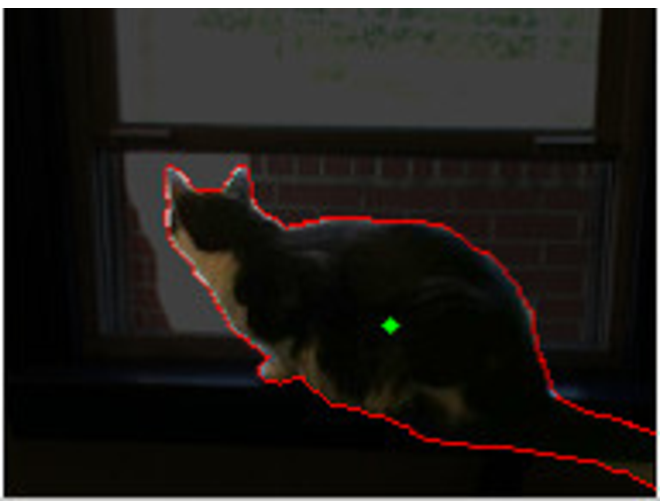}
      \end{center}
      \end{minipage}\\

      \begin{minipage}{0.23\columnwidth}
      \begin{center}
      \includegraphics[width=1\textwidth,keepaspectratio=true,clip]
      {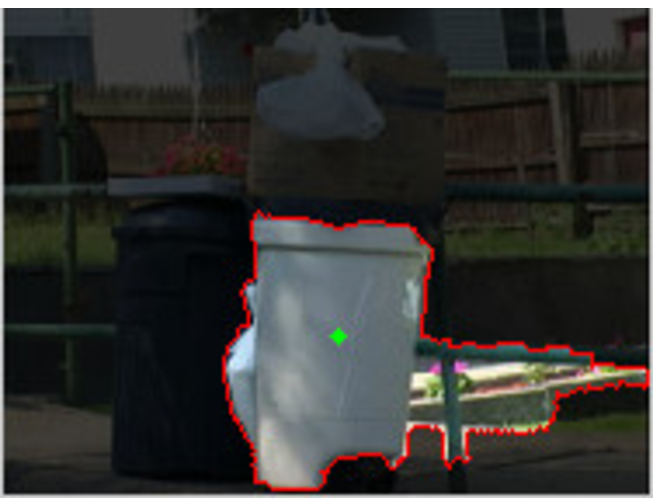}
      \end{center}
      \end{minipage}&

      \begin{minipage}{0.23\columnwidth}
      \begin{center}
      \includegraphics[width=1\textwidth,keepaspectratio=true,clip]
      {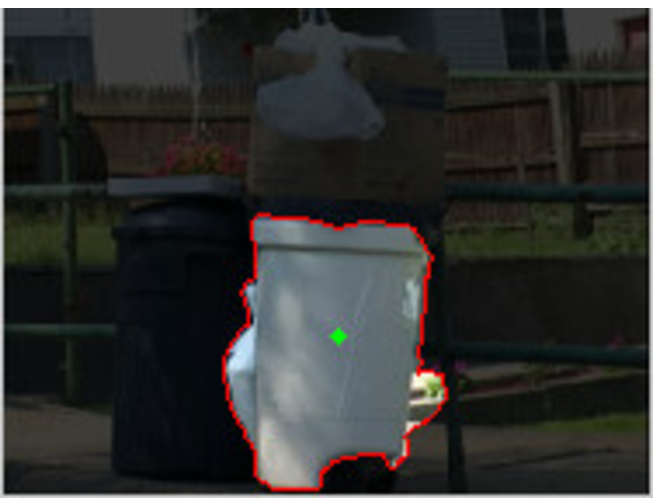}
      \end{center}
      \end{minipage}&

      \begin{minipage}{0.23\columnwidth}
      \begin{center}
      \includegraphics[width=1\textwidth,keepaspectratio=true,clip]
      {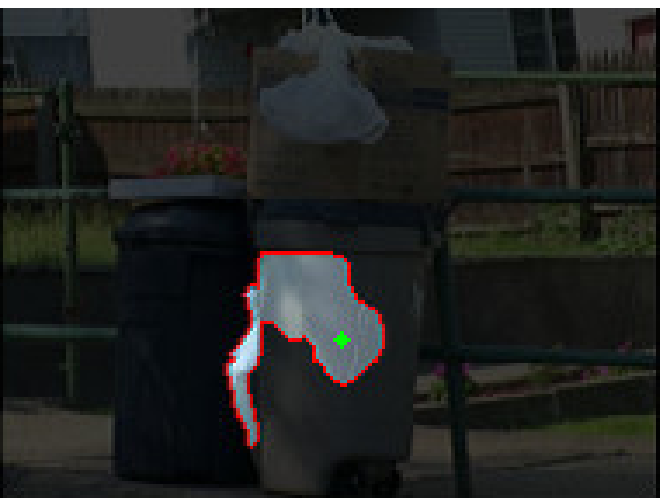}
      \end{center}
      \end{minipage}&

      \begin{minipage}{0.23\columnwidth}
      \begin{center}
      \includegraphics[width=1\textwidth,keepaspectratio=true,clip]
      {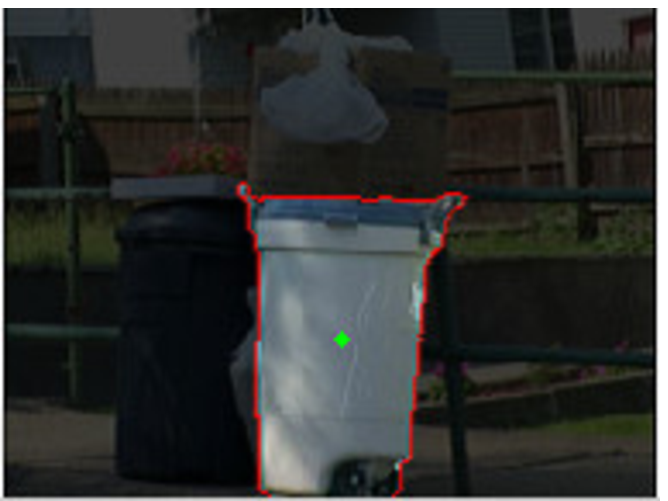}
      \end{center}
      \end{minipage}\\

      (a) & (b) & (c) & (d)
    \end{tabular}
  \end{center}
\caption{Example of a segmentation: (a)  using Canny edges (b) using strengthened edges by the torque (c) using the Chan-Vese method \cite{Chan2001}. (d) Ground truth. The green dot denotes the fixation point.
\label{fig:Examples of Segmentations}}
\end{figure}

Table 3 shows the results of the comparison for the dataset. For each visual cue
the average covering over  28 test images is shown.
As can seen from the table, adding the torque significantly improves the segmentation.
Finally, we also compared with the non-edge based level-set segmentation by Chan and Vese \cite{Chan2001}.
From the the performance of this method, we can see that the segmentation of objects for this data
set, given only the fixation point, is a challenging task.
Examples of segmentation results are shown in Fig.~\ref{fig:Examples of Segmentations}.

\begin{table}[tb]
\begin{center}
\caption{Comparison  of foreground-background  segmentations  based on the  covering criteria. The notation \emph{Torque (Canny)} and \emph{Torque (pb)} refer to  strengthened edges by the torque operator using as base  Canny or  the pb edges, respectively.}
\begin{tabular}{|c|c|}

    \hline
      Visual Cue&  Covering \\
    \hline
      Canny&   0.32\\
      Torque (Canny)&  {\bfseries 0.47}\\
    \hline
      pb&   0.40 \\
      Torque (pb)&  {\bfseries 0.48} \\
    \hline
      Chan-Vese& 0.21\\
    \hline

\end{tabular}
\end{center}
\label{tbl:Covering of Foreground Segmentation with Different Visual Cues}
\end{table}

\subsection{Recognition}
\label{sec:rec}

This section summarizes  experiments from \cite{XuY2012}, where we used the torque mechanism for  object  recognition in the bag-of-features framework. The approach consists of two steps: a patch detection, and a patch description.  First, torque value extrema in space and scale and their corresponding patches are detected.  These  patches are separated into minima, i.e the dark patches on brighter background and maxima, i.e. bright patches on darker background. This detection scheme  is called \emph{Maximal/Minimal Torque Patch (MTP) detector}.

\begin{figure*}[t]
\begin{center}
\begin{tabular}{cccc}
    \includegraphics[height=2.3in]{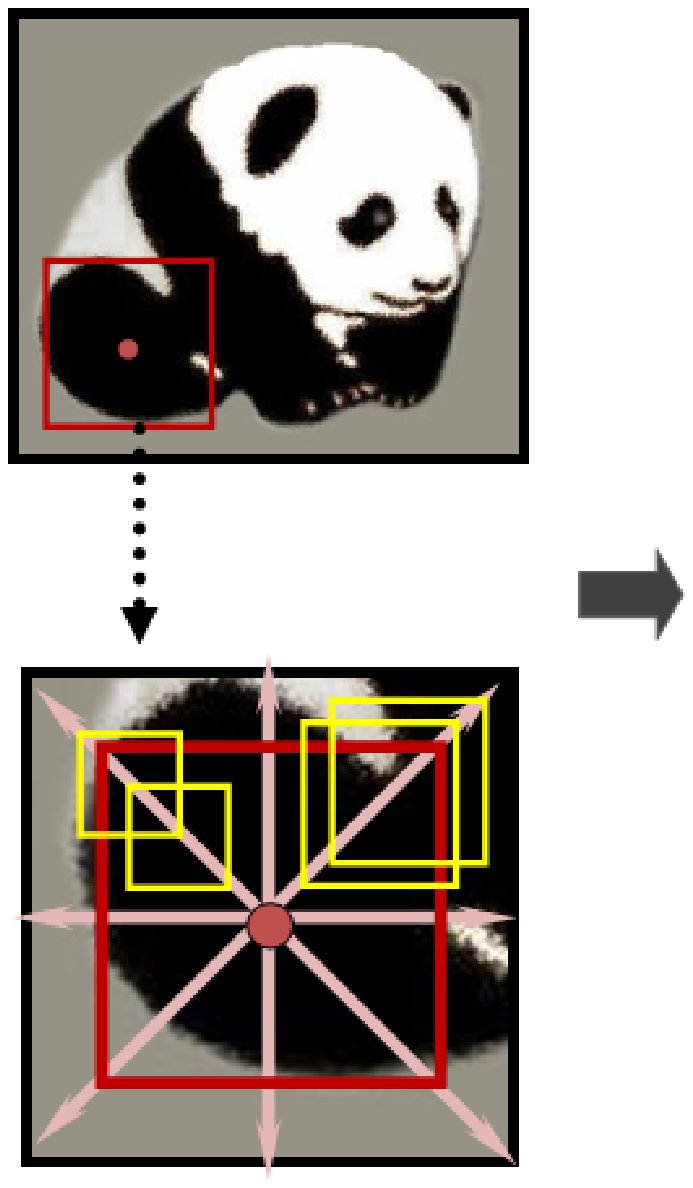}&
    \includegraphics[height=2.3in]{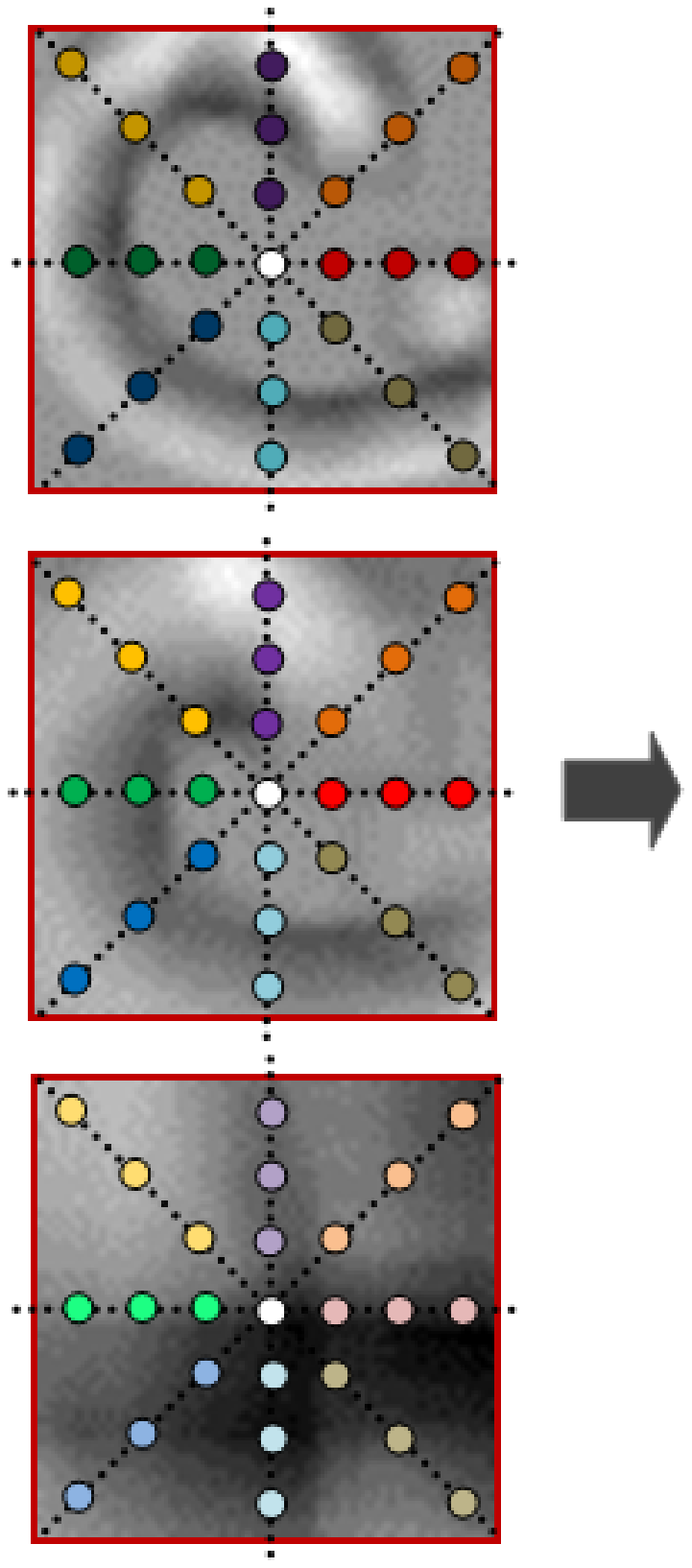}&
    \includegraphics[height=2.3in]{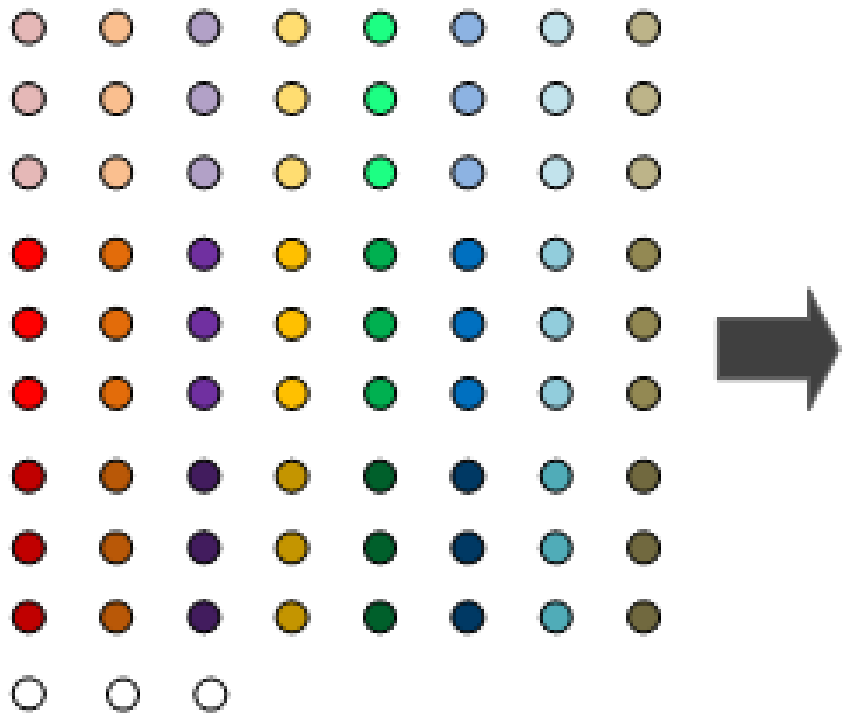}&
    \includegraphics[height=2.3in]{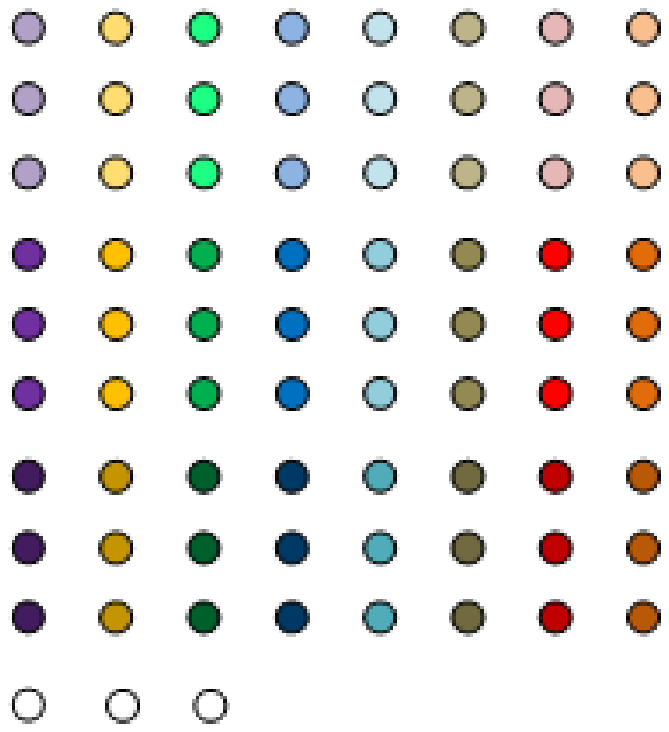}\\
\tiny{(a) An MTP patch}&\tiny{(b) Torque magnitudes}&\tiny{(c) Multi-scale sampling}&\tiny{(d) Orientation alignment}
\end{tabular}
\end{center}
\caption{Outline of MST descriptor. From left to right: (a) An interesting patch is detected by the MTP detector.
  (b) The torque magnitudes of regions  centered at points inside the detected patch are computed.
  (c) The torque values  are  sampled along 8 directions at several scales.
  The sampled values are collected  and concatenated as the local feature of the MTP patch.
  (d) Alignment of the orientation of the feature by circular-shifting.}
  \label{fig:MST framework}
\end{figure*}

Second,  the density and variance of the local edge structure is described in a multi-scale manner in the so-called   \emph{Multiscale Torque (MST) descriptor}. For a given patch   regions of multiple sizes  having an overlap with the patch along the eight axes at discrete
space intervals are selected and their torque values are concatenated into a vector. To keep the number of selected areas the same for all patches, the step size is adapted to the patch size, and  to make it robust to rotation, the patch is rotated such that its x-axis becomes the direction
closest in direction to the vector pointing from the patch center
to the centroid of the edges inside the patch P
(see  Fig.~\ref{fig:MST framework}).

 This descriptor  was evaluated  for  object recognition following  the  {\it bag of features} (BoF) representation paradigm evaluated on the Caltech-101 dataset.
 Table \ref{table:rec} shows the performance of the  method in comparison to a number of   top feature-based recognition methods (which were also  used for comparison in \cite{WYYLHG10}).
 As can be seen,  by itself the torque-based method performs about the same as \cite{JainKG08}, but does not perform as well as methods based on SIFT-based features (such as \cite{BoimanSI08cvpr},  \cite{YangYGH09}, and \cite{WYYLHG10}). This is because, these images are better captured by texture content (extracted through SIFT features). However by combining  the contour-based feature with SIFT, better performance was achieved. Our  implementation of the SIFT feature followed that of \cite{WYYLHG10}, and  we combined the  two features by concatenating them into a single vector and weighing them 1:2 (ours v.s. SIFT). Our method had  an  additional  $2.45\%-4.66\%$ accuracy gain over the best results of other methods with respect to different sizes of the training set. These results demonstrate that the proposed contour-based feature does capture meaningful
information of object contour and adds to objection recognition.
\begin{table*}\label{table:result_101}
\begin{center}
{
\begin{tabular}{|c||c|c|c|c|c|c|}
\hline
& 5 & 10 & 15 & 20 & 25 & 30\\ \hline\hline
Jain~\etal \cite{JainKG08} & - & - & 61.00 & - & - & 69.10 \\ \hline
Boiman~\etal \cite{BoimanSI08cvpr} & - & - & 65.00 & - & - & 70.40 \\ \hline
Yang~\etal \cite{YangYGH09} & - & - & 67.00 & - & - & 73.20 \\ \hline
Wang~\etal \cite{WYYLHG10} & 51.15 & 59.77 & 65.43 & 67.74 & 70.16 & 73.44 \\ \hline\hline
Ours & 48.17 & 57.65 & 62.33 & 65.32 & 67.39 & 68.97\\ \hline
Ours + SIFT & \textbf{53.60} & \textbf{64.01} & \textbf{69.15} & \textbf{72.40} & \textbf{74.52} & \textbf{76.22} \\
\hline
\end{tabular}
}
\end{center}
\caption{Classification accuracy for different methods on the Caltech-101 dataset.}
\label{table:rec}
\end{table*}

\section{Conclusion}

We introduced a new tool of mid-level vision, the \emph{Torque Operator}, and explored its fundamental properties using both theory and experiments.
The torque operator creates maps, which encode the structure of edges within patches, and it tends to generate larger absolute values when edges are aligned in a way surrounding the center of the patch, and the region enclosed by these edges matches the size of patch. This basic property was first discussed and examined on synthesized images. Then, this  property was demonstrated in the  applications of attention, boundary detection,  segmentation, and object recognition.
Experiments showed that for all these  applications the torque operator enhances the performance. An efficient implementation of the torque based on integral images has  also been provided (see \cite{code}).
\section{Extension and Outlook}
We believe that the concept of a grouping mechanism implemented via image processing operations is a powerful concept, and there are many way this concept can be generalized.

This paper discussed a number of  ways on how to use the Torque  operator in bottom-up processing for various  applications. An immediate  extension would be to use  the information it provides in different ways. For example the torque value itself is also expected to be a useful visual cue for segmentation. Furthermore, the torque operator could be applied to image gradient maps instead of edge maps. Such a torque measure can be derived as the  average brightness difference between the  inside  and the boundary of an image patch (see Appendix A).

The general concept of a mid-level operator acting as grouping mechanism can be  developed along a number of directions.
We could have  various mechanisms tuned to different edge configurations. For example, instead of adding curve contribution, such that circles are favored, we can add them such that  radial lines or spirals are favored, and this way create a set of operators tuned to semi-global patterns. We demonstrated such operators for boundary ownership classification \cite{Teoboundary2015}, but this generalization  could be useful also for texture or object recognition.

While in this paper we applied the torque  to edges of single images, the torque operator is applicable also to geometric edges, such as depth edges and motion edges. Because these edges are due to depth discontinuities, which  usually are at the  boundary of surfaces, it is expected that such a torque operator  will perform better for finding objects.

It is clear that pure bottom-up processing has its limitations, and object recognition or object segmentation  in single images requires higher level knowledge. The torque mechanism as a mid-level grouping tool can be modified also to interact with  higher-level processes encoding semantic information. We believe that there is room for research that embeds mid-level operators into object recognition and learning. Such an approach was demonstrated in  \cite{Teo2013} and \cite{Teo2015}, where  torque-like operators  were tuned to  specific shapes.
Contour processing of specific object classes using generalized torque then proceeds in two steps. First the bottom-up torque acts as an attention (or saliency) mechanism. Then  torque-like operators, that are tuned to respond to learned categories of objects, object parts,  or attributes  can be used in task-guided top-down  processing.
\section{Acknowledgements}
The support of the European Union under the Cognitive Systems program (project POETICON++), the National Science Foundation under INSPIRE
grant SMA 1248056, and  DARPA through U.S. Army grant
W911NF-14-1-0384 are gratefully acknowledged.  \newline
 The authors would like to thank Daniel Dementhon for sharing his code, which originated this project.
\appendix

\section{Torque based on Gradient}
\label{sec:gradient}

In previous sections, the torque was defined on edges, where edges were normalized as unit vectors.
Next we consider a slightly different formulation of the torque, which considers the strength of edges.
Images have strong edges and weak edges, and the strength is defined by the gradient. 
As will be shown next, a torque  of patches defined on gradients can be computed simply from the intensity in the area and at  the boundary of the patch.

The {\it gradient torque} is defined as follows:
\begin{align}
\tau_{pq} = \vec{r}_{pq} \times \tilde{\nabla} I\left(q\right),\label{eq:torque of gradient}
\end{align}
where $p$ is the center point and
$q$ is some  point on the image.
$\vec{r}_{pq}$ is the displacement vector from $p$ to $q$.
$\tilde{\nabla} I\left(q\right)$ is the image gradient at $q$ rotated by  90 degree, i.e. $\left(\frac{\partial I}{\partial y},-\frac{\partial I}{\partial x}\right)$, representing an edge-like vector.
Similarly, the {\it gradient torque  for a patch} is defined as:
\begin{align}
\tau_{P,p} = \frac{1}{2\left|P\right|}\int_{q\in P} \tau_{pq} dq,\label{eq:torque of gradient for patch}
\end{align}
where $P$ is the patch, and the integral is taken over the patch.

Next we show that the gradient torque of a patch can be computed as the
difference of the average image intensity inside the patch and the average image intensity on the patch boundary.
Considering a  disk patch, it thus amounts to:
\begin{align}
\tau_{P,p} &= \frac{1}{\pi R^2}\int^{R}_{0}\int^{\pi}_{-\pi}I\left(r,\theta\right)\cdot rd\theta dr \nonumber\\
&- \frac{1}{2\pi R}\int^{\pi}_{-\pi}I\left(R,\theta\right)\cdot Rd\theta,\label{eq:torque intnsity difference}
\end{align}
where $I\left(r,\theta\right)$ is the image intensity at $\left(r,\theta\right)$ in the  polar coordinate system.
Without loss of generality, it is  assumed that the center of the  patch is the origin of the coordinate system so that $\vec{r}_{pq} = \left(x,y\right)$ for $q=\left(x,y\right)$.
Then, using the substitution $\left(x,y\right)=\left(r\cos\theta,r\sin\theta\right)$, eq.(\ref{eq:torque intnsity difference}) can be  deduced as follows:
\begin{align}
\tau_{P,p} &= \frac{1}{2\pi R^2}\int^{R}_{0}\int^{\pi}_{-\pi}
\left\{x\left(-\frac{\partial  I}{\partial x}\right)-y\frac{\partial I}{\partial y}\right\}rd\theta dr\\
&= \frac{1}{2\pi R^2}\int^{R}_{0}\int^{\pi}_{-\pi}\cos\theta\left(\sin\theta\frac{\partial I}{\partial \theta}-r\cos\theta\frac{\partial I}{\partial r}\right)\\ \nonumber
& - \sin\theta\left(\cos\theta\frac{\partial I}{\partial \theta}+r\sin\theta\frac{\partial I}{\partial r}\right)r d\theta dr\\
&= - \frac{1}{2\pi R^2}\int^{R}_{0}\int^{\pi}_{\pi} r^2\frac{\partial I}{\partial r} d\theta dr\\
&= -\frac{1}{2\pi R^2} \int^{\pi}_{-\pi} \left\{R^2I\left(R,\theta\right)-\int^{R}_{0}2rI\left(r,\theta\right) dr\right\}d\theta. \label{eq:forty}
\end{align}
Eq. (\ref{eq:forty})  is equivalent to eq. (\ref{eq:torque intnsity difference}), which concludes the proof.

\bibliographystyle{plain}
\bibliography{references}

%
%

\end{document}